\documentclass[11pt,a4paper,oneside]{amsart}
\usepackage[colorlinks,citecolor=blue,urlcolor=blue]{hyperref}
\usepackage[foot]{amsaddr}
\usepackage[
hmarginratio={1:1},     
vmarginratio={1:1},     
textwidth=450pt,        
textheight=700pt,
heightrounded,          
]{geometry}
\usepackage[utf8]{inputenc}
\usepackage[T1]{fontenc}
\usepackage[english]{babel}

\usepackage{subcaption}
\usepackage{caption}
\usepackage{graphicx}
\usepackage{amsmath}
\usepackage{amsfonts}
\usepackage{amssymb}
\usepackage{afterpage}
\usepackage{xfrac}
\usepackage[authoryear]{natbib}

  \title{Statistical embedding: Beyond principal components}

  \author{Dag Tj{\o}stheim\textsuperscript{1}}
  \address{\textsuperscript{1}University of Bergen and Norwegian Computing Center}
  \email{Dag.Tjostheim@uib.no}

  \author{Martin Jullum\textsuperscript{2}}
  \address{\textsuperscript{2,3}Norwegian Computing Center}
  \email{Martin.Jullum@nr.no}

  \author{Anders L{\o}land\textsuperscript{3}}
  \email{Anders.Loland@nr.no}

\begin{document}

\begin{abstract}
There has been an intense recent activity  in embedding of very high dimensional and nonlinear data structures, much of it in the data science and machine learning literature. We survey this activity in four parts. In the first part we cover nonlinear methods such as principal curves, multidimensional scaling, local linear methods, ISOMAP, graph based methods and diffusion mapping, kernel based methods and random projections. The second part is concerned with topological embedding methods, in particular mapping topological properties into persistence diagrams and the Mapper algorithm. Another type of data sets with a tremendous growth is very high-dimensional network data. The task considered in part three is how to embed such data in a vector space of moderate dimension to make the data amenable to traditional techniques such as cluster and classification techniques. Arguably this is the part where the contrast between algorithmic machine learning methods and statistical modeling, the so-called stochastic block modeling, is at its greatest. 
In the paper, we discuss the pros and cons for the two approaches.  
The final part of the survey deals with embedding in $\mathbb{R}^ 2$, i.e.~visualization. Three methods are presented: $t$-SNE, UMAP and LargeVis based on methods in parts one, two and three, respectively. The methods are illustrated and compared on two simulated data sets; one consisting of a triplet of noisy Ranunculoid curves, and one consisting of networks of increasing complexity generated with stochastic block models and with two types of nodes.   
\newline

\noindent \textit{Key words and phrases:} Statistical embedding, principal component, nonlinear principal component, multidimensional scaling, local linear method, ISOMAP, graph spectral theory, reproducing kernel Hilbert space, topological data analysis and embedding, persistent homology, persistence diagram, network embedding, spectral embedding, Skip-Gram, neighborhood sampling strategies, visualization, $t$-SNE, LargeVis, UMAP   
\end{abstract}

\maketitle
\section{Introduction \label{Intro}}

With the advent of the Big Data revolution, the availability of data has exploded. The dimension of the data can be in the thousands, if not in the millions, and the relationships between data vectors can be exceedingly complex. Also, data are arriving in new forms. One recent addition to data types is network data, often internet based, sometimes with millions of nodes, and literally billions of edges (relationships between nodes). How does one understand the structure of such data sets? How can the essential structure of the data be preserved and characterized in an embedding in a possibly still high dimension, but much lower dimension than that of the original data set? How does one describe the interaction between various types of nodes in a way amenable to analysis? Another example is the analysis of porous media, in oil exploration say, or of astronomical or physiological data. Such data contain cavities and complicated geometric structures. Still another example is in natural languages with texts containing million of words. Is it possible to characterize language segments so as to discriminate one type of text from another?

All of these examples have to do with the characterization and simplification of highly complex and often unorganized data. From a mathematical and statistical point of view these tasks are examples  of embedding problems. One classic method is the traditional principal component analysis. It is still a much (the most?) used approach. For a number of low-dimensional characterization problems it works well, but in other situations it fails or simply cannot be applied, and other embedding methods must be sought. There has been a recent surge in methods beyond principal components initiated by the Big Data revolution, and the aim and motivation behind this  paper is to make a concentrated survey of such methods.

The goal of the survey could be said to be two-fold. First, to try to give a quite comprehensive survey of embedding methods and applications of these methods. Much of recent developments have taken place in the data science literature, including machine learning, and often published in proceedings of conferences. The second objective of this article has been to make the statistical community more aware of current methods in this branch of data science. We believe that there are potential synergy effects to be harvested. An example of such a development is the integration of so-called stochastic block models in embeddings of networks. The theory and use of stochastic block models have had a strong recent growth, and it has resulted in fruitful interaction between algorithmic modeling and statistical modeling. Clearly there is a need for more of this, since statistical methodology has been lagging behind. More specifically, one may think of finding better and more adequate measures of uncertainty inherent in some of these algorithmic methods, and to find possibly better rationales for some of the methods that presently do include ad hoc choices to be made in their implementation, in particular in setting of input or hyper parameters, and potentially bring fresh insight to issues such as heterogeneity and nonstationarity.

Here is a brief overview of the contents of the paper. Section
\ref{Principal} gives a brief summary of principal components and
points out some strengths and weaknesses. Included in the theory are
often assumptions of multivariate normal distribution and the use of
linear transformations. There are now a number of novel nonlinear
methods, some of them in fact with roots going far back in time. In
Section \ref{Nonlinear}, we look in particular at methods such as
principal curves and surfaces, multidimensional scaling, local linear
embedding, embedding via graphs (note that in this survey the terms
``graph'' and ``network'' will be used interchangeably), ISOMAP and
Laplace eigenmaps, diffusion maps, kernel principal components using reproducing kernel Hilbert spaces 
and random projections. Section \ref{Topological} has to do with the emerging field
of topological data analysis and topological manifold embedding. The
idea is to seek a type of embedding where topological and/or
geometrical patterns, including cavities of the data sets, are well
described. Section \ref{Network} deals with embedding of network data,
especially ultra high dimensional networks. This is a topic of great
practical interest, as can be understood from the recent advances within social network analysis. Networks have been examined
spanning from the famous karate club example, from networks of books
on American politics, both analyzed in \cite{newm:2006b}, to criminal
fraud networks, hidden cells in terrorist networks (see
e.g.\ \cite{morn:gigu:peti:2007} and \cite{budu:lee:kong:2015}) and
natural language analysis
(\cite{miko:suts:chen:corr:dean:2013b}). Some of the network research
publications have rapidly resulted in thousands of
citations in relevant fora, which are often found to be machine learning conference proceedings and internet based publications. 
Arguably this is the theme where the contrast between algorithmic machine learning methods and statistical modeling represented by stochastic block modeling is at its most pronounced, although recently  the gap has been narrowing. 
We discuss the pros and cons for the two approaches in Sections \ref{SBM1} and \ref{Data science}. In particular, 
in the last subsection of Section \ref{Network} we discuss algorithmic data science versus statistical modeling more generally, briefly reviewing recent advances in parametric
stochastic block modeling and autoregressive models applied to networks.
Open problems in heterogeneous, directed and dynamic networks are also briefly covered in  Section \ref{Network}. 

Finally, in Section \ref{Visualization}, we go on to the extreme case of having an embedding of dimension 2, the plane. This has to do with visualization, of course, and we are presenting three visualization methods, $t$-SNE, LargeVis, and UMAP, whose basis can be found in each of the preceding sections, namely nonlinear type embedding, network embedding and topological embedding. They are compared to principal component visualization.


To avoid an overlong paper some of the more technical and detailed
aspects of the surveyed methods are relegated to the Supplement \citep{supp}. We will also refer 
to previous review articles covering parts of our material, again found mainly in the data science literature and with the emphasis more on algorithms rather than statistical properties and concepts. To our knowledge ours is the first of such broad coverage. There are many unsolved statistical problems, and we will try to point out some of these as we proceed.

We have chosen to illustrate our methods by two types of simulation
experiments. First, a triple of noisy Ranunculoid (a concept
originating in flower forms in botany) curves encapsulated in one
another, cf.\  Fig.~\ref{fig:sub1a},  (a situation in which principal components do not work) illustrates a number of the nonlinear methods of Section \ref{Nonlinear} and the topological embedding of Section \ref{Topological}. As a
second example we have included a network based simulation, generated by stochastic block models, with
two types of nodes and varying degrees of complexity in their
interaction. Among other things these are used to illustrate and
compare the three visualization methods of Section \ref{Visualization}, for several choices of their input parameters. In the paper we also refer to real
data experiments that have been conducted especially in the network
embedding literature.  

In our treatment of embeddings in this paper we also seek to demonstrate at various points in the survey that there are important and challenging classes of problems of statistical modeling nature, whose solution will contribute to general advancement of the field. These problems have been summarized into three keypoints in Section \ref{Conclusion} on concluding remarks.

\section{Principal components \label{Principal}}

Principal component analysis (PCA) was invented by \cite{pear:1901} as an analogue of the analysis of principal axes in mechanics. It was later independently developed by Harold Hotelling in the 1930s, see e.g.\ \cite{hote:1933} and \cite{hote:1936}.

Given   $p$-dimensional observations $X_1,\ldots,X_n$, the Hotelling approach was along the lines that have since become standard: Let $X_i, i=1,\ldots,n$ have components $X_{ij},j = 1,\ldots,p$. The first principal component $V_1 = \{a_{j1}\}$ consists of the weights which gives the linear combination $\sum_{j=1}^{p}a_{j1}X_{ij}$  maximum variance subject to the constraint that the Euclidean norm $||V_1|| = 1$. The $k$th principal component $V_k = \{a_{jk}\}$ corresponds to the linear combination $\sum_{j=1}^{p} a_{jk}X_{ij}$ with the maximum variance subject to $||V_k|| = 1$, and it being orthogonal to previously found $V_j, 1 \leq j \leq k-1$. Or said in another way, the principal components constitute a sequence of projections in $\mathbb{R}^p$ of the data, mutually uncorrelated and ordered in variance.

Let $\boldsymbol \Sigma$ be the $p \times p$ population covariance matrix. Then it is well known, see e.g. \cite{joli:2002},  that the principal components $V_k$ are obtained by solving the eigenvalue problem
\begin{equation}
\boldsymbol \Sigma V_k = \lambda_k V_k,
\label{2a}
\end{equation}
where the largest eigenvalue $\lambda_1$ corresponds to the first principal component $V_1$, and where the variance explained by the $k$th principal component is given by $\lambda_k/\sum_{i=1}^{p}\lambda_i$. 

The estimated principal components are obtained by considering an estimate of $\boldsymbol \Sigma$. Let ${\bf X}$ be the $n \times p$ centered data matrix ${\bf X} = \{(X_{ij}-\bar{X}_j)\}$ with $\bar{X}_j = n^{-1}\sum_iX_{ij}$, then an estimate of $\boldsymbol \Sigma$ is obtained from $n^{-1}[{\bf X}^{T}{\bf X}]$, and the estimated eigenvectors and eigenvalues are obtained from
\begin{equation}
{\bf X }^{T}{\bf X} \hat{V} = \hat{\lambda}\hat {V}.
\label{2b}
\end{equation}
In practice, especially for high dimensions, it is computationally faster to use a singular value decomposition of the data matrix ${\bf X}$ itself.

The approach of Pearson is different, and the essence of his method is that he looks at a set of $m$  principal components as spanning a hyper-plane of rank $m$ in $\mathbb{R}^p$ such that the sum of the distances from the data points to this hyper-plane is minimized. The first principal component is then the line in $\mathbb{R}^p$ obtained by such a minimization. As will be seen it is the Pearson approach which is most amenable to generalizations to the nonlinear case. 

Before we close this section there is cause to ask why  linear principal component analysis is so useful. It is clearly the most used statistical embedding method. Why? 
There are several reasons for this. One is its potential to reduce the dimension of the original data. If a few principal components explain a large percentage of the variation, this in many cases means that the ensuing analysis can be concentrated to those components. These components can also be used henceforth in a factor analysis. And the number of needed components can often be decided by a clear cut percentage of variation explained, which, as was seen above, is straightforward  to compute given the eigenvalues of the covariance matrix.

The collection of $m$ principal components can also be used as feature extractors in clustering and classification problems where there is more than one class of observations involved. But it should be noted that the first principal components, although explaining most of the variance, may not generally be the best feature extractors for classification. This is because several classes may have essentially the same set of largest principal components, and that the discriminatory power is more concentrated in components  with a lower degree of explained variance.

Principal components have been used with great success in a number of
different fields, so diverse as e.g.\ quantitative finance,
neuroscience, meteorology, chemistry, and recognition of handwritten characters. Many applications and the basis of the theory are given in the book by \cite{joli:2002}. It is also quite robust and can work reasonably well for certain types of nonlinear systems as seen in the comparative review by \cite{vand:post:vand:2009}. 

However, there are also several shortcomings of linear principal components, which have inspired much recent research. The most obvious fault is the fact that it is a linear method, and data are often nonlinearly generated or located on or close to a submanifold of $\mathbb{R}^{p}$. This is sometimes aggravated by the fact that the PCA is based on the covariance matrix, and it is well-known that a covariance between two stochastic variables is not always a good measure of statistical dependence. This has been particularly stressed in recent dependence literature, a survey of which is given in \cite{tjos:otne:stov:2021}. Especially there exist statistical models and data where the covariance is zero although there may be a strong statistical dependence. An example is the so-called ARCH/GARCH time series models for financial risk whose dependence cannot be measured in terms of autocorrelation and cross-correlation function. 

To do statistical inference in PCA often a Gaussian assumption is added as well. For Gaussian variables the covariance matrix describes the dependence relations completely, so that it would be impossible to improve on the PCA embedding by a nonlinear embedding. But increasingly, data sets are appearing where the Gaussian assumption is not even approximately true. Moreover, in the age of Big Data the dimension of data may be extremely large, not making it amenable to principal component analysis which involves the solution of a $p$-dimensional eigenvalue problem. Finally, data may come in other forms such as networks. It is not clear how principal components can be applied under such circumstances.

As stated in the introduction, the purpose of this paper is to review a number of methods that can handle these problems. There has been an enormous growth in the literature recently both methodologically and in applications to new situations.

\section{Nonlinear embeddings \label{Nonlinear}}

As briefly indicated above, there are a number of reasons why PCA may fail. Principal components are found by solving an eigenvalue problem based on variances and covariances of multidimensional data. If the dependence structure of the data is not well described by such second order quantities, in general a nonlinear embedding is recommended or even required. There are a variety of possible nonlinear dependence structures, for each of which there are particular nonlinear algorithms adapted to the given structure. Below we enumerate these very briefly, with more detailed coverage and references in the subsequent subsections.

For the so-called principal curve method \citep{hast:1984} the data are supposed to be concentrated roughly on a curve or more generally on a submanifold. Although in this case the data are not well represented by a linear model, they may still be well approximated by a local linear model giving rise to the LLE method \citep{rowe:saul:2000} or to ISOMAP \citep{tene:silv:lang:2000}. Alternatively, the data may lie on chained non-convex structures, see for instance the example in Figure 1. For such and similar structures one may try to map the dependence properties to a graph, leading to a Laplace eigenvalue problem \citep{belk:niyo:2002}, and in its continuation to diffusion maps \citep{coif:lafo:2006}. In still other situations it may be advantageous to use a nonlinear transformation of the data points, and then solve a resulting eigenvalue problem, as is done in kernel principal components \citep{scho:smol:mull:2005}. One of the classical nonlinear methods is multidimensional scaling (MDS) \citep{torg:1952}, where an embedding is sought by preserving distances between individual data points. A combined linear and distance preserving method is represented by random projections, whose rationale is based on \citet{john:lind:1984}. All of these methods are presented in more details in the following subsections and most are illustrated in Figure 1. Figure 3 contains examples of embeddings of the data in Figure 1 by using topological data analysis presented in Section \ref{Topological}.

\subsection{Principal curves and surfaces\label{Principal curves}}

As mentioned in Section \ref{Principal} it is the Pearson's
hyper-plane fitting that is perhaps the best point of departure for
nonlinear PCA. Principal curves and surfaces were introduced in
\cite{hast:1984} and \cite{hast:stue:1989}. A brief summary is given
in \citet[pp. 541-544]{hast:tibs:frie:2009}.
Essentially, the idea is to replace the hyper-plane by a hyper-surface. It is simplest in the case of principal curves, generalizing the first principal component. Let $f(s)$ be a parameterized smooth curve in $\mathbb{R}^p$. The parameter $s$ in this case is a scalar and can for instance be arc-length along the curve. For each $p$-dimensional data value $X$, one lets $s_f(X)$ be the point on the curve closest to $X$. Then $f(s)$ is called a principal curve for the distribution of the random vector $X$ if
$$
f(s) = E(X|s_f(X)=s).
$$
This means that $f(s)$ is the average of all data points that project onto it. This is known as the self-consistency property. In practice it turns out (\cite{duch:stue:1996}) that there are infinitely many principal curves for a given multivariate distribution, but one is interested mainly in the smooth ones.

\subsubsection{Algorithm for finding one principal curve $f(s)$}
\begin{enumerate}
\item \textit{Definitions of coordinate functions and $X$.} Consider the coordinate functions $f(s) = [f_1(s),\ldots,f_p(s)]$ and let $X$ be the $p$-dimensional observational vector given by $X^T = (X_1,\ldots,X_p)$.
\item \textit{The two alternating steps.}
\begin{equation}
E(X_j|\hat{s}_f(X)=s) \to \hat{f}_j(s); \;j=1,\ldots,p 
\label{31a}
\end{equation}
and 
\begin{equation}
\mbox{argmin}_{s^{\prime}} ||X-\hat{f}(s^{\prime})||^2 \to \hat{s}_f(X).
\label{31b}
\end{equation}
\end{enumerate}
Here the first step \eqref{31a} fixes $s$ and enforces the self-consistency requirement. The second step \eqref{31b} fixes the curve and finds the closest point on the curve to each data point. The principal curve algorithm starts with the first linear principal component, and iterates the two steps in \eqref{31a} and \eqref{31b} until convergence is obtained using a given tolerated error. The conditional expectation in step \eqref{31a}  is determined by a scatter plot smoother by smoothing each $X_j$ as a function of arc-length $\hat{s}(X)$, and the projection in \eqref{31b} is done for each of the observed data points.

There are unsolved mathematical problems inherent in this method and proving convergence is in general difficult.

Principal surfaces have the same form as principal curves. The most commonly used is the two-dimensional principal surface with coordinate functions
$$
f(y_1,y_2) = [f_1(y_1,y_2),\ldots,f_p(y_1,y_2)].
$$
The estimates in step  \eqref{31a} and  \eqref{31b} above are obtained from two-dimensional surface smoothers. 
The scheme with a quantification of percentage reduction of variance seems to be lost in a principal curve and principal surface set-up.
A different but related approach is taken by \cite{ozer:erdo:2011}, 
where principal curves and surfaces are studied in terms of density ridges.
See also Section \ref{Manifold} for further developments and more references for the more general case of so-called manifold learning. 

\begin{figure}[ht!]
  \begin{center}
    \begin{subfigure}{.45\textwidth}
      \centering
      \includegraphics[width=1\textwidth]{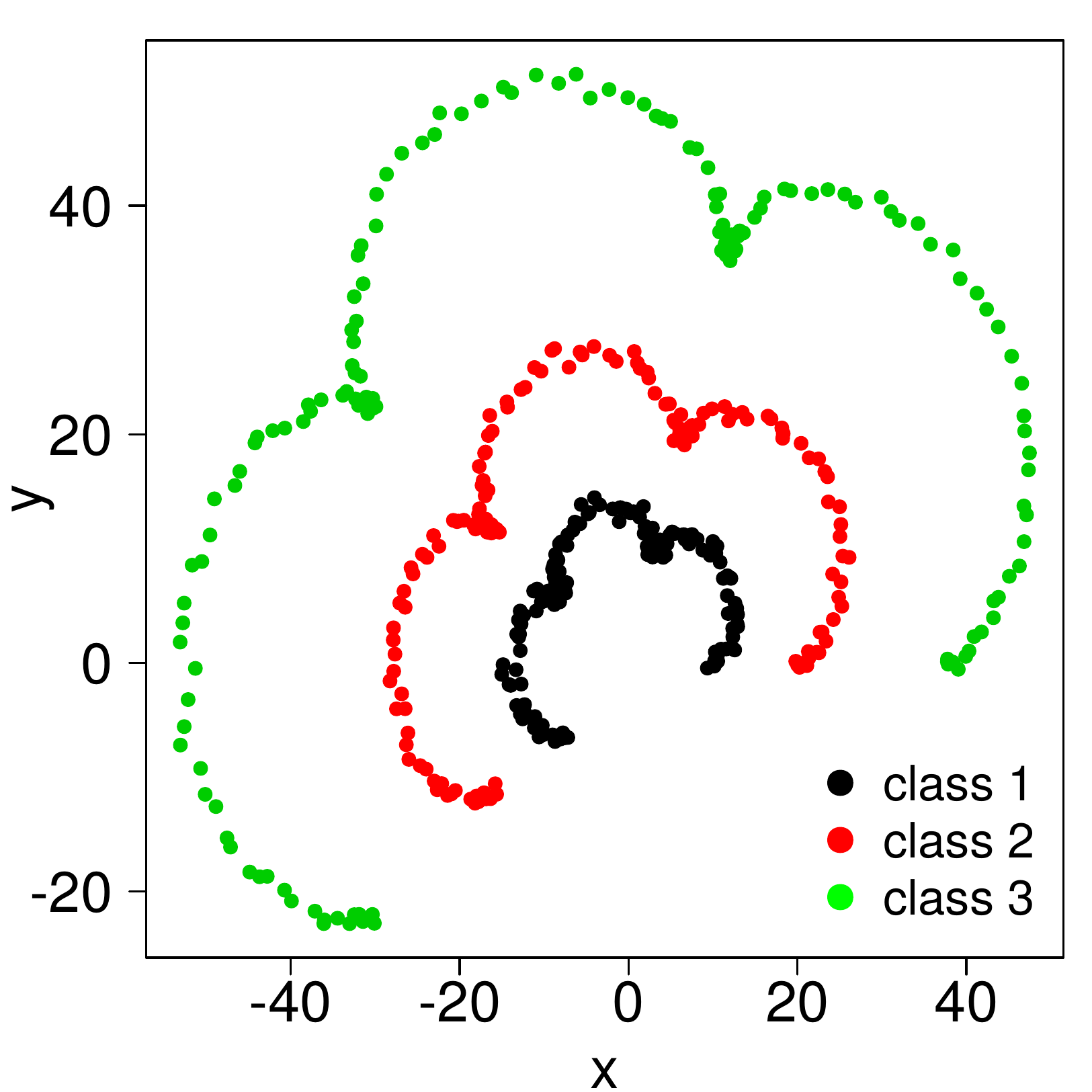}  
      \caption{The raw data.}
      \label{fig:sub1a}
    \end{subfigure}
    \begin{subfigure}{.45\textwidth}
      \centering
      \includegraphics[width=1\textwidth]{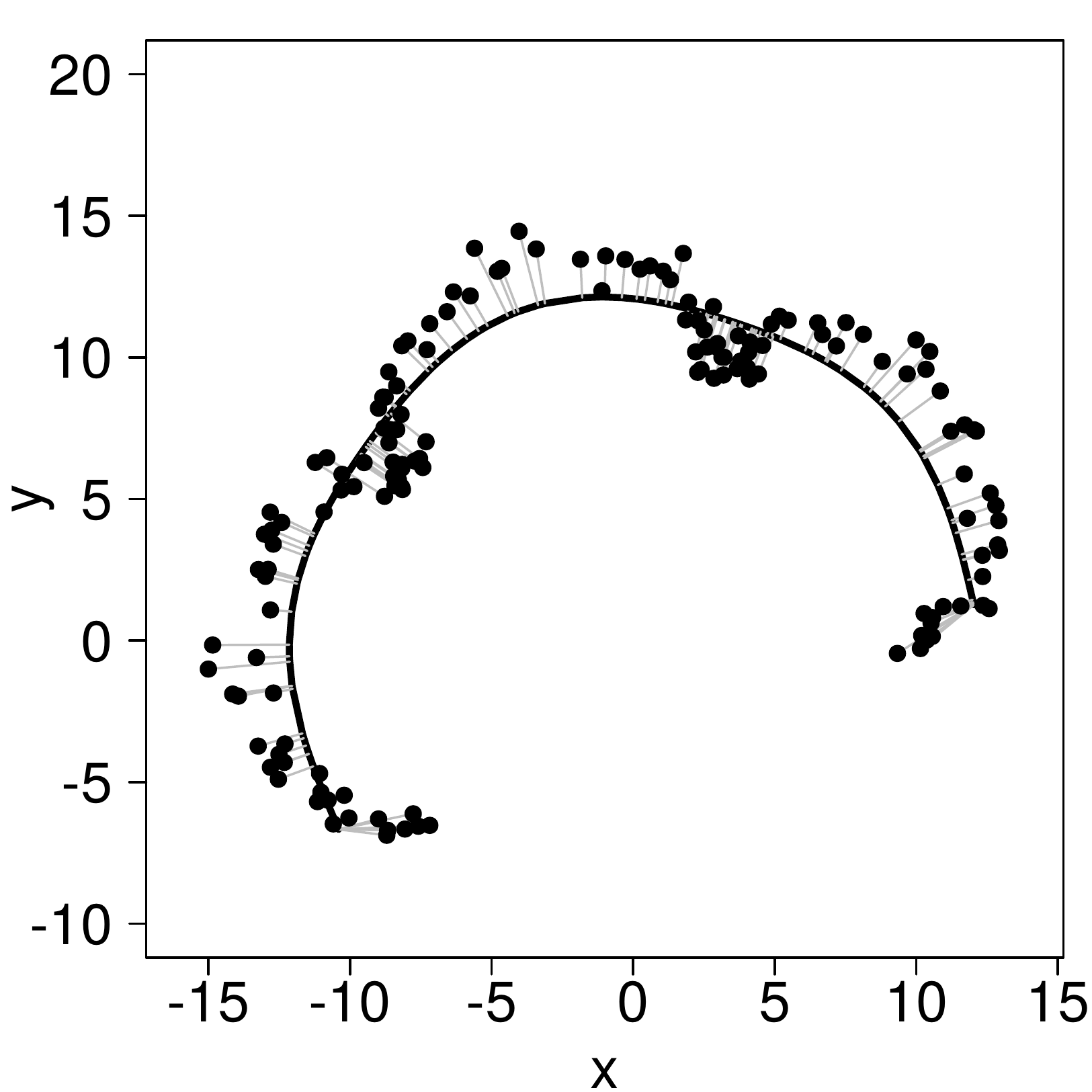}  
      \caption{Principal curve on the innermost curve of Fig.~\ref{fig:sub1a}.}
      \label{fig:sub1b}
    \end{subfigure}
    \begin{subfigure}{.45\textwidth}
      \centering
      \includegraphics[width=1\textwidth]{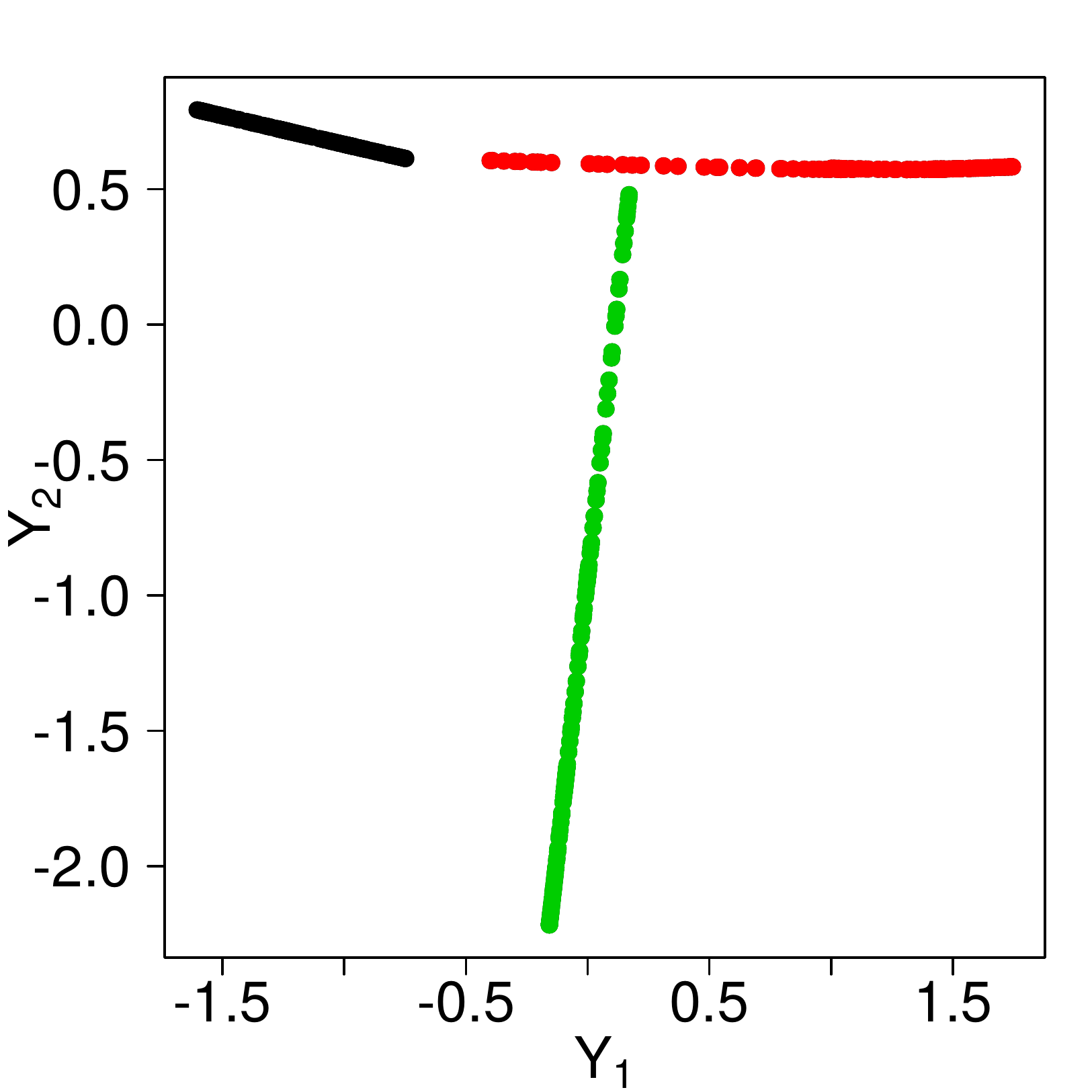}  
      \caption{Local linear embedding.}
      \label{fig:sub1c}
    \end{subfigure}
    \begin{subfigure}{.45\textwidth}
      \centering
      \includegraphics[width=1\textwidth]{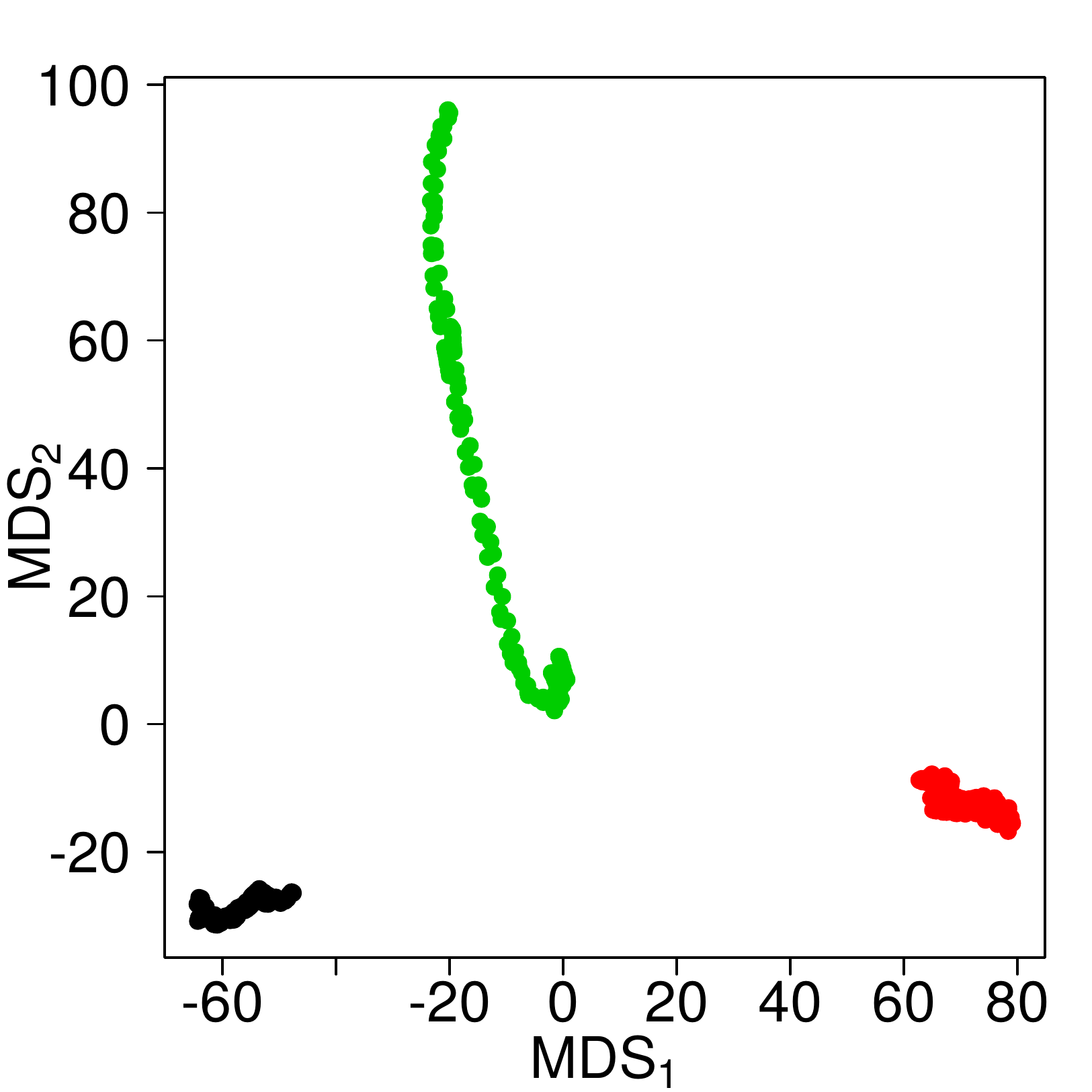}  
      \caption{ISOMAP.}
      \label{fig:sub1d}
\end{subfigure}
    \begin{subfigure}{.45\textwidth}
      \centering
      \includegraphics[width=1\textwidth]{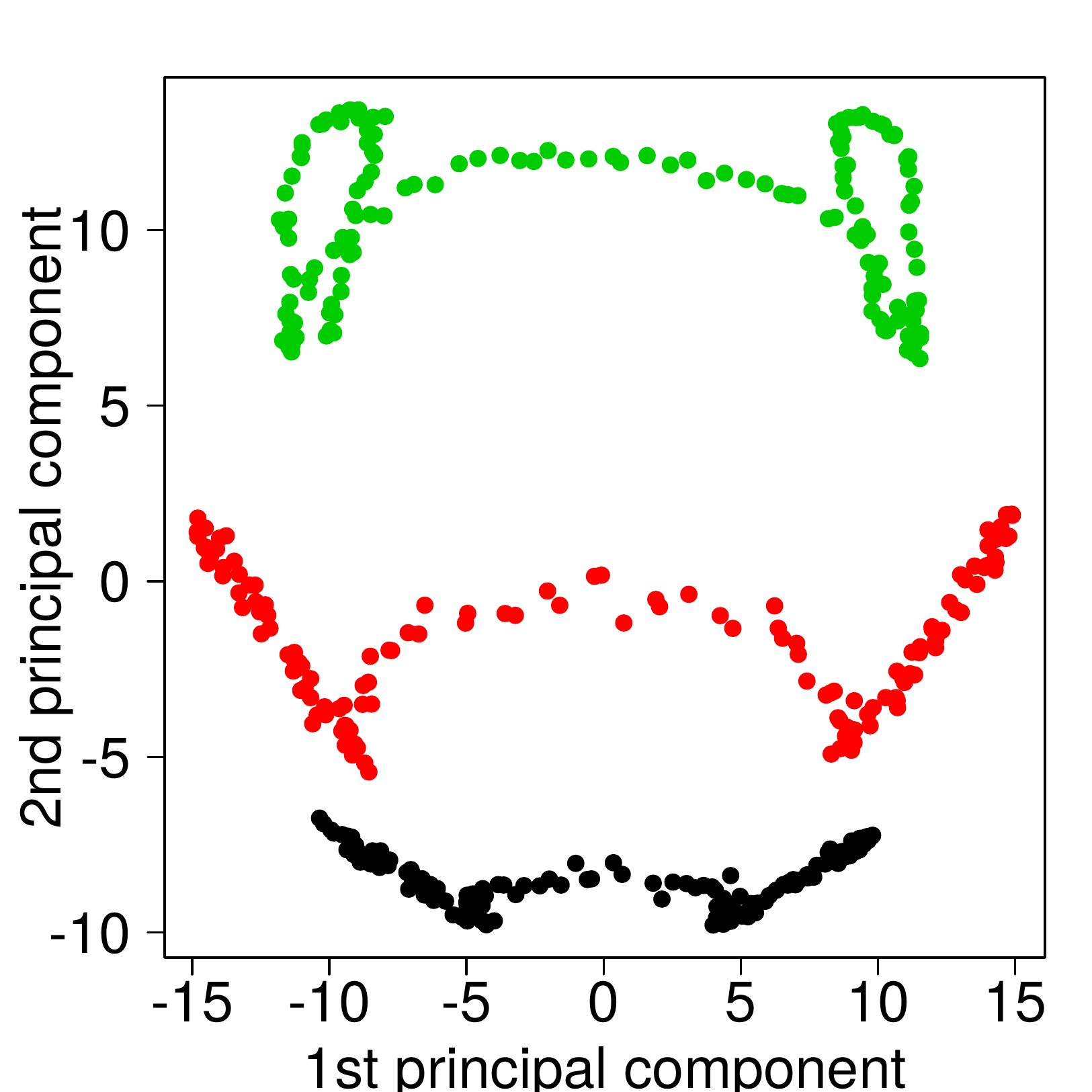}  
      \caption{Kernel principal components (with a Bessel kernel).}
      \label{fig:sub1e}
    \end{subfigure}
\end{center}
\caption{Four different embedding methods applied to three parametric curves from the so-called Ranunculoid and perturbed by Gaussian noise with a standard deviation of \sfrac{1}{2}.
 \label{fig:1}}
\end{figure}

In Fig.~\ref{fig:1} we present a data set that will be used for
illustration purposes throughout this section and also in Section
\ref{Topological} on topological data analysis. The raw data are
presented in Fig.~\ref{fig:sub1a}. It consists of parts of three parametric curves, each being obtained from the so-called Ranunculoid, but with three different parameter sets. In addition the curves have been perturbed by Gaussian noise. In Fig.~\ref{fig:sub1b} we have illustrated the construction of a principal curve on the innermost curve of Fig.~\ref{fig:sub1a}. It is seen that the main one-dimensional structure of the curve is well picked up, but it does not quite get all the indentions of the original curve. Compared to a linear principal regression curve it is a big improvement. (Note that a nonparametric regression is not an option here, since the x- and y-coordinates of Fig.~\ref{fig:sub1a} are on  the same basis, and there are several y-values for many of the x-values.) 

\afterpage{\clearpage}

\subsection{Multidimensional scaling\label{Multidimensional scaling}}

The idea of multidimensional scaling (MDS) goes far back, but it, or
similar ideas, has recently got a revival in statistical embedding
through algorithms such as LLE, ISOMAP (see the next subsections), and $t$-SNE (see Section \ref{Visualization}). It can be roughly formulated as finding suitable coordinates for a set of points given their mutual distances. This problem was first considered by \cite{youn:hous:1938}. 
These methods were further developed and applied to scaling of psychometric distances between pairs of stimuli by \cite{torg:1952}. 
A fine review of the essentials of multidimensional scaling is given in \citet[pp. 570-572]{hast:tibs:frie:2009}. 
Here the emphasis is on viewing multidimensional scaling as a general method for dimensionality reduction of data in $\mathbb{R}^p$. They therefore start with a set of observations $X_1,\ldots,X_n \in \mathbb{R}^p$ where $d_{ij}$ is some form of distance measure (not necessarily Euclidean) between observation $X_i$ and $X_j$. In fact, in the general theory of multidimensional scaling the $d_{ij}$ may be considered as a dissimilarity measure between objects (e.g psychological stimuli) $i$ and $j$. One example can be found in \cite{kuno:suga:1966}, where judgment of psychometric distance between piano pieces were represented as a configuration of points in two-dimensional Euclidean space.

From a dimension reduction point of view, multidimensional scaling seeks values $Y_1,\ldots,Y_n \in \mathbb{R}^m$, often $m=2$, as in the above piano piece example, for visualization purposes, by minimizing the so-called stress function
$$
S(Y_1,\ldots,Y_n) = \sum_{i \neq j}(d_{ij}-||Y_i-Y_j||)^2,
$$
which means choosing $\{Y_j, j =1,\ldots,n\}$ such that one strives to preserve distances when going from $\mathbb{R}^p$ to $\mathbb{R}^m$.
This is known as the least squares or Kruskal-Shephard scaling. A gradient descent algorithm can be used to minimize $S$. A variation on this is the so-called Sammon mapping,  \cite{samm:1969}, which minimizes 
$$
S_{Sm}(Y_1,\ldots,Y_n) = \sum_{i \neq j} \frac{(d_{ij}-||Y_i-Y_j||)^2}{d_{ij}}.
$$
In so-called classical scaling one starts instead  with similarities $s_{ij}$. Classical scaling is not equivalent to least squares scaling. The loss functions are different, and the latter mapping can, in contradistinction to PCA, be nonlinear. 


It should be noticed that in the case of multidimensional scaling it is an embedding from $\mathbb{R}^p$ to $\mathbb{R}^m$, we have an embedding from one Euclidean space to another. On the other hand principal surfaces as in Section \ref{Principal curves} and many of the other methods in this survey consider embedding from $\mathbb{R}^p$ to a lower dimensional {\it manifold}. In particular, this is the case in Sections \ref{LLE} and \ref{ISOMAP}. Preserving distance is a key factor of the random projection method to be treated in Section \ref{Random projection}.


\subsection{LLE -- Local linear embedding\label{LLE}}

Principal curves and surfaces represent an early example of local modeling and manifold embedding. Manifold embedding will be taken up from a more general point of view in Section \ref{Topological} with its connections to recent advances in TDA (Topological Data Analysis). However, it is convenient at this point to briefly mention the early work of \cite{rowe:saul:2000}
that resembles the principal surface methodology in that it is a local
method. In fact, it is a local {\it linear} model, and locally linear
methods are well known and much used in nonparametric regression. But
here the viewpoint is different since there is no clearly defined
dependent variable. Actually in that respect, it is like the recent local Gaussian modeling of \cite{tjos:otne:stov:2021a}.

Local linear embedding (LLE) recovers global nonlinear structure from locally linear fits. Suppose that the data $X_1,\ldots,X_n$ are $p$-dimensional vectors sampled from an inherent $m$-dimensional manifold. One assumes that each data point  lies on or close to a locally linear patch of the manifold. The local geometry of these patches is characterized by linear coefficients that reconstruct each data point from its neighbors.

\subsubsection{The LLE algorithm}
The algorithm consists of three main steps:
\begin{enumerate}
\item \textit{Find the the nearest neighbors $N(i)$ of $X_i$}, for example by a nearest neighborhood algorithm, such as kNN (k-nearest neighbors).
\item \textit{Construct weights $w_{ij}$  by minimizing the cost function \eqref{33a}} subject to the constraint that $w_{ij}= 0$ if $x_j$ does not belong to the set of neighbors of $X_i$, and such that $\sum_j w_{ij} = 1.$ Weights for non-neighbors are 0.
 \begin{equation}
   M(w) = \sum_{i}||X_i-\sum_{X_j \in N(i)}w_{ij}X_j||^2,
   \label{33a}
 \end{equation}
\item \textit{Map each high dimensional observation $X_i$ to a low-dimensional vector $Y_i$} representing global internal coordinates on the manifold. This is done by choosing $m$-dimensional coordinates to minimize the embedding cost function 
\begin{equation}
M(Y) = \sum_i||Y_i-\sum_j w_{ij}Y_j||^2. 
\label{33b}
\end{equation}
It should be noticed that this is a minimization problem over $Y$. The
$w_{ij}$ weights are known and equal to those obtained in step 2. Then, optimizing with respect to
$Y_i$ in (\ref{33b}) can be achieved by solving a sparse $m \times m$
eigenvalue problem.
\end{enumerate}
  
The assumption of \cite{rowe:saul:2000} is here that one can expect the $w_{ij}$-characterization of local geometry in the original data space to be equally valid for local patches of the manifold. In particular, the same weights $w_{ij}$ that reconstruct the $i$th data point in $p$ dimensions should also reconstruct its embedded manifold coordinates in $m$ dimensions.

It will be seen later in this survey that such a
two-step (steps 2 and 3) procedure is used also in other dimension reduction
algorithms, as in the $t$-SNE visualization routine described in
Section \ref{Network}. From Fig.~\ref{fig:sub1c} it is seen that the three parts of the Ranunculoid in Fig.~\ref{fig:sub1a} are clearly separated with LLE, especially in the $Y_2$-direction.

\subsection{Embedding via graphs and ISOMAP\label{ISOMAP}}

Some of the primary purposes of statistical embedding is to use the embedded vectors or coordinates for feature extraction, clustering and classification. The most used clustering method is probably the $K$-means algorithm. (See e.g.\ \citet[chapter 14.3]{hast:tibs:frie:2009}.) This method does not work well if the clusters form non-convex subsets of the data space. Examples of this are the clusters consisting of 3 concentric noisy circles in $\mathbb{R}^2$, or of the more complicated structure of the three curves in Fig.~\ref{fig:sub1a}.

For a given point cloud in $\mathbb{R}^p$ a method of circumventing such problems is to embed the points in a similarity graph or network. Given a set of data points $X_1,\ldots,X_n$, a similarity measure $s_{ij} \geq 0$ between $X_i$ and $X_j$ can simply be the Euclidean distance between $X_i$ and $X_j$, or there could be other  similarity measures. The intuitive goal of clustering is to divide the points into groups such that the similarity between two groups is weak, whereas the similarity between points within a group is typically strong. If we do have similarity information between the points, a convenient way to represent this is to form a similarity graph $G = (V,E)$. Each node $v_i \in V$ in the graph represents a data point $X_i$. Two nodes in the graph are connected if their similarity $s_{ij}$ is positive or exceeds a threshold. The similarities $s_{ij}$ then are weights $w_{ij}$ on the edges $E$ of the graph. The graph is undirected if $w_{ij} = w_{ji}$. The problem of clustering can now be reformulated  using the similarity graph: one wants to find a partition of the graph such that the edges between different groups have low weight, and the edges within a group have  high weights.

Given a point cloud in $\mathbb{R}^p$ there are several ways of constructing a corresponding similarity graph:

\begin{itemize}

\item[i)] The $\varepsilon$-neighborhood graph: Here one connects all points, and give them weight $w_{ij}=1$, that have pairwise distances less then $\varepsilon$. One can see at once that this would represent a possible solution to the clustering of three noisy concentric circles mentioned above if the noise is moderate.

\item[ii)] $k$-nearest neighbor graph: Here one can connect node $v_i$ with node $v_j$ if $v_j$ are among the $k$ nearest neighbors of $v_i$. Symmetrization leads to an undirected graph and $w_{ij} = s_{ij}$.

\item[iii)] The fully connected graph: All points with positive similarity are connected with each other, and we take $w_{ij} = s_{ij}$. As an example of a similarity measure one can take $s_{ij} = \exp(-||X_i-X_j||²/2\sigma^2)$, where $\sigma$ is a parameter that controls the strength of the similarity. 

\end{itemize} 
Two early references for the use of graph embedding are 
\cite{tene:silv:lang:2000} and \cite{desi:tene:2002}.

\subsubsection{The ISOMAP algorithm}
It is described in  \cite{tene:silv:lang:2000}. Apart from clustering, it has gained considerable use as a nonlinear dimension reduction method, by combining graph representation with multidimensional scaling seeking distance preservation, see op.~cit.\ references for details.

The input is the distances $d_X(i,j)$ between all pairs of $X_i$ and $X_j$ of the $N$ data points. The output is $d$-dimensional vectors $Y_i$ in $\mathbb{R}^m$. The algorithm consists of three main steps:
\begin{enumerate}
\item \textit{Construct the neighborhood graph} $G$ according to i) or ii) above. Set edge lengths equal to $d_X(i,j)$.
\item \textit{Compute shortest paths}  $d_G(i, j)$  between all pairs in the graph $G$, for example by Dijkstra's algorithm or the Floyd–Warshall algorithm \citep{cormen:leisersion:rivest:stein:2022}.
\item \textit{Construct $m$-dimensional embeddings} $Y_i$ by applying multidimensional scaling from Section \ref{Multidimensional scaling} to the matrix of graph distances $D_G = \{d_G(i,j)\}$.
\end{enumerate}

The results of applying the ISOMAP algorithm on the curves in Fig.~\ref{fig:sub1a} are given in Fig.~\ref{fig:sub1d}. It is seen that the curves are well-separated both in the $\mbox{MDS}_1$ and $\mbox{MDS}_2$ directions.

\subsection{Graph representation and Laplace eigenmaps\label {Laplace eigenmaps}}

In this subsection we will just give a brief presentation of Laplace eigenmaps and graph spectral theory mainly based on \cite{belk:niyo:2002,belk:niyo:2003}.
Here the point of departure is, as it is for all of this section, a point cloud in $\mathbb{R}^p$, and then the aim is to reduce the dimension by searching for a manifold embedding of lower dimension. 

In Section \ref{Network} we will {\it start} with a network and use graph spectral theory to find an embedding of the network in Euclidean space or on a manifold such that it can subsequently be used for purposes of clustering and classification. A few more details of graph spectral theory will be given then.

To introduce Laplacian eigenmaps we need  some more graph notation: The weighted adjacency matrix of the graph is the matrix ${\bf A}= \{a_{ij}\},i,j = 1,\ldots,n$, where $a_{ij}= w_{ij}$ is the weight on the edge between nodes $v_i$ and $v_j$. If $a_{ij}=0$, this means that the nodes $v_i$ and $v_j$ are not connected by an edge. We still assume that the graph is undirected so that $a_{ij} = a_{ji}$. The degree of a node $v_i \in V$ is defined as 
\begin{equation}
d_i = \sum_{j=1}^{n}a_{ij} = \sum_{j=1}^{n}w_{ij},
\label{25i}
\end{equation}
with $a_{ii}=0$.  The degree matrix $D$ is defined as the diagonal matrix with the degrees $d_1,\ldots,d_n$ along the diagonal.
\subsubsection{The Laplacian eigenmap algorithm} The algorithm consists of three main steps:
\begin{enumerate}
\item \textit{A graph is constructed} using the strategy outlined in (i), (ii) or (iii) of Section \ref{ISOMAP}. This is used to establish the edges of the graph.
\item \textit{The weights of the edges are determined.} \cite{belk:niyo:2003} present two choices. The first choice, as in Section \ref{ISOMAP}, is to choose the so-called heat kernel
\begin{equation}
w_{ij} = \exp^{-||X_i-X_j||/t}
\label{34a}
\end{equation}
if the nodes are connected using the $\varepsilon$-strategy of Section \ref{ISOMAP}, and putting $w_{ij}=0$ if they are not connected. The kernel parameter $t \in \mathbb{R}$ is up to the user to choose. A second alternative is just to let $w_{ij}= 1$ if $v_i$ and $v_j$ are connected, and $w_{ij}=0$ if not. 
\item \textit{Find the Laplacian eigenmaps.}  Assume that the graph $G$ as constructed above is connected. If not, use the algorithm given below for each connected component. 
Define the Laplacian matrix by ${\bf L} = {\bf D}-{\bf A}$, where ${\bf D} = \{d_{ii}\}$ is the degree matrix defined above and ${\bf A} = \{a_{ij}\}$ are the weights of the adjacency matrix. The Laplacian is symmetric, positive semidefinite and can be thought of as an operator acting on functions defined on the nodes of the graph $G$. The Laplacian eigenmaps are then obtained by solving the eigenvalue problem
\begin{equation}
{\bf L}f_i = \lambda_i {\bf D}f_i,\; i= 0,1,\ldots,p-1,  
\label{34r}
\end{equation}
with 
$$
0 = \lambda_0 \leq \lambda_1 \leq \lambda_{p-1},
$$
where it is easily verified that 0 is a trivial eigenvalue corresponding to the eigenvector $f_0 = [1,1,\ldots,1]$. This eigenvector is left out, and the next $m$ eigenvectors are used for an embedding in $m$-dimensional Euclidean space
$$
X_i \to \sum_{j=1}^m\langle X_i,f_j \rangle f_j,
$$
where $\langle \cdot,\cdot \rangle$ is the inner product in $\mathbb{R}^{p}$. The Laplacian eigenmaps preserve local information optimally in a certain sense \citep{belk:niyo:2003}. In fact these authors show that the Laplacian of a graph is analogous to the Laplace-Beltrami operator on manifolds, and they demonstrate that the eigenfunctions of the Laplace-Beltrami operator have properties desirable for embedding.
\end{enumerate}



\subsection{Diffusion maps\label{Diffusion maps}}

The representation of the Laplace matrix and the corresponding Laplace-Beltrami diffusion operator is just one way of finding a meaningful geometric description of a data set. As will be seen in this subsection, it is possible to introduce an associated Markov chain that can be used to construct coordinates called diffusion maps. By iterating the Markov transition matrix one obtains multiscale geometries that can be useful in the context of data parametrization and dimension reduction.

Following \cite{coif:lafo:2006}, it is convenient to think of the data set ${\bf X}$ as a measure space $({\bf X}, {\mathcal B},\mu)$ with an associated kernel $k$ satisfying $k(x,y) = k(y,x)$ and $k(x,y) \geq 0$. In terms of Section \ref{Laplace eigenmaps}, $k$ may be associated with the adjacency matrix ${\bf A}$, and $\mu(x)$ with the discrete measure with $\mu(x_i) = 1/n$, where $n$ is the number of observations. Generally we let $d(x) = \int_{\bf X} k(x,y)d\mu(y)$, which corresponds to the definition of degree in (\ref{25i}). One possibility for choosing $k$ is to choose the heat kernel in (\ref{34a}).

The next step is to introduce the probability transition distribution $p(x,y) = k(x,y)/d(x)$. Then clearly $\int_{\bf X} p(x,y)d\mu(y) = 1$, and $p$ can be viewed as a transition kernel of a Markov chain on ${\bf X}$. The operator $Pf(x) = \int_{\bf X} p(x,y)f(y)d\mu(y)$ is the corresponding diffusion operator.

Whereas $p(x,y)$ represents the probability of a one-step transition from node $x$ to node $y$, the probability of a transition from $x$ to $y$ in $L$ steps is given by the $L$-step transition $p_L(x,y)$, the kernel of the $L$-th power, $P^L$ of $P$. A main idea of the diffusion framework is that running the Markov chain forward in time, or equivalently, taking larger powers of $P$, will allow one to reveal relevant geometric structures of different scales.

The Markov chain has a stationary distribution, it is reversible, and if ${\bf X}$ is finite and the graph of the data is connected, then it is ergodic \citep[cf.][]{coif:lafo:2006}. Further, $P$ has a discrete sequence of eigenvalues $\{\lambda_i\}$ and eigenfunctions $\psi_i$ such that $1=\lambda_0 \geq \lambda_1 \geq \cdots$, and $P\psi_i = \lambda_i\psi_i$. This corresponds to the eigenvalue problem in (\ref{34r}).

Let $\pi(x)$ be the stationary distribution of the Markov chain. \cite{coif:lafo:2006} show that the family of so-called diffusion distances $\{D_L\}$ can be written as 
\begin{equation}
\label{diff1}
D_L(x,y)^2 =  \int_{\bf X} (p_L(x,u)-p_L(y,u))^2\frac{d\mu(u)} {\pi(u)} 
\end{equation}
$$
= \sum_{i \geq 1} \lambda_i^{2L}(\psi_i(x)-\psi_i(y))^2.
$$
Since the eigenvalues in (\ref{diff1}) are less than one, the expansion can be broken off after a finite number of terms $m(\delta,L)$, where $m(\delta,L) = \max\{i \in \mathbb{N}\}, \mbox{ such that } |\lambda_i|^L > \delta|\lambda_1|^L$, where $\delta$ is a measure of the precision desired in this approximation. Each component $\lambda_i^L\psi_i(x), i=1,\ldots,m(\delta,L)$ is termed a diffusion coordinate, and the data are mapped into an Euclidean space of dimension $m(\delta,L)$.

By choosing the kernel $k$ appropriately, various diffusion operators can be obtained. The Laplace-Beltrami operator mentioned in the preceding subsection is one choice. Another one is the Fokker-Planck operator. In addition, each of these operators can be raised to a power $L$ giving rise to a diffusion operator on different scales. We refer to \cite{coif:lafo:2006} for more details.

There are a number of applications of diffusion maps. For an application to gene expression data, see \cite{hagh:buet:thei:2015}.

\subsection{Kernel principal components\label{Kernel principal}}

The standard linear Fisher discriminant seeks to discriminate between two or more populations by using the global Gaussian likelihood ratio method in an attempt to separate  the populations linearly by separating hyper-planes. This is of course not possible for the data in Fig.~\ref{fig:sub1a}. An alternative is to use a local Gaussian Fisher discriminant which leads to nonlinear hyper-surfaces \citep{otne:jull:tjos:2020}. Still another possibility is to use transformations of the original data into nonlinear features and then try to find linear hyper-planes in this feature space. To find the linear hyper-planes scalar products between vectors are used; this being the case both in the linear Fisher discriminant and in case there is a nonlinear feature space. As a function of the original coordinates of observations, the inner product in the feature space is termed a kernel. The support vector machine (SVM) discrimination analysis is based on such an idea.

An analog procedure can be used in so-called kernel PCA 
\citep{scho:smol:mull:2005}. 
Consider a set of data vectors $X_1,\ldots, X_n$ with $X_i \in \mathbb{R}^p$ that sums to the zero-vector. Recall that in ordinary principal components analysis the estimated principal components are found by solving the eigenvalue problem ${\bf C}f = \lambda f$, where, ${\bf C}$ is the empirical $p \times p$ covariance matrix given by
$$
{\bf C} = \frac{1}{n}\sum_{i=1}^{n}X_i X_i^{T},
$$
and corresponding to the matrix ${\bf X}^{T}{\bf X}$ in Section \ref{Principal}. In kernel PCA the starting point is to map the data vector $X_i$ into a nonlinear feature vector $\Phi(X_i)$, $\Phi: \mathbb{R}^p \to F$, where $F$ is an inner product space in general different from $\mathbb{R}^p$, such that $\sum_{i=1}^{n} \Phi(X_i) = 0$.

Consider the $n \times n$ matrix ${\bf K}_{\Phi} = \{\langle \Phi(X_i),\Phi(X_j) \rangle \}$ and the eigenvalue problem
\begin{equation}
{\bf K}_{\Phi} \alpha = n \lambda \alpha,
\label{25e}
\end{equation}
where $\alpha$ is the column vector with entries $\alpha_i,\ldots,\alpha_n$. Let $f^{l}$ be the $l$th eigenvector corresponding to non-zero eigenvalues. 
It can be shown that \citep{scho:smol:mull:2005}  for principal components extraction, one can compute the projections of the image of a data point $X$ onto the eigenvectors $f^{l}$ according to
\begin{equation}
\langle f^{l}, \Phi(X) \rangle = \sum_{i=1}^{n} \alpha_i^{l} \langle \Phi(X_i), \Phi(X) \rangle. 
\label{25g}
\end{equation}
It is very important to observe that neither (\ref{25e}) nor
(\ref{25g}) requires the $\Phi(X_i)$ in explicit form. One just needs
to know the values $k(X,Y) \doteq \langle \Phi(X), \Phi(Y) \rangle$ of
their inner product. The function $k(X,Y)$ is the kernel and using it
instead of explicit values of $\Phi(X)$ and $\Phi(Y)$ is the content
of the so-called {\it kernel trick} (\cite{aize:brav:rozo:1964},
\cite{bose:guyo:vapn:1992}). The point is that one can start with a
suitable kernel instead of having to do the mapping $\Phi(X)$. It can
be shown by methods of functional analysis that there exists for any
positive definite kernel $k$, a map $\Phi$ into some inner product
space $F$, such that $k$ constitutes the inner product of this
space. This space would in general be of infinite dimension (function space), so there it is the opposite of dimensionality reduction. To show that this works and to put this into a rigorous mathematical context,
one uses the framework and the properties of a reproducing
kernel Hilbert space (RKHS). A recent tutorial is given in \cite{gret:2019} Some common choices of kernels include the
polynomial kernel $k(X,Y) = (\langle X, Y \rangle)^d$, for some
integer $d$ and inner product $\langle \cdot, \cdot \rangle$ in
$\mathbb{R}^p$, and the radial basis functions $k(X,Y) = \exp(-||X-Y||^2/2\sigma^2)$. The latter  should be compared to the heat kernel weighting function of Laplacians as surveyed in the previous subsection. Finally there are the sigmoid kernels $k(X,Y) = \mbox{tanh}(\kappa\langle X, Y \rangle + \theta)$ for some tuning parameters $\kappa$ and $\theta$.

Substituting kernel functions for  $\langle \Phi(X), \Phi(Y) \rangle$
one obtains the following algorithm for kernel PCA: One computes the
dot product matrix
$${\bf K}_{\Phi} = \langle \Phi(X_i), \Phi(X_j) \rangle =
k(X_i,X_j),
$$
solve the eigenvalue problem for   ${\bf K}_{\Phi}$, normalize the eigenvector  expansion coefficient $\alpha^{k}$, and extract principal components (corresponding to the kernel $k$, of which there are several choices) of an observational point $X$ by computing projections on the eigenvectors as in Equation (\ref{25g}). Kernel PCA has the advantage that no nonlinear optimization is involved; one only has to solve an eigenvalue problem as in the case of the standard PCA with a feature space that is fixed a priori by choosing a kernel function. The general question of choosing an optimal kernel for a given problem is unsolved both for kernel PCA and SVM, but the three kernels mentioned above generally perform well, \citep{scho:smol:mull:2005}. A connection between kernel principal components and kernels used in diffusion maps is pointed out in Section 2.7 of \cite{coif:lafo:2006}. 

The results of using the kernel principal component method on the data
in Fig.~\ref{fig:sub1a} can be seen in Fig.~\ref{fig:sub1e}. It is seen that the curves are clearly separated along the second kernel principal component. The two dents in the two innermost curves of Fig.~\ref{fig:sub1a} are also reproduced. 

\begin{figure}[ht!]
  \begin{center}
    \begin{subfigure}{.45\textwidth}
      \centering
       \includegraphics[width=1\textwidth]{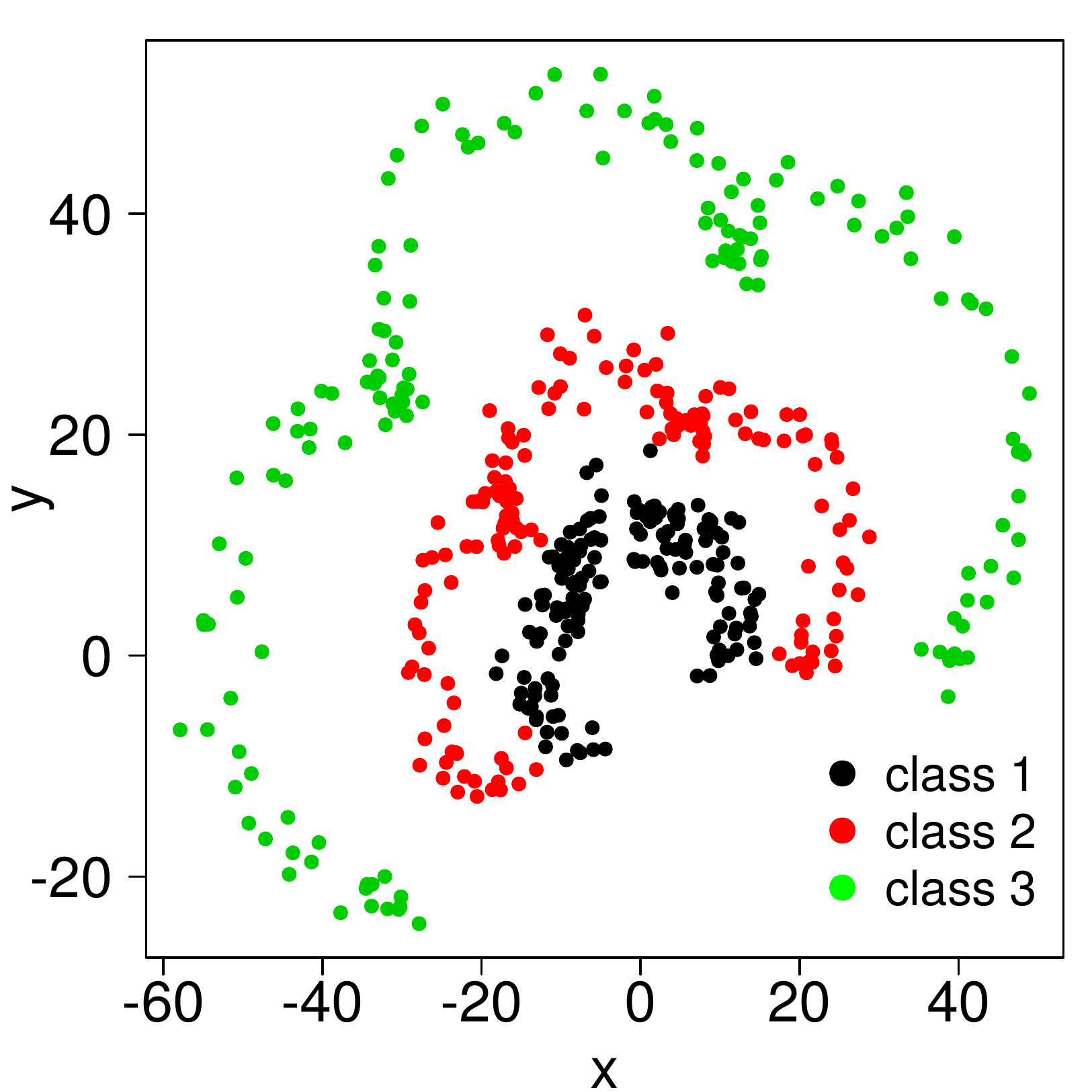}  
      \caption{The raw data.}
      \label{fig:sub2a}
    \end{subfigure}
    \begin{subfigure}{.45\textwidth}
      \centering
       \includegraphics[width=1\textwidth]{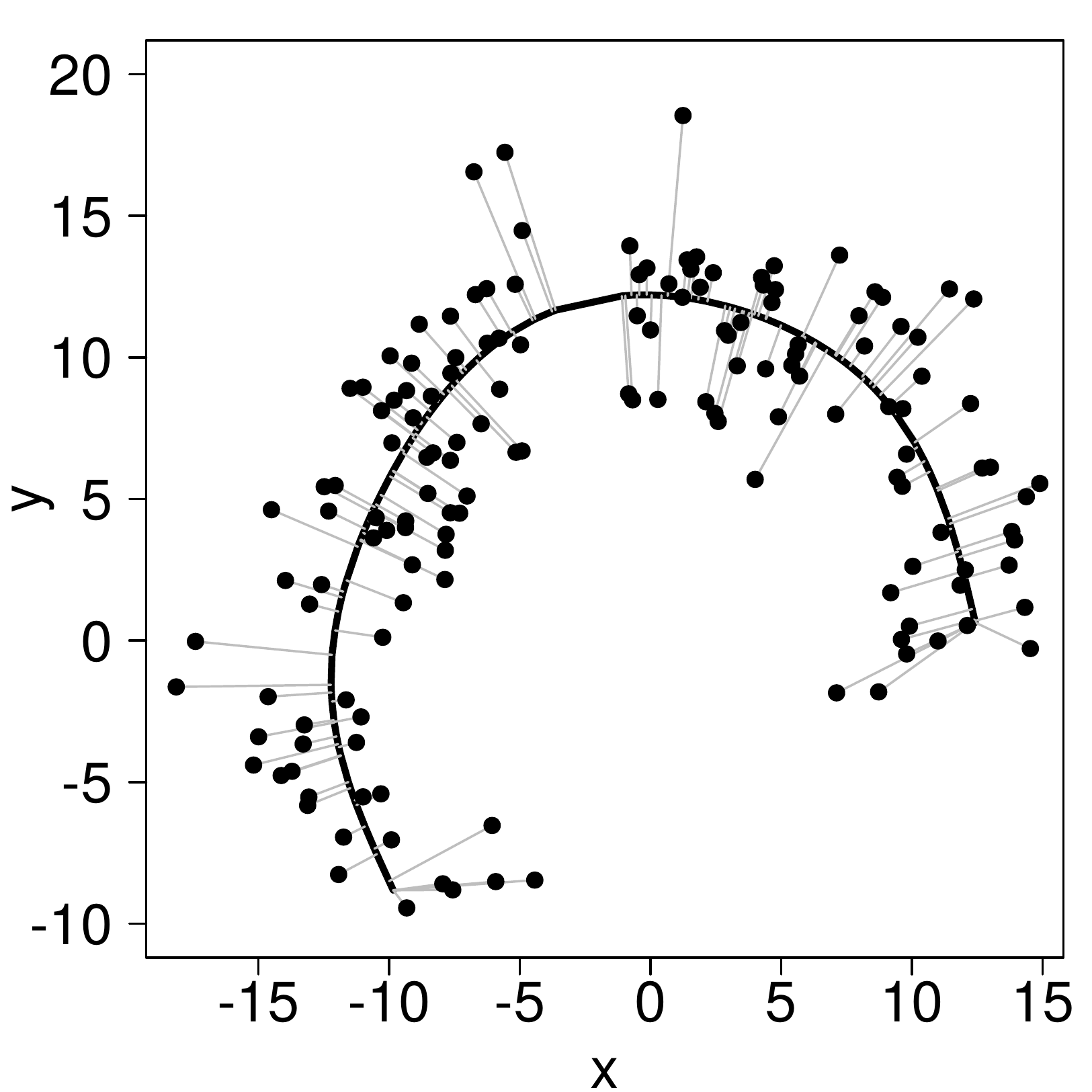}  
      \caption{Principal curve on the innermost curve of Fig.~\ref{fig:sub2a}.}
      \label{fig:sub2b}
    \end{subfigure}
    \begin{subfigure}{.45\textwidth}
      \centering
       \includegraphics[width=1\textwidth]{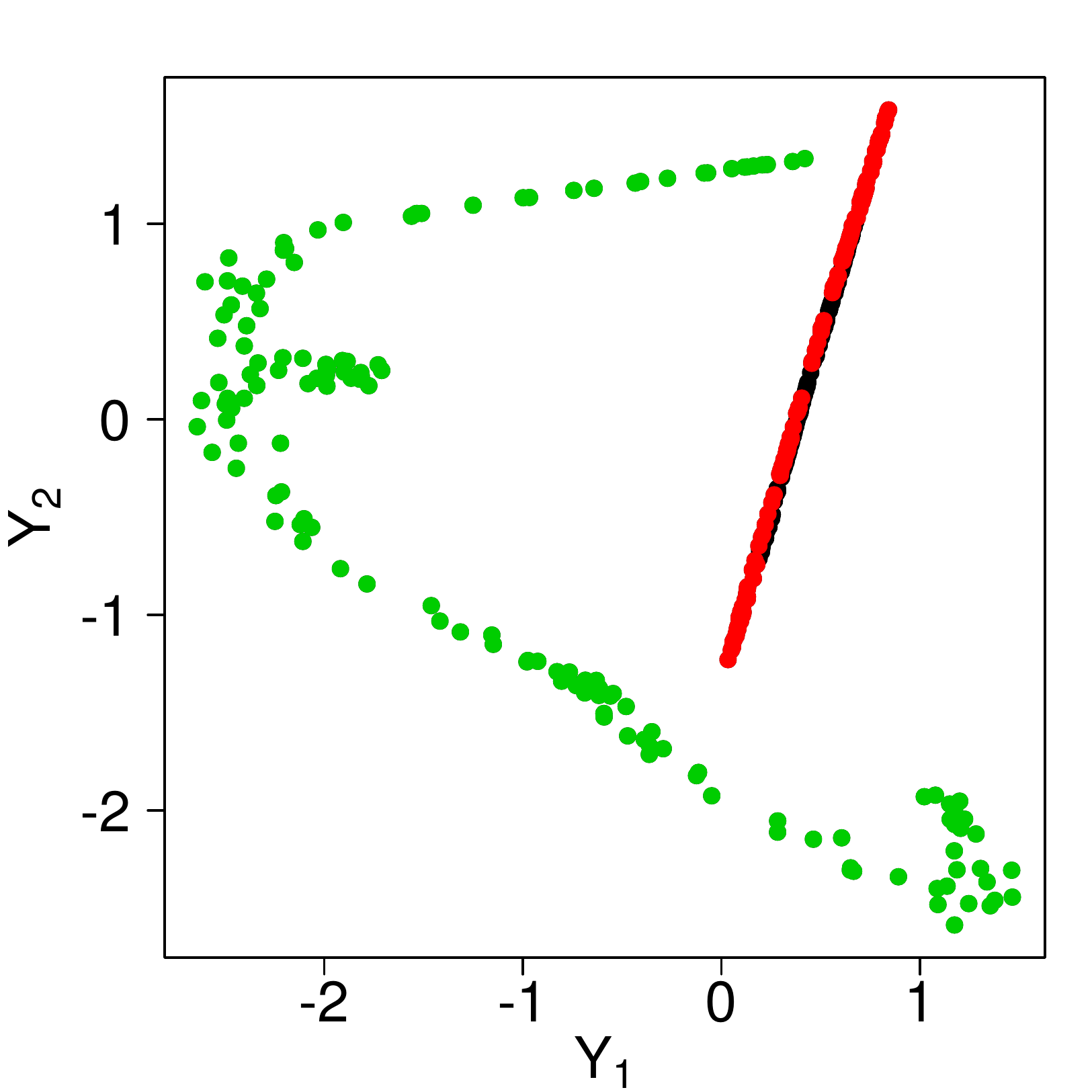}  
      \caption{Local linear embedding.}
      \label{fig:sub2c}
    \end{subfigure}
    \begin{subfigure}{.45\textwidth}
      \centering
       \includegraphics[width=1\textwidth]{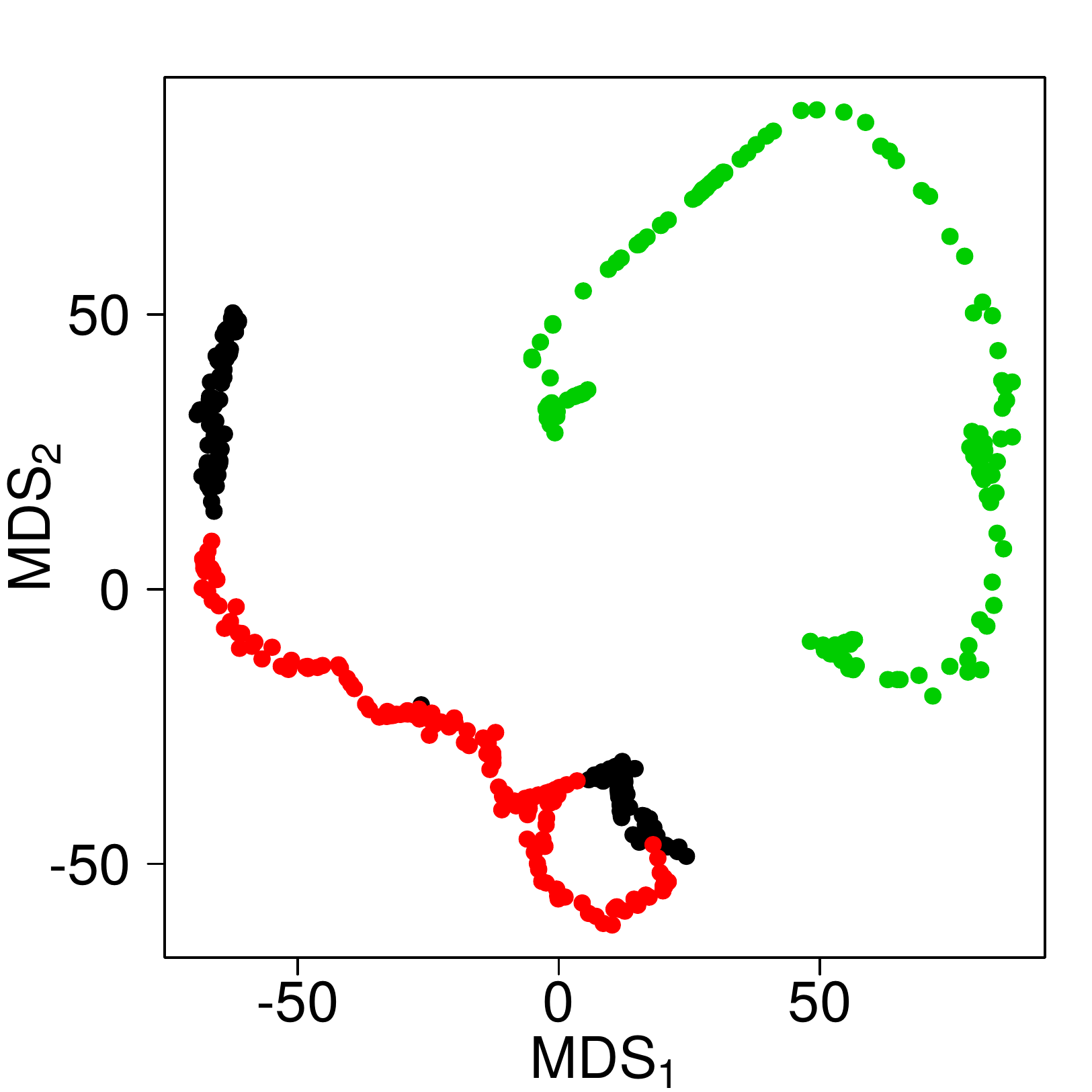}  
      \caption{ISOMAP.}
      \label{fig:sub2d}
    \end{subfigure}
    \begin{subfigure}{.45\textwidth}
      \centering
       \includegraphics[width=1\textwidth]{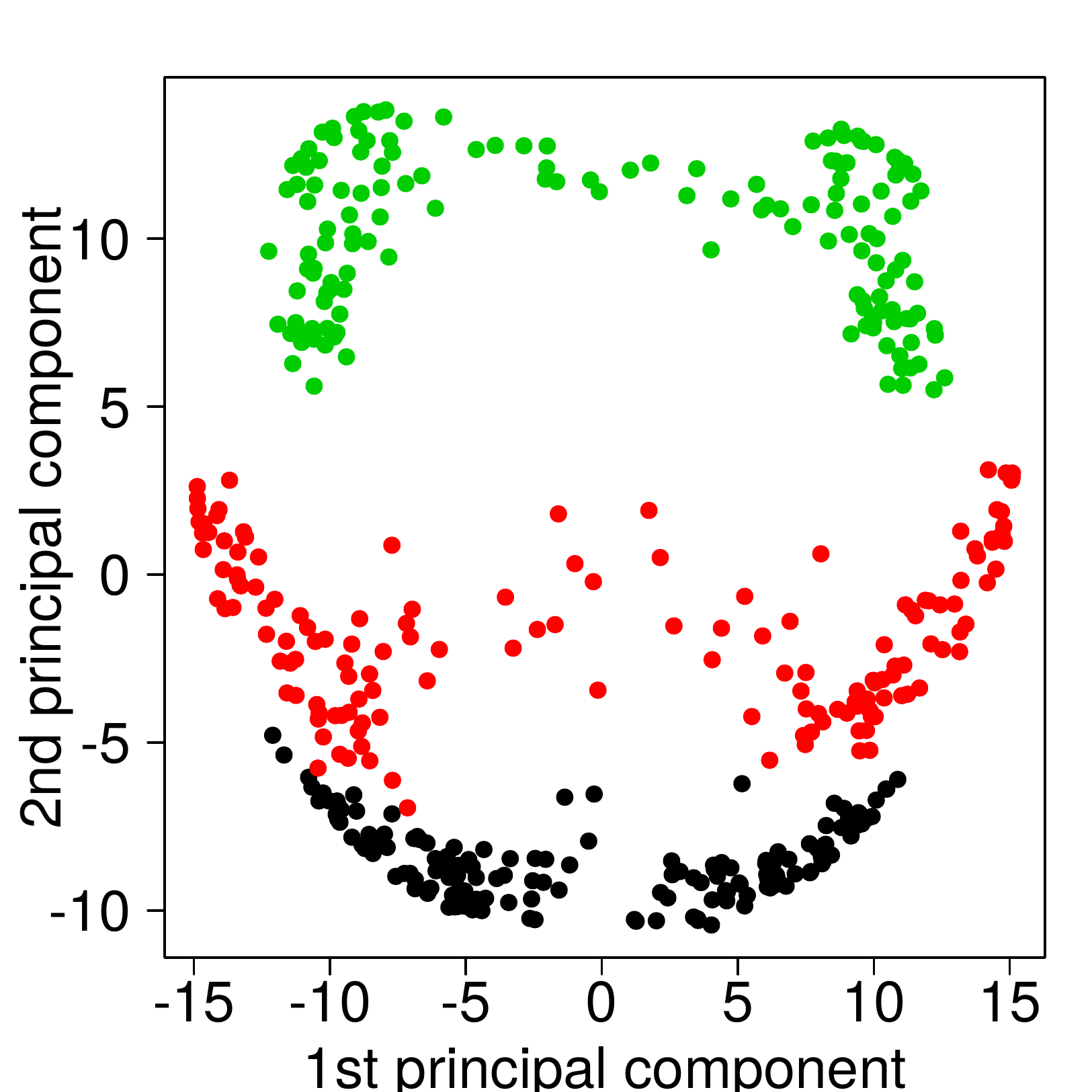}  
      \caption{Kernel principal components (with a Bessel kernel).}
      \label{fig:sub2e}
    \end{subfigure}
\end{center}
\caption{Four different embedding methods applied to three parametric
  curves from the so-called Ranunculoid and perturbed by more Gaussian  noise than in Fig.~\ref{fig:sub1a} (we have used a standard deviation of 2 instead of \sfrac{1}{2} for the noise here).
 \label{fig:2}}
\end{figure}

It is of interest to look at the curves in Fig.~\ref{fig:sub1a} and their nonlinear
representations when the noise is increased. This is done in Fig.~\ref{fig:2}. In
Fig.~\ref{fig:sub2a} it is seen that with the increased noise the two innermost
curves are not separated any more, but rather forms a quite
complicated closed curve. The principal curve for the innermost curve
(with the other two removed) is seen in Fig.~\ref{fig:sub2b}. The overlap of the two
innermost curves is clearly seen for the local linear embedding, the
ISOMAP and the kernel principal component in Figs.~\ref{fig:sub2a}-\ref{fig:sub2c}. It seems that
only kernel principal component  is close to separating the original
three curves. For the two others the two innermost curves coalesce. In
fact for local linear embedding the innermost curve more or less
degenerates to two points.

The ISOMAP picture is also interesting. The
innermost curve is split into two opposite curves. This is consistent
with the gap in the innermost curve in the middle of it. It is also
worth noting that the loop formed on the left hand side of
the two innermost curves is reproduced at the bottom of the ISOMAP
plot.

\afterpage{\clearpage}

\subsection{Random projection\label{Random projection}}

A number of embedding methods depends on a linear or nonlinear transformation of the data. This is for instance the case for principal components, where the transformation is found by solving an eigenvalue problem involving the data. Similar reasoning in terms of eigenvalues is also done for graph representations and eigenmaps as in Sections \ref{Laplace eigenmaps} and \ref{Diffusion maps}. For all of these embedding methods extensive manipulation of the data is necessary to find a suitable transformation.

To be  more specific, let us return to the principal component method of Section \ref{Principal}. Here there is a $n \times p$ data matrix ${\bf X}$. Estimated principal components $\hat{V}_1, \ldots, \hat{V}_m$ are then found by solving the eigenvalue problem (\ref{2b}). Let us denote by $\hat{{\bf V}}$ the $p \times m$ matrix $\hat{{\bf V}} = [V_1,\ldots,V_m]$ of the first $m$ principal components. Then an embedding to the $m$-dimensional space is essentially done by the transformation $\tilde{{\bf X}} = {\bf X}\hat{{\bf V}}$. For a large $p$ this is burdensome computationally. Similarly, the dimension of the eigenvalue problem may be in the millions for the eigenvalue problem (\ref{34r}) for graph representation, and when cross-validation routines are added for a possible classification problem the amount of computations is prohibitive. As will be seen in Section \ref{Skip-Gram main} this has prompted alternative methods, where the eigenvalue problem is avoided.

There is, however, another and very different way to avoid the high computational cost. This is via the so-called random projection method, whose rationale is based on the famous Johnson-Lindenstrauss lemma, \cite{john:lind:1984} . In a random projection algorithm the transformation matrix $\hat{{\bf V}}$ based on the data is simply replaced  by a matrix ${\bf U}$ such that $\tilde{{\bf X}} = {\bf X}{\bf U}$, where each element of the matrix ${\bf U}$ is obtained by drawings from a random variable.
In a normal random projection, cf. \citet[Section 2.1]{li:hast:chur:2007}, the elements $U_{ij}$ are all sampled iid from a standard normal $U_{ij} \sim N(0,1)$.
This certainly implies an enormous saving of computational cost, but one may ask whether it makes sense. After all, the matrix ${\bf U}$ is drawn independently of the data ${\bf X}$.

Here the Johnson-Lindenstrauss lemma comes to ones help. This says that under relatively mild conditions distance relationships are kept approximately invariant under the random projection. There are many formulations of this lemma. We state the one used in \citet[Lemma 2]{li:hast:chur:2007}: If $m > G (2 \log n - \log \delta)/\varepsilon^2$, where $G = 4/(1-2 \varepsilon/3 )$, then with probability at least $1-\delta$, and remarkably, independent of ${\bf X}$ and $p$, the squared $l_2$ distance between any pair of projected data points can be approximated within a factor of $(1\pm \varepsilon)$, $(0 < \varepsilon < 1)$, of the squared $l_2$ distance of the original data after normal random projections. Alternative formulations and proofs can be found in e.g. \cite{ghoj:ghod:karr:crow:2021}. It could be noted that \cite{li:hast:chur:2007} also considers drawing from the Cauchy distribution and using the $l_1$ distance, that may be more robust.

Several attempts have been made to apply the random projections to clustering, classifications and regression. Perhaps not unexpectedly, it has been found that random projections may fail exactly because the transformation ${\bf U}$ is constructed without taking the intrinsic structure of the original data into account. This issue has been sought avoided in various ways for example by considering several random projections in combination with much used classification algorithms. We refer to \cite{cann:samw:2017} and \cite{xie:li:xue:2018} for further reading.

\subsection{A few other techniques\label{Other}}

There are several other alternative methods in nonlinear dimension reduction.
Perhaps the most used one is Independent Components Analysis (ICA). The main concepts of the method are described in a much cited paper  by \cite{hyva:oja:2000}.

In traditional factor analysis latent factors are obtained as latent variables in an eigenvector PCA decomposition. The latent factors are uncorrelated but not unique, since the uncorrelatedness is preserved by orthogonal matrix transformations of the factors, the so-called factor rotation, and in psychometry various factor rotations, varimax and quartimax, see \cite{joli:2002}, have been given special interpretations. This non-uniqueness is intrinsically linked to the Gaussian distribution where independence is equivalent to uncorrelatedness.

In ICA the aim is again to obtain latent factors, and in format the decomposition is the same as the PCA decomposition except that the components are now required to be independent. This means that not only the second order crossmoment (covariance) is assumed to vanish but also all higher order crossmoments. This results in uniqueness. The derivation of the decomposition is done using entropy concepts such as the mutual information, and the Kullback-Leibler distance between probability densities. One might remark that ICA essentially starts from a factor analysis solution to dimension reduction and looks for rotations that lead to independent components. From this point of view ICA is just another factor rotation along with the traditional varimax and quartimax.

Two other methods will be very briefly mentioned. These are both neural network based methods. One of them consists in so-called autoencoding in deep neural networks, and can be represented by \cite{hint:sala:2006}. The other is the method of Self Organizing Maps which can be said to have originated by another much cited paper, \cite{koho:1982}.
The latter is also covered in \citet[chapter 14.4]{hast:tibs:frie:2009}.


\section{Topological embeddings and topological data analysis (TDA) \label{Topological}}

The present section concerns topological embeddings and data analysis. 
We will divide our exposition in three parts, manifold learning,
persistent homology, and finally the Mapper algorithm. It is the persistent homology part that is usually identified
with TDA. Our point of departure is in all cases a point cloud in
$\mathbb{R}^{p}$. In part one the objective is to examine whether
there is a possibility of embedding the point cloud in a lower
dimensional manifold. In the two latter parts the aim is to try to find
additional topological features that may characterize the point cloud
and its embedding. In order to avoid an overlong paper, parts of the
TDA survey have been moved to the Supplement \citep{supp}. Two main introductory references to manifold learning and TDA are \cite{wass:2018} and \cite{chaz:mich:2021}.

\subsection{Manifold learning\label{Manifold}}

Already in the \cite{pear:1901} treatment of principal components, 
the point cloud of data is embedded on a hyper-plane in $\mathbb{R}^{p}$. The approach of ISOMAP and local linear embedding are early examples of representing the data in a lower dimensional manifold.

A main aspect of manifold learning is that one looks for a
non-Euclidean subspace to make an embedding that may not easily be
achieved in an Euclidean space $\mathbb{R}^{m}$, but more efficiently
on a manifold. One trivial example is the case where the point cloud
in the plane is concentrated on a circle with only small additional
perturbations. The data can then essentially be reduced from
two-dimensional space (the plane), not to the line
$(\mathbb{R })$, but to the circle which is a one-dimensional
manifold. For a more complex example we refer to the Ranunculoid of
Fig.~\ref{fig:sub1a}. An extension to the perturbed circle example is the Swiss roll as a two-dimensional manifold in $\mathbb{R}^{3}$. 

In the  more general case manifold learning consists in finding a smooth compact submanifold $S$ of $\mathbb{R}^{p}$ on which the point cloud data may be reasonably located. ``Finding'' in general comprise both estimating the dimension of $S$ and estimating $S$ itself. But often the dimension is assumed known.

One may estimate $S$ by trying to cover the data cloud by a collection of balls of radius $\varepsilon$, such that 
\begin{equation}
\hat{S} = \cup_{i=1}^{n}B(X_i,\varepsilon),
\label{41a}
\end{equation}
 where $n$ is the number of observations and $B(X_i,\varepsilon) =
 \{x: ||x-X_i|| \leq \varepsilon\}$, and where $X_i$ is observation
 number $i$ of the point cloud. This was suggested by
 \cite{devr:wise:1980}  in another context. If the observations $X_i$
 are all exactly on $S$ and with $\varepsilon$ depending on $n$, it is
 possible to prove convergence of $\hat{S}$ to $S$ at the rate of
 $O_P(\log n/n)^{1/r}$, where $r$ is the dimension of $S$, and the
 distance between $S$ and $\hat{S}$ is measured in terms of the
 Hausdorff distance between sets. 

It is not likely that a sample will fall precisely on $S$. A more realistic model is that one observes $Y_i = X_i+\delta_i$, where $X_i$ comes from a distribution with support on $S$, and $\delta_i$ are samples from a noise distribution. In this case the convergence rate of the estimation of $S$ is very slow; see 
\cite{geno:pero:verd:wass:2012}. 
An interesting example of two-dimensional data, but where there is a set $S$ of dimension 1 with a high concentration of data, is the data set of galaxies treated in \cite{chen:ho:free:geno:wass:2015a,chen:ho:tenn:mand:crof:2015b}.

As mentioned, in a theoretical analysis, often the dimension  $r$ of the embedding manifold is assumed known. In practice one may need to estimate $r$; see \cite{levi:bick:2005}, \cite{litt:magg:rosa:2011}, and \cite{kim:rina:wass:2019}. 
It may be possible to estimate an $r$-dimensional and high density region $R$ that is close to $S$. One way to make this more precise is through the idea of density ridges. A density ridge is a low-dimensional set with large density. 


The ridge set can then be estimated by the ridge of the kernel density estimator. The properties of this estimator is studied in \cite{geno:pero:verd:wass:2014} and \cite{chen:geno:wass:2015c}. A popular algorithm for finding the ridge set estimator was given by \cite{ozer:erdo:2011}, the so-called SCMS algorithm.  
Recently, \citet{qiao:polo:2021} proposed two novel algorithms for estimating ridge lines in ridge regression. They provide theoretical guaranties for their convergence in probability using the Hausdorff distance between the estimated and theoretical ridge. There are no analog results for the SCMS algorithm, which also, as pointed out in Section 2.3 of \cite{qiao:polo:2021}, may encounter difficulties in the vicinity of saddle points.

\subsection{Persistent homology and persistence diagrams\label{Persistent}}

In our context the concept of homology can be seen as coming from a desire to answer the question of whether two sets are topologically similar. For instance is an estimate $\hat{S}$ of $S$
 topologically similar to $S$, or is it at all possible to find an estimate of $S$ that is topologically similar to $S$? The answer to this question depends on what is meant by ``similar''.

Two sets $S$ and $T$ equipped with topologies are homeomorphic if there exists a bi-continuous map from $S$ to $T$. \cite{mark:1958} proved that, in general, the question of whether two spaces are homeomorphic is undecidable for dimension greater than 4.

However, it is possible to use the weaker notion of homology, and it is much easier to determine whether two spaces are homologically equivalent. Strictly speaking homology is a way of defining topological features algebraically using group theory. See e.g.\ \cite{carl:2009} for a precise definition. Intuitively it means that one can compare connected components, holes and voids for two spaces. The zeroth order homology of a set corresponds to its connected components. The first order homology corresponds to one-dimensional holes (like a donut), whereas the second order homology corresponds to two dimensional holes (like a soccer ball) and so on for higher dimensions. If two sets are homeomorphic, then they are homologically equivalent, but not vice versa.

Homology is a main topic of TDA. To establish a link with the previous subsection, consider the estimate $\hat{S} = \cup_{i=1}^{n}B(X_i,\varepsilon)$ of Equation (\ref{41a}). One of the first results about topology and statistics is due to \cite{niyo:smal:wein:2008}. They showed that under certain technical conditions the set $\hat{S}$ has the same homology as $S$ with high probability. 

In many ways topological data analysis has been identified with the subject of persistent homology. This is concerned with the homological structure of data clouds at various scales of the data, and to see how the homology changes (how persistent it is) over these various scales, cf. also Section \ref{Diffusion maps}. Two main introductory sources are \cite{wass:2018} and \cite{chaz:mich:2021}.

The field of TDA is new. It has emerged from research in applied topology and computational geometry initiated in the first decade of this century. Pioneering works are \cite{edel:letc:zomo:2002} and \cite{zomo:carl:2005}. An early survey paper at a relatively advanced mathematical level but with a number of interesting and illustrative examples is \cite{carl:2009}. \cite{wass:2018} and \cite{chaz:mich:2017} are somewhat less technical and more oriented towards statistics. See also \cite{ghri:2017}.

For our purposes of statistical embedding, TDA brings in some new aspects in that topological properties are emphasized in the embedding. This is done to start with in so-called persistence diagrams which depict the persistence, or lack thereof, of certain topological features as the scale in describing a data cloud changes. In complicated situations persistence diagrams can be computed from simplical complexes. This is a particularly interesting concept since it generalizes the embedding of a point cloud in  a graph. A one dimensional simplical complex can be identified with a graph, whereas generalizations allow for describing cycles and voids of the data. This is of special interest for certain types of data, such as porous media and physiological or cell data.

To introduce the persistence diagram, recall the estimator $\hat{S}$ in (\ref{41a}) as a union of balls $B(X_i,\varepsilon)$ of radius $\varepsilon$. One may question what happens to this set as the radius of the balls increases. Consider for example a data cloud that contains a number $n$ of isolated points that resembles a circular structure.  Let each point be surrounded by a neighborhood consisting of a ball centered at each data point and having radius $\varepsilon$. Then initially and for a small enough radius $\varepsilon$, the set $\cup_{i=1}^{n} B(X_i,\varepsilon)$ will consists of $n$ distinct connected sets (homology zero). But as the radius of the points increases, some of the balls will have non-zero intersection, and the number of connected sets will decrease. For $\varepsilon$ big enough one can easily imagine that the set $\cup_{i=1}^{n} B(X_i,\varepsilon)$ is large enough so that it covers the entire circular structure obtaining an annulus-like structure of homology 1, but such that there still may exist isolated connected sets (of homology 0) apart from the annulus. Continuing to increase the radius, one will eventually end up with one connected set of zero homology. 

This process, then, involves a series of births (at $\varepsilon$-radius zero $n$ sets are born) and deaths of sets as the isolated sets coalesce. 
A  useful plot is the  persistence diagram, which has the time (radius) of birth on the horizontal axis and the time (radius) of death on the vertical axis. 
The birth and death of each feature is represented by a point in the diagram. All points will be above or on the diagonal then. For the circle example mentioned above the birth and death of the hole will be well above the diagonal, and it has a time of death which may be considerably larger than its time of birth. The birth and death points of the connected components on the other hand may be quite close to the diagonal if the distances between points are small enough. 

\begin{figure}[ht!]
  \begin{center}
    \begin{subfigure}[b]{.32\textwidth}
      \centering
      \includegraphics[width=1\textwidth]{example1a}  
      \caption{The raw data.}
      \label{fig:sub3a}
    \end{subfigure}
    \begin{subfigure}[b]{.32\textwidth}
      \centering
      \includegraphics[width=1\textwidth]{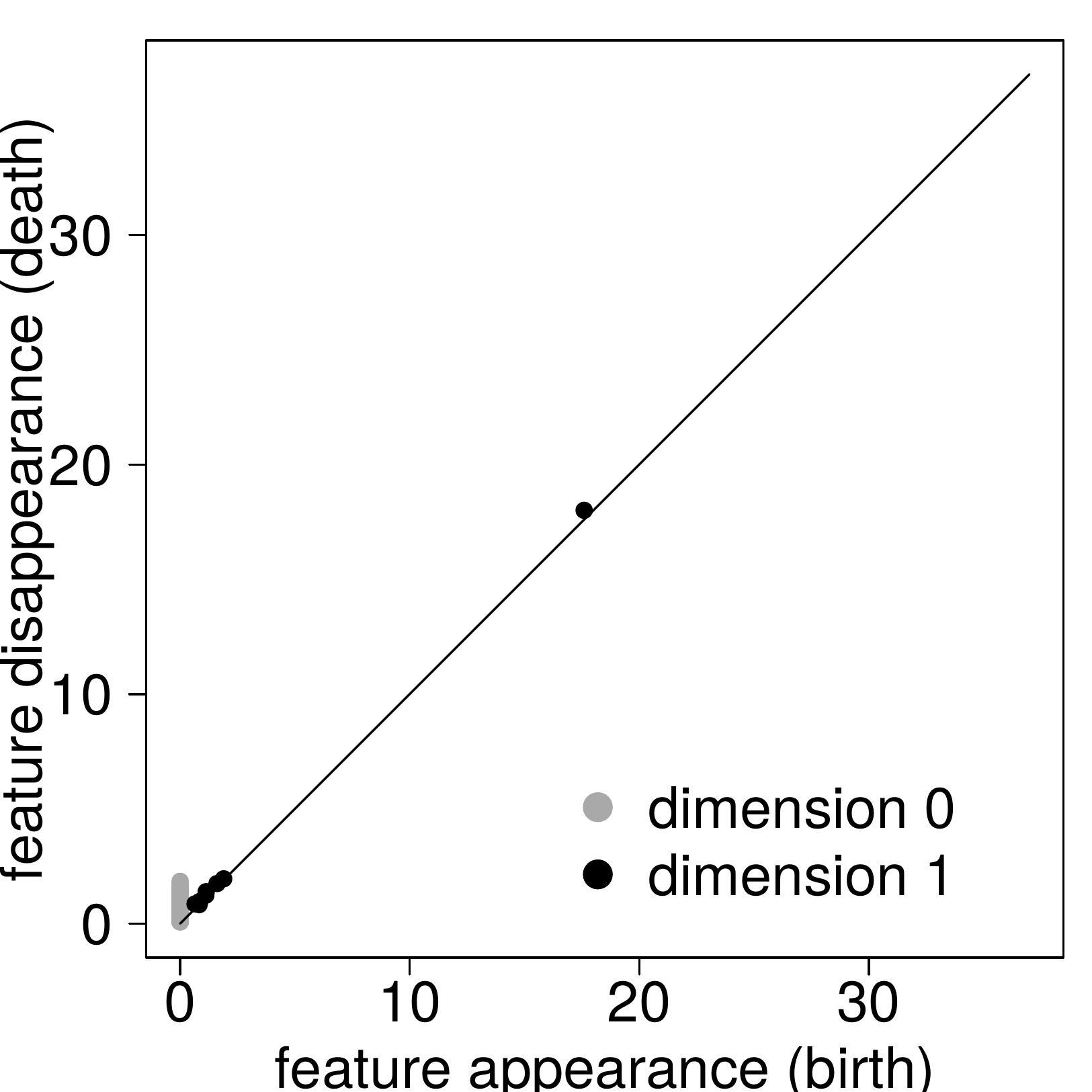}  
      \caption{Class 1.}
      \label{fig:sub3b}
    \end{subfigure}
    \begin{subfigure}[b]{.32\textwidth}
      \centering
      \includegraphics[width=1\textwidth]{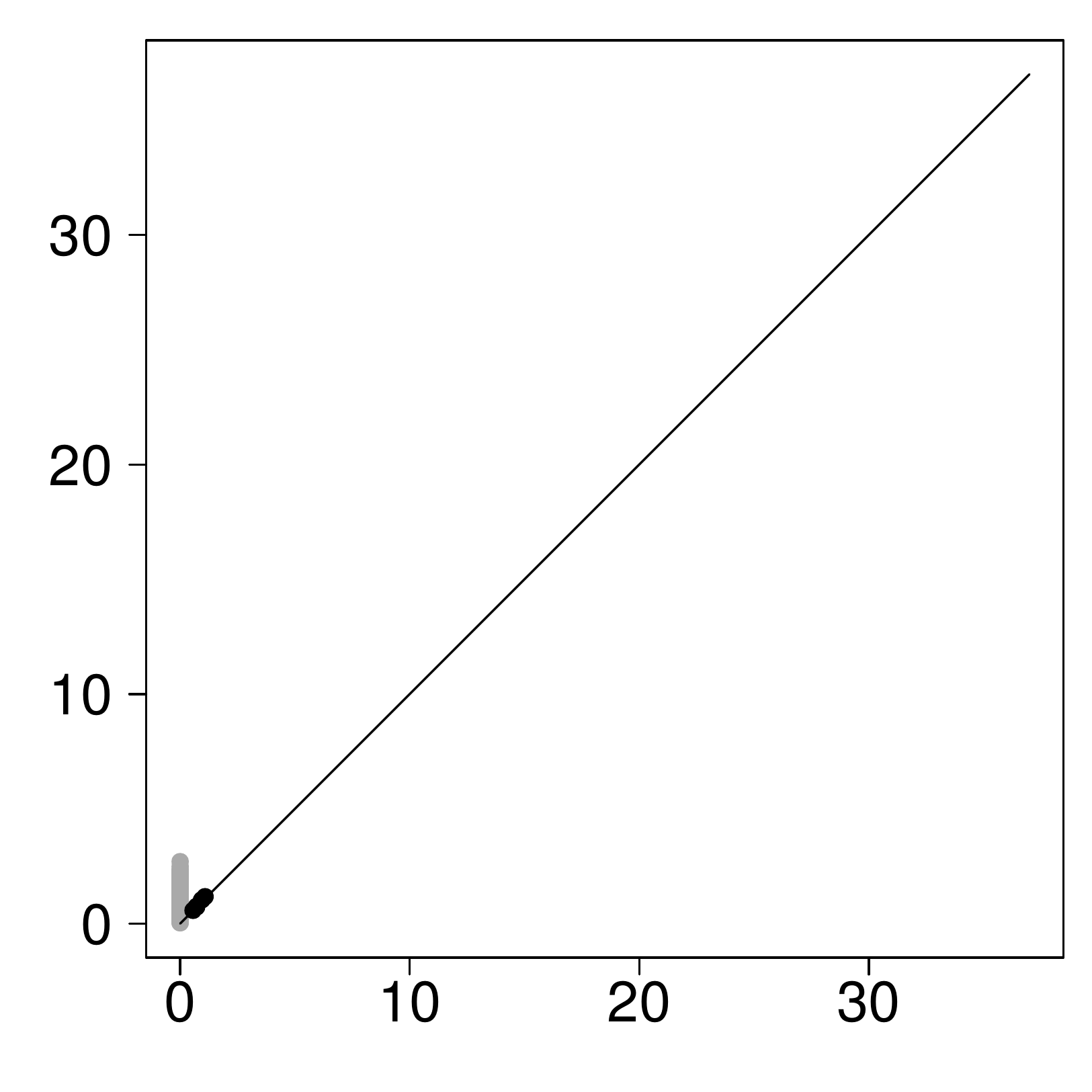}  
      \caption{Class 2.}
      \label{fig:sub3c}
    \end{subfigure}
    \begin{subfigure}[b]{.32\textwidth}
      \centering
      \includegraphics[width=1\textwidth]{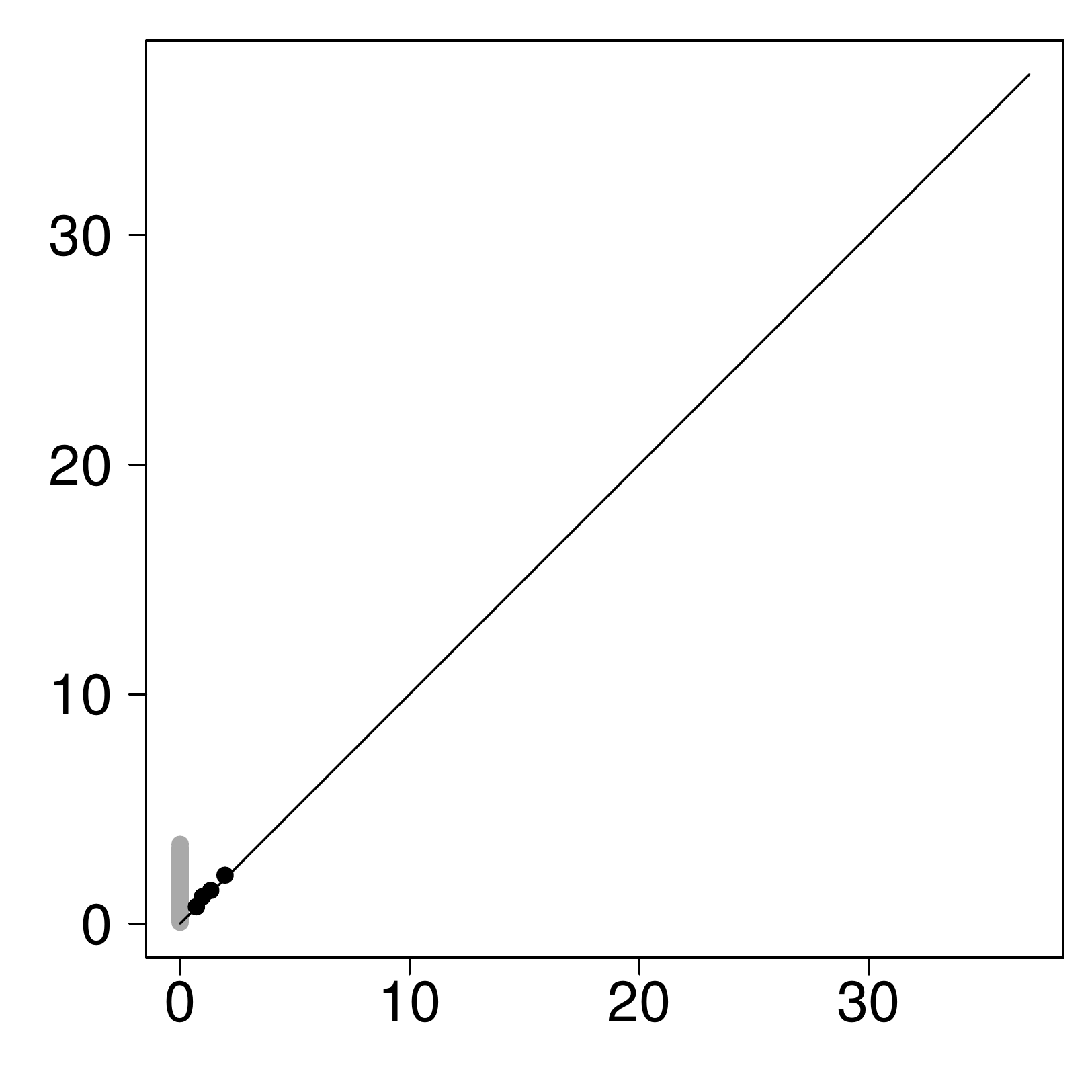}  
      \caption{Class 3.}
      \label{fig:sub3d}
    \end{subfigure}
    \begin{subfigure}[b]{.32\textwidth}
      \centering
      \includegraphics[width=1\textwidth]{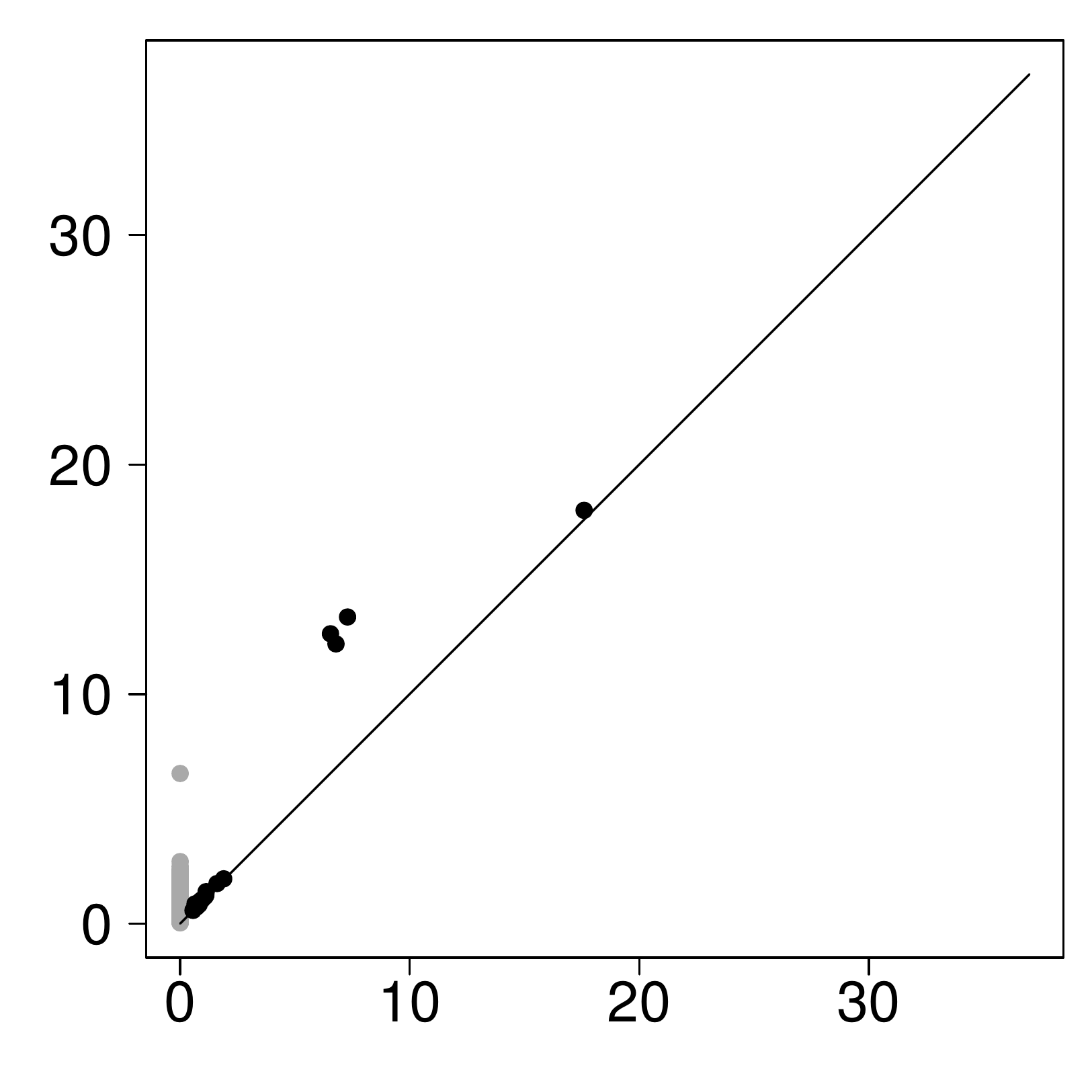}  
      \caption{Classes (1,2).}
      \label{fig:sub3e}
    \end{subfigure}
    \begin{subfigure}[b]{.32\textwidth}
      \centering
      \includegraphics[width=1\textwidth]{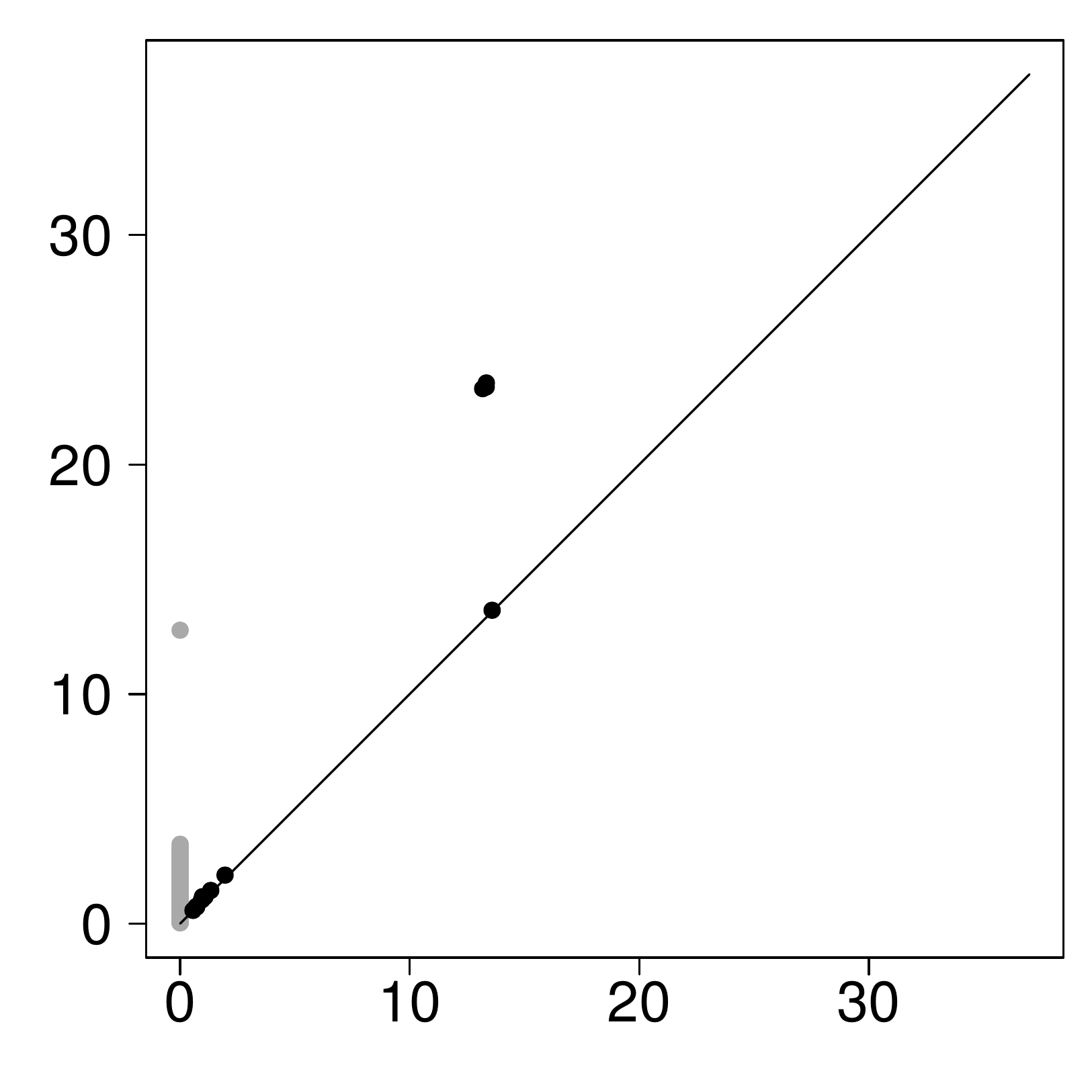}  
      \caption{Classes (2,3).}
      \label{fig:sub3f}
    \end{subfigure}
    \begin{subfigure}[t]{.32\textwidth}
      \centering
      \includegraphics[width=1\textwidth]{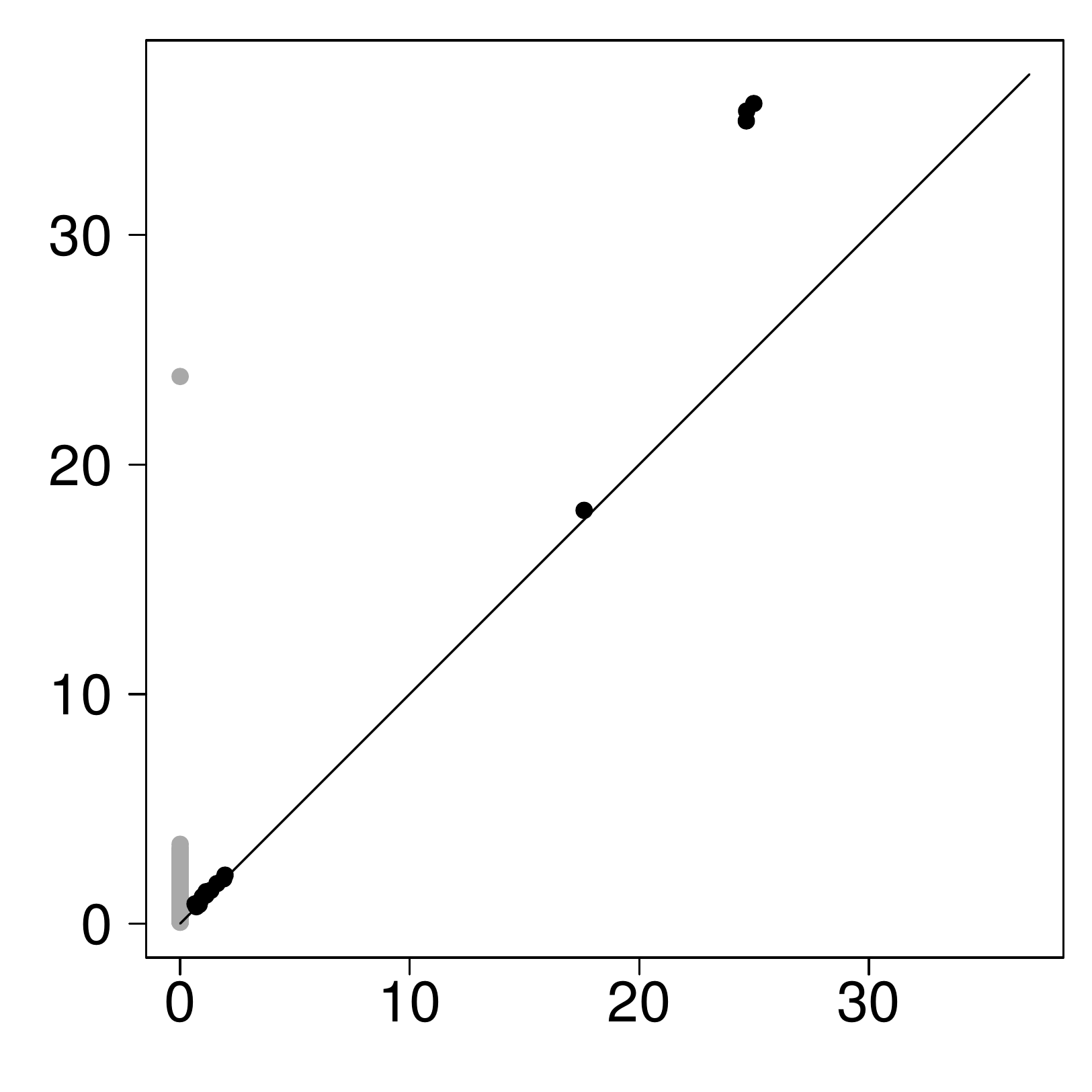}  
      \caption{Classes (1,3).}
      \label{fig:sub3g}
    \end{subfigure}
    \begin{subfigure}[t]{.32\textwidth}
      \centering
      \includegraphics[width=1\textwidth]{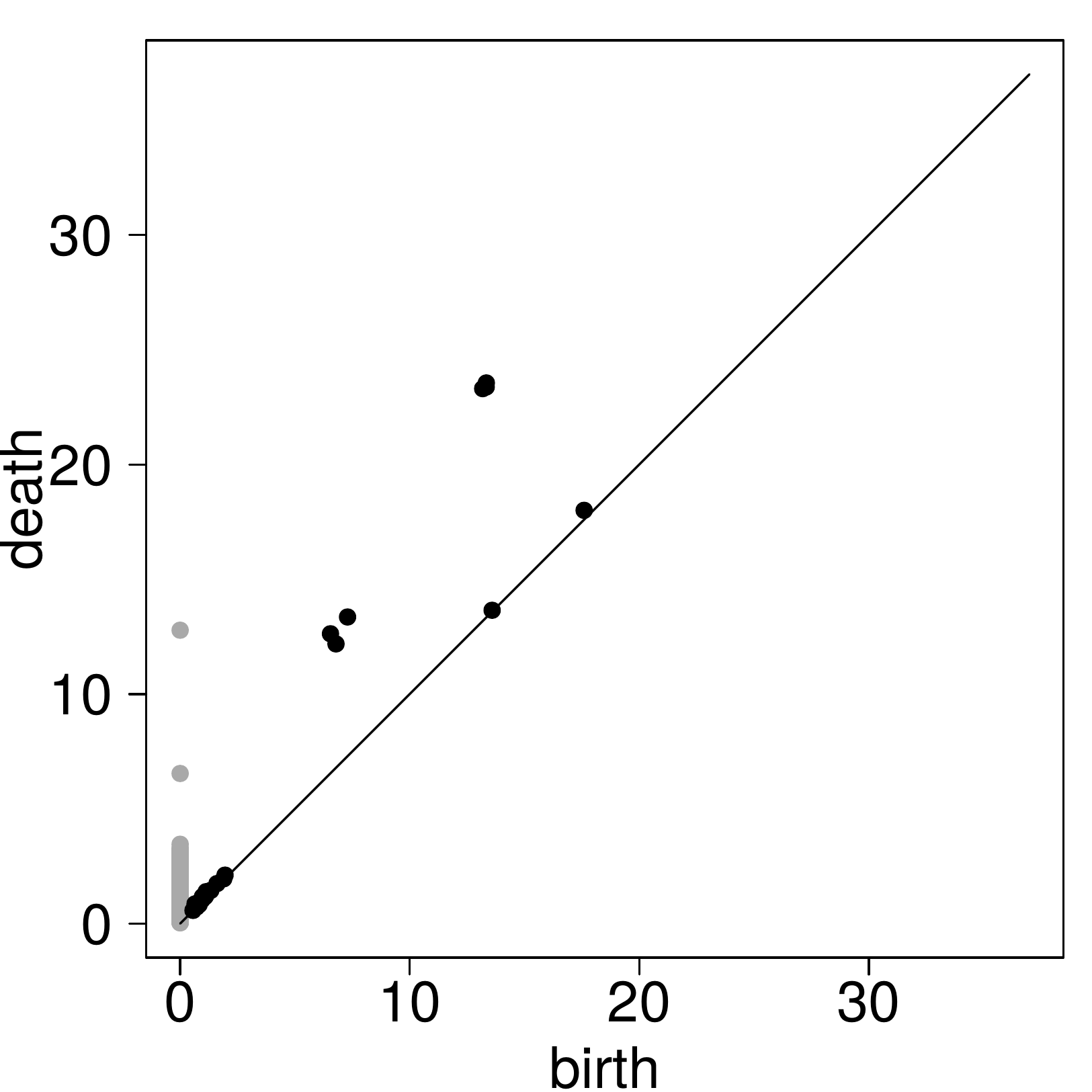}  
      \caption{Classes (1,2,3).}
      \label{fig:sub3h}
    \end{subfigure}
\end{center}
\caption{Persistence diagrams for combinations of classes 1, 2 and 3.
 \label{fig:3}}
\end{figure}

We will go through the steps of this procedure in a rather more
complicated example than the circle, namely that of the noisy
Ranunculoid structure of Fig.~\ref{fig:sub1a}. We will start by
considering each of the three curves, then pair of curves and finally
all three curves. The corresponding persistence diagrams are displayed
in Fig.~\ref{fig:3}, and these diagrams furnish the topological
embedding signature of the data, which is rather different from and
presents additional information compared to the embeddings in Figures
\ref{fig:1} and \ref{fig:2}.  

Consider first the individual curves in Figures
\ref{fig:sub3b}-\ref{fig:sub3d} (and where Fig.~\ref{fig:sub3a} is
identical to Fig.~\ref{fig:sub1a}). Here, class 1, 2 and 3 in Figures
\ref{fig:sub3b}-\ref{fig:sub3d} represent the persistence diagram of
the innermost to the outermost curves, respectively. The gray points
represent sets of homology zero (isolated sets) and black points
represent sets of homology one, i.e., one-dimensional holes. The gray
column at the left is just the time of death for all the sets around
the individual points as the radius for the individual neighborhoods
increase. Naturally the column is highest for the outermost curve in
Fig.~\ref{fig:sub3d}, where the distances between points are
largest. The black points at the right hand side of the columns mark
small holes that temporarily arise in this process due to indents in the
point spreads. Probably these points would not have been there if
circles had been used instead of the Ranunculoid curves. For the
innermost curve there is a black point at the far right with a short
lifetime. This is due to the opening in this curve, which is just great
enough for there to form an annulus as the radii increase. 

Next, to the diagram of the pairwise curves: The pair (1,2) consists
of the two innermost curves, and the persistence diagram is  displayed
in Fig.~\ref{fig:sub3e}. The points of curve 1 can again be
found. In addition at birth time zero, there is a gray point above the
gray column. This is just due to the fact that there are two curves at
the starting point. As time (and radii) increase the two curves
coalesce and we have a death at the gray point above the gray
column. The three black points being born at approximate time 6 and
living for about time 6 to time 12 come from holes that are created as
curve 1 and 2 are approximating each other. The explanation for the
pair (2,3) is much the same. In this case it takes more time before
the curves 2 an 3 coalesce, so the gray point at time zero are farther
up. Here too 3 holes are formed as the curves 2 and 3 approach each
other. One hole has very short lifetime, it is almost on the diagonal,
where as the two others almost coincide and have far longer
lifetime. This has to do with the different levels of indention on the
two curves. Finally, for the pair (1,3), the gray point at zero is
even farther up, reflecting the increased distance between the curves
1 and 3. Again the pattern of curve 1 is dominating as for the pair
(1,2). The indents of curve 1 are small in comparison with the indents
of curve 3, and this explains that it takes longer time for holes to
appear as these two curves are approaching each other. 

The diagram for the triple of curves (1,2,3) in Fig.~\ref{fig:sub3h}
is roughly obtained by superposition of the pattern for the pairwise
curves. There is a difference at birth time zero, though. The
uppermost point for the pair (1,3) has disappeared. The explanation is
obvious. The curves 1 and 2 coalesce first due to least distance
between them. Curve 3 is then coalescing with the set combined curve 1
and 2, which has a distance from curve 3 equal to the distance between
2 and 3, such that the second gray point at zero correspond to the
gray point at zero for the pair (2,3). 

\begin{figure}[ht!]
  \begin{center}
    \begin{subfigure}[b]{.32\textwidth}
      \centering
      \includegraphics[width=1\textwidth]{example2a}  
      \caption{The raw data.}
      \label{fig:sub4a}
    \end{subfigure}
    \begin{subfigure}[b]{.32\textwidth}
      \centering
      \includegraphics[width=1\textwidth]{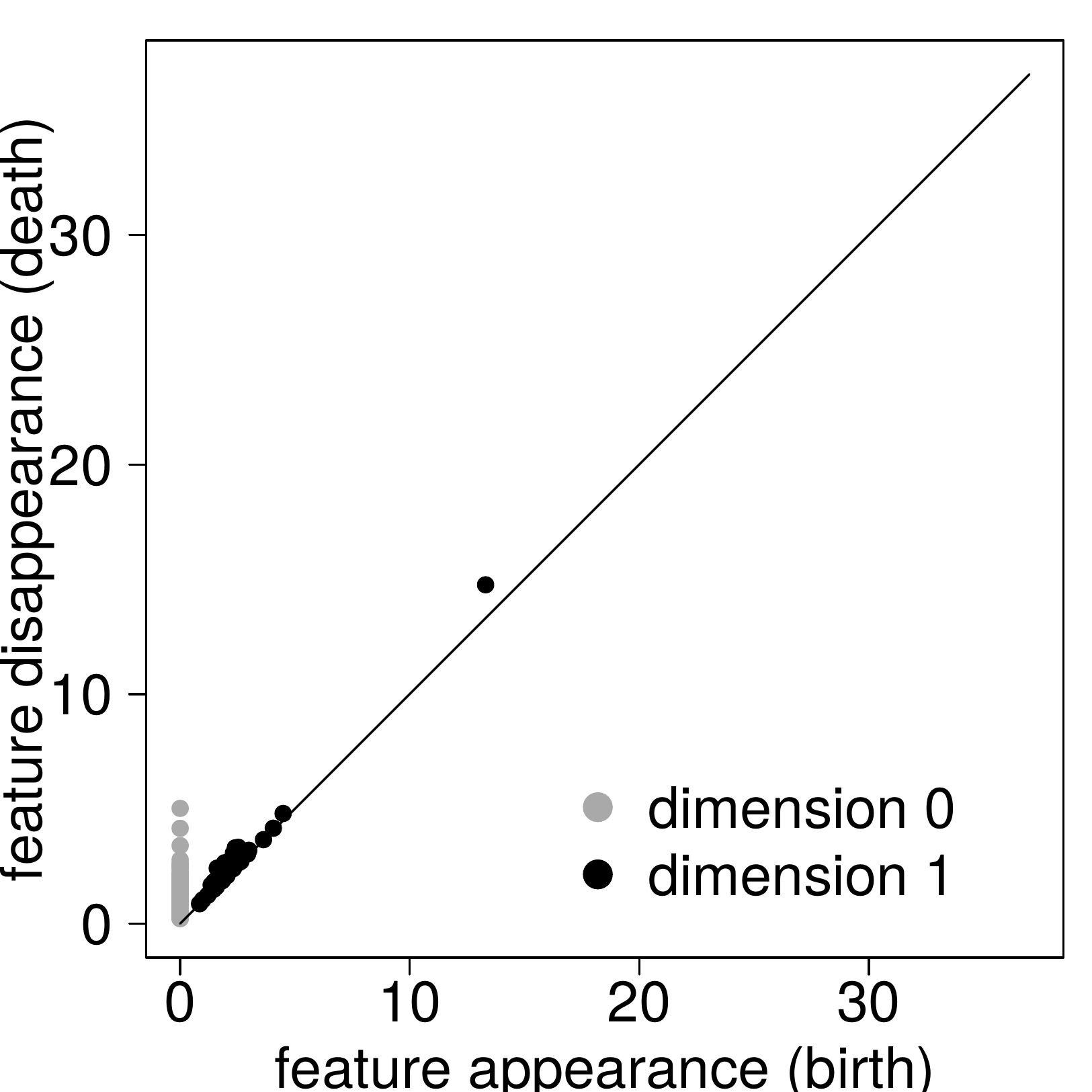}  
      \caption{Class 1.}
      \label{fig:sub4b}
    \end{subfigure}
    \begin{subfigure}[b]{.32\textwidth}
      \centering
      \includegraphics[width=1\textwidth]{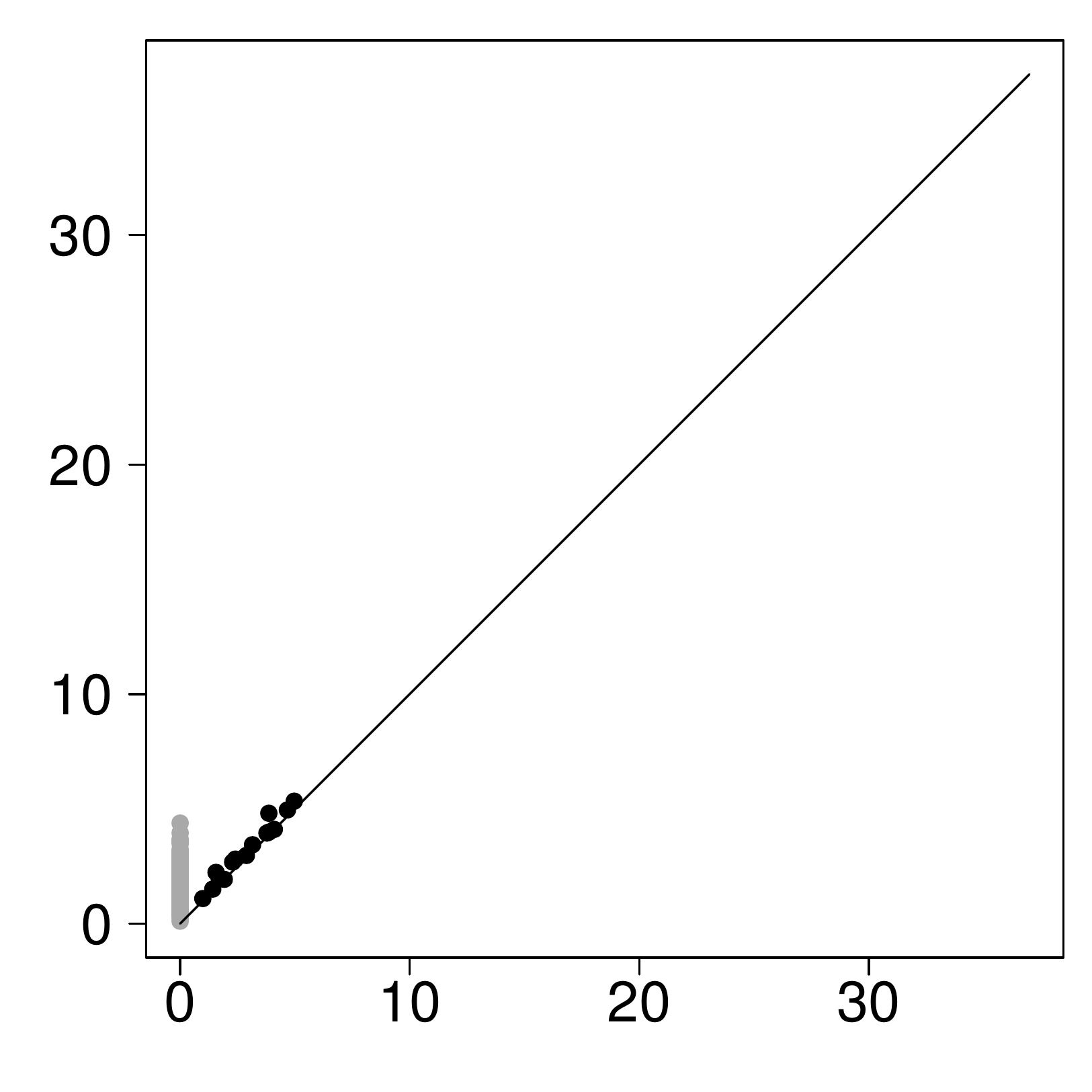}  
      \caption{Class 2.}
      \label{fig:sub4c}
    \end{subfigure}
    \begin{subfigure}[b]{.32\textwidth}
      \centering
      \includegraphics[width=1\textwidth]{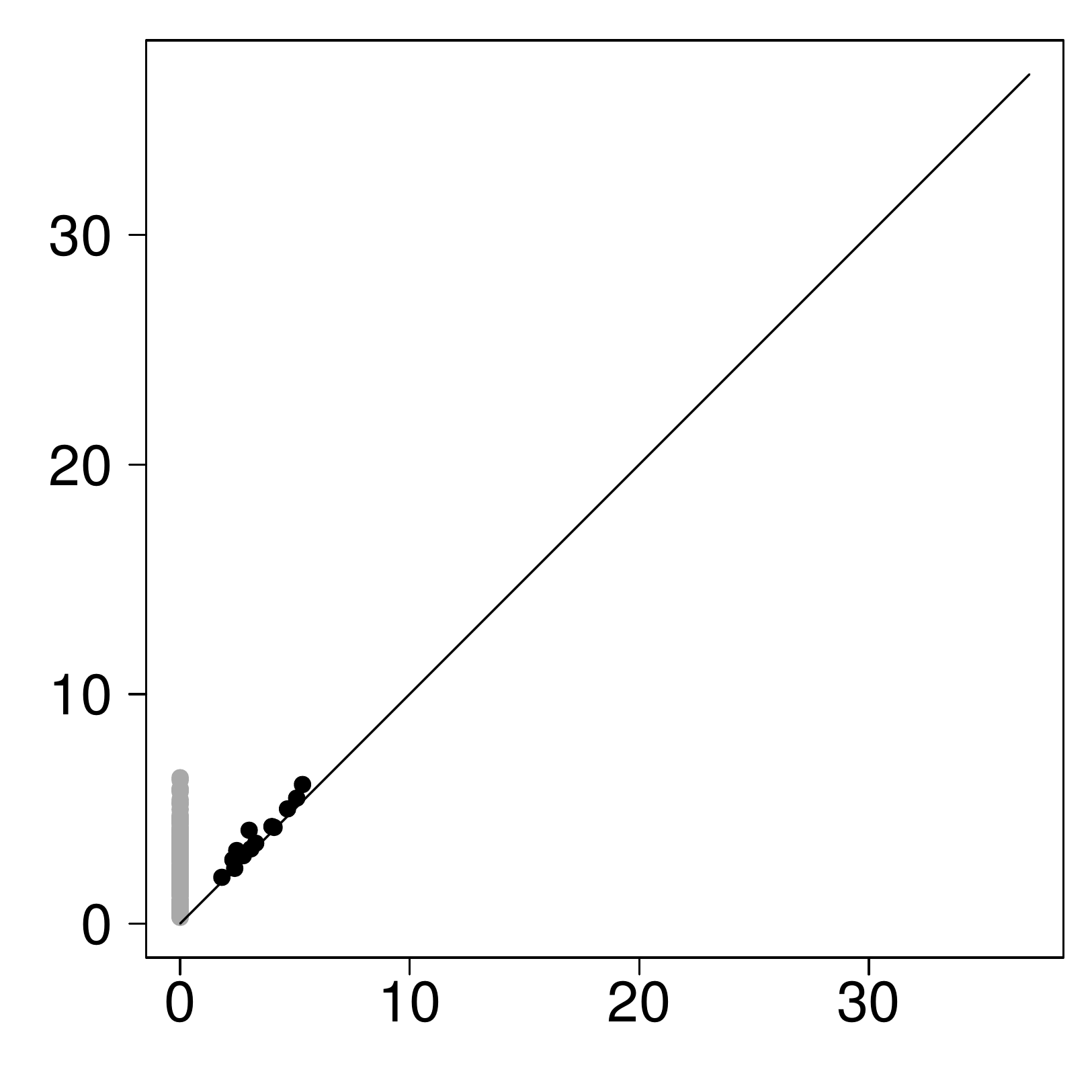}  
      \caption{Class 3.}
      \label{fig:sub4d}
    \end{subfigure}
    \begin{subfigure}[b]{.32\textwidth}
      \centering
      \includegraphics[width=1\textwidth]{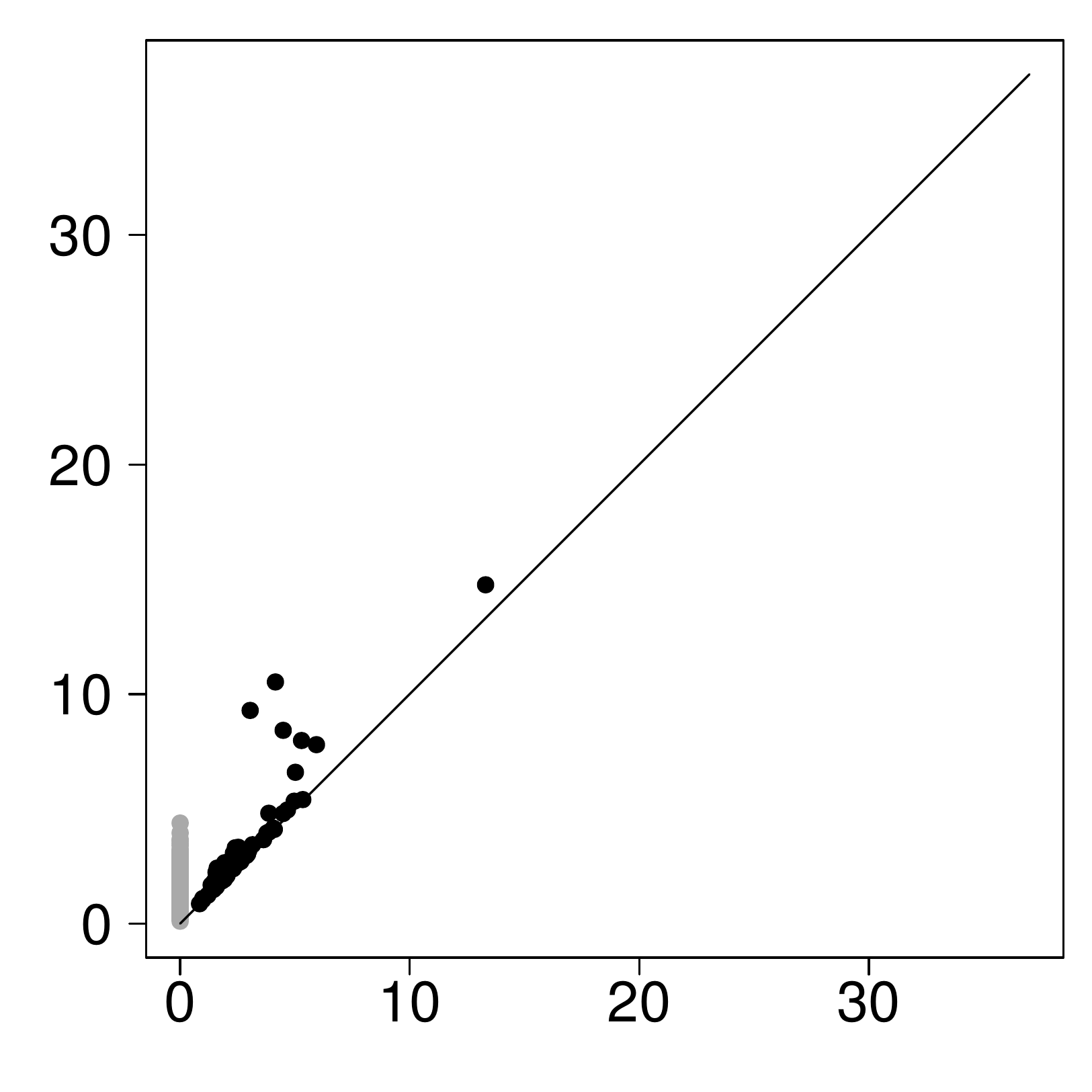}  
      \caption{Classes (1,2).}
      \label{fig:sub4e}
    \end{subfigure}
    \begin{subfigure}[b]{.32\textwidth}
      \centering
      \includegraphics[width=1\textwidth]{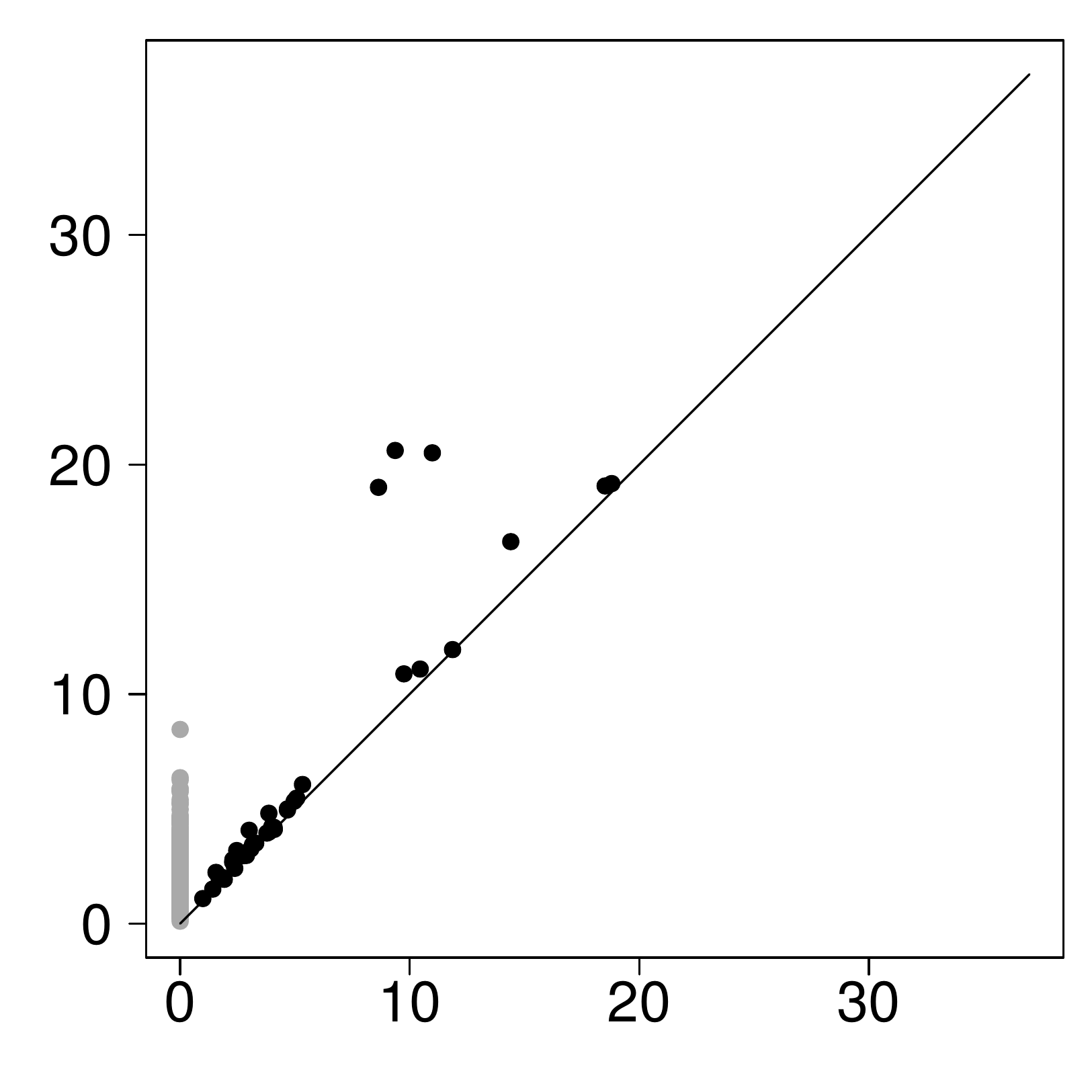}  
      \caption{Classes (2,3).}
      \label{fig:sub4f}
    \end{subfigure}
    \begin{subfigure}[t]{.32\textwidth}
      \centering
      \includegraphics[width=1\textwidth]{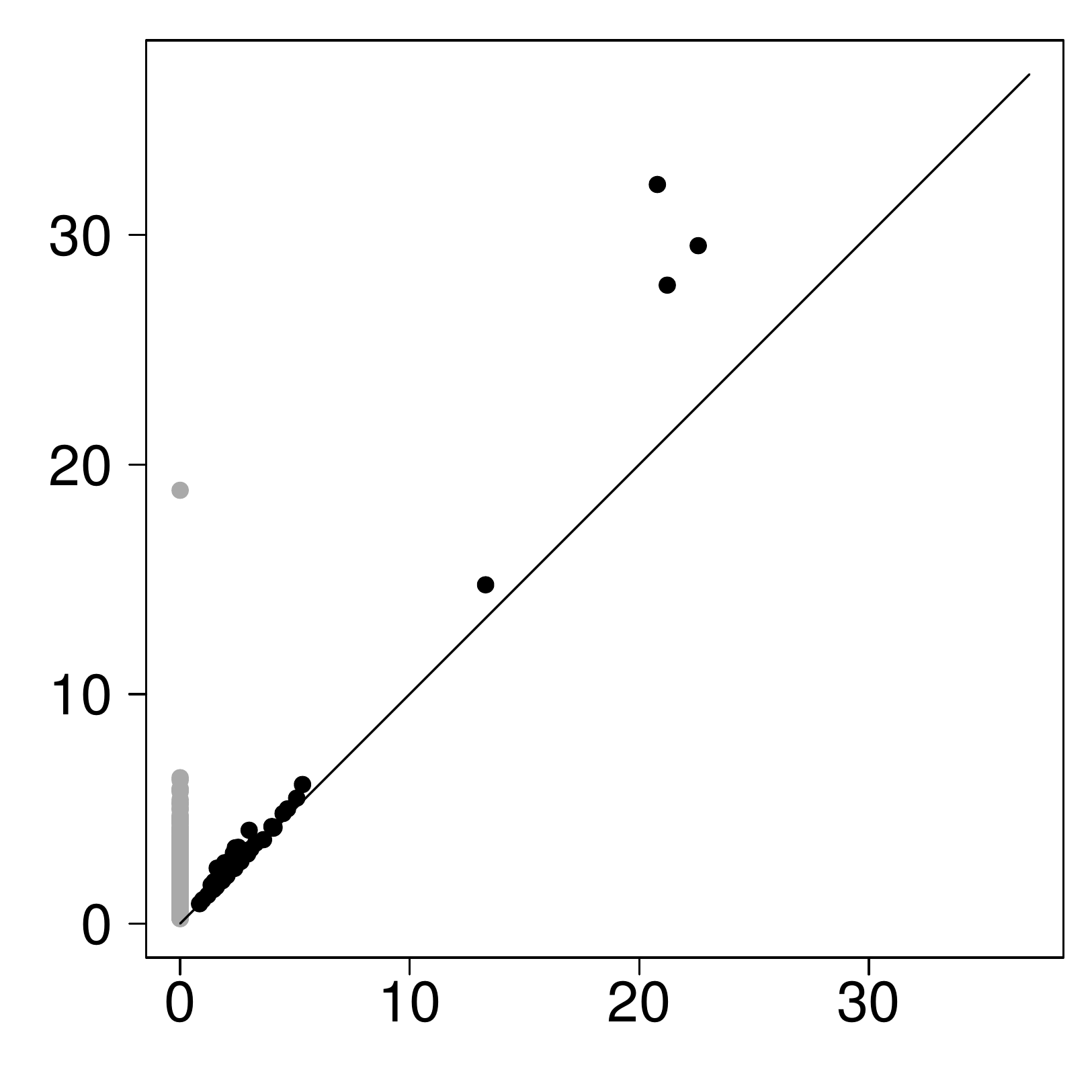}  
      \caption{Classes (1,3).}
      \label{fig:sub4g}
    \end{subfigure}
    \begin{subfigure}[t]{.32\textwidth}
      \centering
      \includegraphics[width=1\textwidth]{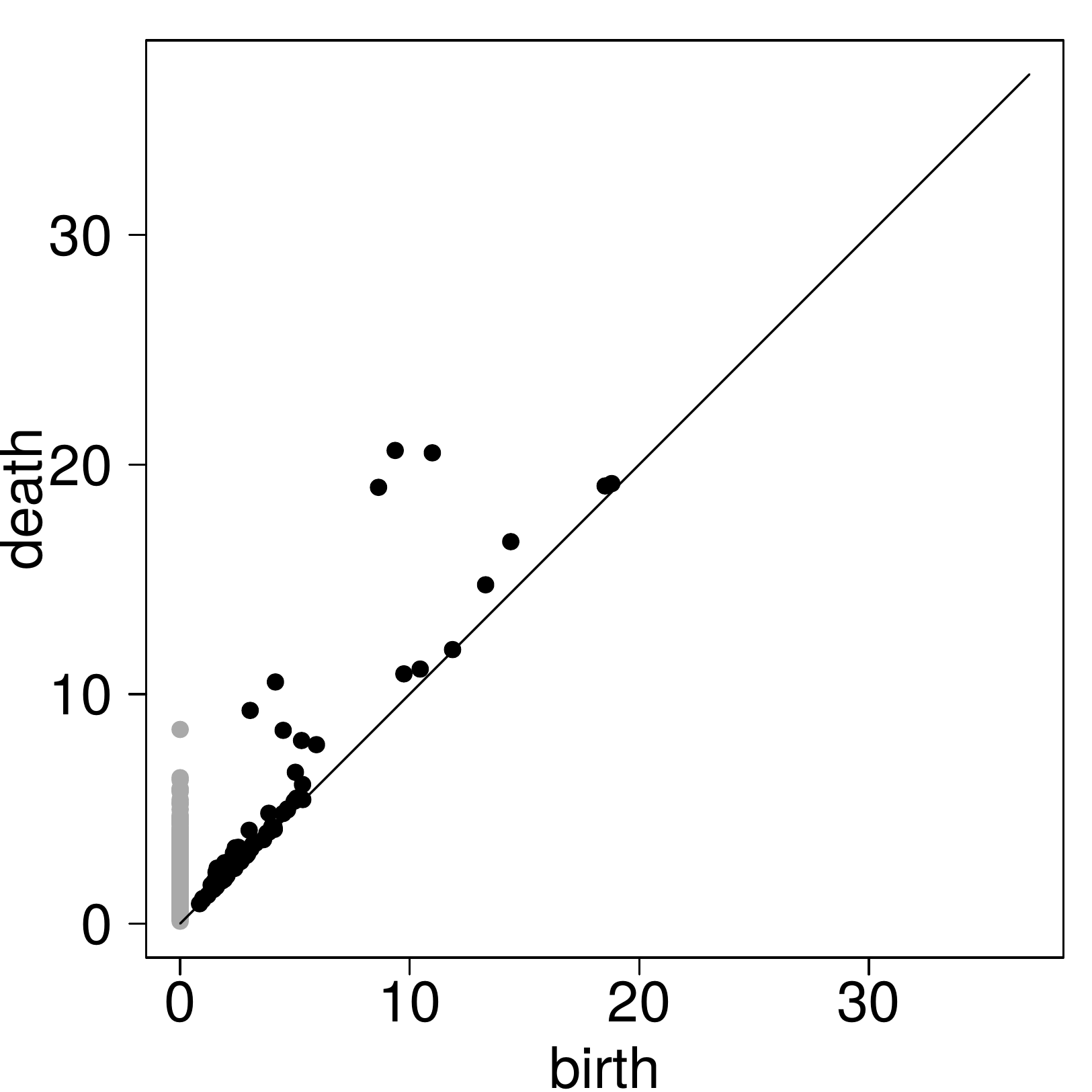}  
      \caption{Classes (1,2,3).}
      \label{fig:sub4h}
    \end{subfigure}
\end{center}
\caption{Persistence diagrams for combinations of classes 1, 2 and 3.
 \label{fig:4}}
\end{figure}

One can also construct persistence diagrams for the more noisy curves of Fig.~\ref{fig:2}. This is shown in Fig.~\ref{fig:4}.  The pattern is a bit more complex as is expected, but the individual points can be interpreted as before. In particular, due to the more irregular patterns of the noisy curves, the gray columns to the left extend farther up, and the birth of holes of dimension 1 has an earlier birth, there are more of them, and they exhibit a somewhat more complex pattern.

Note that the data set composed of the three Ranunculoids carry
topological information that is revealed by the persistence diagrams
embeddings of Figs.~\ref{fig:3} and \ref{fig:4}, but which cannot be discerned in the
embeddings of Figs.~\ref{fig:1} and \ref{fig:2}.

The idea is that this description of a  point cloud in the plane, as
indicated above,  may be generalized to higher dimensions and much
more complicated structures with multiple holes and voids of
increasing  homology. The number of sets of different homologies are
described by the so-called Betti numbers, $\beta_0, \beta_1,
\ldots$. In a non-technical jargon $\beta_0$ is the number of
connected components ($\beta_0 = n$, $n$ being the number of isolated
points in the start of our example), $\beta_1$  is the number of
one-dimensional holes, so $\beta_1= 1$ if there is only one connected
ring structure, and $\beta_0 = 1, \beta_1=0$ when the radius is so
great that there is only one connected set altogether. The hole is
one-dimensional since it suffices with a one-dimensional curve to
enclose it, whereas the inside of soccer ball is two-dimensional, it
can be surrounded by a two-dimensional surface, and has $\beta_0 =1,
\beta_1=0$ and $\beta_2=1$. A torus has $\beta_0=1, \beta_1=2,
\beta_2=1$. In Figs.~\ref{fig:3} and \ref{fig:4} it is a trivial exercise to find the Betti numbers (0 or 1) for any chosen interval of time (radius) of these figures.

The extension of the persistence diagrams to
more general structures requires relatively advanced use of
mathematical tools. We  only indicate some main concepts in Section
\ref{Intro} of the Supplement \citep{supp}.
The space of persistence diagrams is not a function space, and is
sometimes replaced by persistence landscapes which do form a function
space and may be more amenable to machine learning and statistical
analysis. The latter also brings in the need for statistical
inference, and recently statistical tools like the bootstrap has been
introduced in TDA. We refer to Sections 1.2 and 1.3 in the Supplement for more information and a number of references. Section 1 of that Supplement is concluded by formulating some explicit and open statistical problems in TDA.

There are many applications of TDA in general and of persistence diagrams in particular. Two recent applications to cancer research are \cite{bukk:ando:darc:2021} and \cite{craw:mono:chen:mukh:raba:2020}, 
where the latter introduces a variation of a persistent homology transformation to facilitate the difficulties in integration with traditional statistical models. In this type of cancer studies time series are important. The use of TDA to analyze time series data is discussed in \cite{ravi:chen:2019}.

\subsection{The Mapper\label{Mapper}}

In Section \ref{Nonlinear} we have outlined a number of methods for projecting high dimensional data to lower dimensions, thus making the projected data more amenable for characterization such as e.g. clustering and classification. Some of these methods strive to make the distance between points invariant, others not. But in all cases there is a risk of missing important topological information during the projection operation. The Mapper algorithm suggested in a seminal paper by \cite{sing:memo:carl:2007} tries to handle this issue by back-projecting the characterization in the lower dimensional space to the original space by considering preimages of the clustering, say, in the low dimensional space. More precisely, the Mapper algorithm consists of the following steps:

Consider a point cloud of data ${\bf X}$, and let $f$ be the mapping of ${\bf X}$ to a lower dimensional space, obtained by principal components or one of the other dimensionality reduction methods of Section \ref{Nonlinear}. Let ${\bf Y} = f({\bf X})$ be the set of data points in the lower dimensional space, often assumed to be $\mathbb{R}^m$ or even $\mathbb{R}^1$. Then
\begin{itemize}
\item[1.] Cover the range of values ${\bf Y} = f({\bf X})$ by a collection ${\mathcal U} = \{U_1,\ldots,U_S\}$ of intervals, or possibly more general sets, which overlap.
\item[2.] Apply a clustering algorithm to each of the preimages $f^{-1}(U_s), s=1,\ldots,S$. Even though $U_s$ may be connected, $f^{-1}(U_s)$ of course may not be connected due to the potential complicated topological relationships in the original space. This defines a pullback cover ${\mathcal C} = \{C_{1,1},\ldots,C_{1,k_{1}},\ldots,C_{S,1},\ldots,C_{S,k_{S}}\}$ of the point cloud ${\bf X}$, where $C_{s,k}$ denotes the $k$th cluster of $f^{-1}(U_s)$. 
\item[3.] Each node $v_{s,k}$ of the Mapper corresponds to one element $C_{s,k}$, and two nodes $v_{s,k}$ and $v_{s^{\prime},k^{\prime}}$ are connected if and only if $C_{s,k} \cap C_{s^{\prime},k^{\prime}}$ is not empty.
\end{itemize}
The algorithm results in a graph (or more generally a simplical complex).
The essential design problems consist in the choice of the transformation $f$ and the covering $U_1,\ldots,U_S$ in the lower dimensional space. Unfortunately, according to \cite{chaz:mich:2021}, Mapper is quite sensitive to the choice of covering, the number of covering sets and the overlap between them. Small changes in these design parameters may result in large changes in the output, making the method potentially unstable. A classical strategy consists  in exploring some range of design parameters, and selecting the ones that turn out to provide the most informative output from the user's perspective. Clearly, a more stringent approach would be desirable. This could also be said  for determination of parameters in other embedding algorithms; see keypoint 1 in the concluding remarks of Section \ref{Conclusion}.

There is a statistical analysis including parameter selection in \cite{carr:mich:oudo:2018}. They demonstrate aspects of statistical convergence and ensuing optimality problems. They also derive confidence regions of topological features such as loops and flares.

The Mapper algorithm has found many applications, especially for its capability of detecting loops and flares in the mapping of the original data space. A recent example of applications to cell description is given in \cite{carr:raba:2020}.

\section{Embedding of networks \label{Network}}

In Sections \ref{ISOMAP} and \ref{Laplace eigenmaps} graphs (or networks) were used as a tool in embedding a point cloud in $\mathbb{R}^{p}$, making it possible among other things to do cluster analysis involving non-convex clusters. In the present section the {\it starting point} is a network or collection of networks, and the task is to embed the network in an Euclidean space $\mathbb{R}^m$ or to map it to a  manifold. This is used to obtain a vector representation of each node of the network.

Why is it important to be able to embed a network in such a way? The main reason is simply that for many purposes it is easier to work with a set of $n$ vectors than with a network consisting of $n$ nodes. One has standard methods for dealing with vectors. For example one can do clustering of vectors, which in a social network could correspond to  finding and grouping communities in the network. And one can also compare and classify networks by looking at their embedded sets of $n$-dimensional vectors. 

With the increasing use of the internet and Big Data, the analysis of large networks is becoming more and more important. There is a very wide field of applications ranging over such diverse areas as e.g.\ finance, medicine and sociology, including criminal networks. A broad overview can be found in the recent book by \cite{newm:2020}. A fine detailed survey is \cite{cui:wang:pei:zhu:2019}.

With ultra-high dimension and very large data sets, there is a need
for fast methods. With the recent technique of Skip-Gram, described in
some detail in Section \ref{Skip-Gram main} and in Section 2 in the
Supplement \citep{supp}, one is able to handle networks with millions of nodes and billions of edges such that each node is represented by a vector of dimension 500-600, say. On such vectors one can use standard discrimination and clustering. One may also do further embedding to lower dimensional vectors, as described in Section \ref{Visualization}, to visualize data of very high dimension.

In our survey of network embedding methods, we will start  with spectral graph
methods in Section \ref{Spectral} after a brief introduction on
characterization of graphs in Section \ref{Graph theory}. The spectral
method requires the solution of an eigenvalue problem, and this puts a
limitation on the number of nodes and edges. This restriction is to a
large degree bypassed in neural network based methods, in particular
in the Skip-Gram algorithm. This algorithm was originally introduced
in natural language analysis, which has independent interest in that
the words in a language text can be embedded in a vector in
$\mathbb{R}^m$ reflecting not only the word count in a text but also
the syntax of the text. A language text is not a network, and
therefore the detailed embedding analysis of a language text is
covered in Section 2 of the Supplement. Ideas and methods developed in such a framework have proved vitally important, however, for fast and efficient embedding of networks
as is demonstrated in Section \ref{Skip-Gram main}. That section is chiefly concerned with symmetric undirected networks, but briefly mentioning  directed networks, heterogeneous networks and dynamic networks, where there are many open statistical and data processing problems, in the ensuing sections.

The research on embedding of networks has mainly been published in machine learning journals and conference proceedings. There are several issues of statistical interest, and one may think that there is a potential synergy effect that both the statistics and machine learning community could benefit from. We will try to make this more clear in the sequel. One issue is the lack of statistical modeling and inference in the algorithmic machine learning industry. It is important to realize that there now exists a  growing statistical literature that is in process of being integrated in algorithms on finding communities in networks. We refer to Sections \ref{SBM1} and \ref{SBM2}. See also the three keypoints formulated in the concluding remarks in Section \ref{Conclusion}.

\subsection{A few elementary concepts of graph theory and matrix representations\label{Graph theory}}

We have already introduced some elementary graph concepts in Sections \ref{ISOMAP} and \ref{Laplace eigenmaps}. In this brief introductory section we supplement these to more fully explain the spectral based clustering algorithms for networks.  

We consider a graph $G=(V,E)$, where $V$ and $E$ are the sets of nodes
and edges, respectively. The graph is supposed to be undirected, which
means that an edge goes in both directions between two neighboring
nodes. Let $n = |V|$ be the number of nodes in $(V,E)$. Then the graph
can be represented by a $n \times n$ matrix ${\bf M}$, such that an
element $M_{ij}$ of this matrix represents some property of the pair
of nodes $v_i$ and $v_j$. When $V$ is large, this matrix may be
huge. Later, representation  matrices of dimension $m \times n$ will
be introduced where $m < < n$. Diagonal elements $M_{ii}$ encode
information of the node $v_i$ only, such as the degree of $v_i$
(number of edges emanating from $v_i$ or more generally as in Equation
(\ref{25i}) for a weighted graph).

A simple example of such a matrix is the adjacency matrix $A$, which was mentioned in Section \ref{Laplace eigenmaps}.

It is clearly desirable for network patterns to be independent of the way one labels the nodes, and one is therefore interested in quantities that can be derived from ${\bf M}$ that are invariant to permutations.  One such quantity is the spectrum of ${\bf M}$. It is well known from elementary linear algebra that the spectrum of a matrix is invariant to similarity transformations such as that produced by a permutation matrix. Unfortunately, cospectrality of two adjacency matrices does not necessarily mean that the corresponding graphs are isomorphic.\footnote{Two graphs $G$ and $H$ are isomorphic if there exists a bijection $f$ from $G$ to $H$ such that two nodes $u$ and $v$ in $G$ are adjacent if and only if $f(u)$ and $f(v)$ are adjacent in $H$.} This may not be a serious problem in practice, though, as is indicated in \cite{wils:zhu:2008}, and following recent literature on embedding of networks it will be ignored in the sequel. 

An adjacency matrix $A$ for an undirected graph is symmetric with real eigenvalues, both negative and positive. In many applications it is useful to have a non-negative definite matrix. Such a matrix has non-negative eigenvalues. One example of such a matrix is the Laplace matrix, a version of which was introduced in Section \ref{Laplace eigenmaps} for a general weighted undirected graph. It is given by
\begin{equation}
{\bf L} = {\bf D}-{\bf A},
\label{51a}
\end{equation}
where ${\bf A}$ is the adjacency matrix and ${\bf D} = \mbox{diag}(d_i)$ is the diagonal matrix having the degree of the nodes along the diagonal.

The normalized Laplacian $L_N$ is defined by
\[
L_{N,ij} = \left\{
\begin {tabular}{cc}
1 & if $i=j$ \\
$-1/\sqrt{d_i d_j}$ & if $i$ and $j$ are adjacent \\
0 & otherwise.
\end{tabular}
\right.
\]
This matrix can also be written ${\bf L}_N = {\bf D}^{-1/2} {\bf L}{\bf D}^{-1/2}$. It is non-negative definite and it has all its eigenvalues $0 \leq \lambda \leq 2$.

\subsection{Spectral embedding and graph clustering\label{Spectral}}

A basic task in network clustering is  community structure detection. It is perhaps best thought of as a data technique used to throw light on the structure of large-scale network data sets, such as social networks, web data networks or biochemical networks. It is normally assumed that the network of interest divides naturally into subgroups, and the task is to find those groups. 

For the purpose of community grouping and division a criterion  is required that can measure both the internal structure within each group, where the goal is to maximize the dependence between members of a group, but such that the dependence between each group is minimized. There are two main methods for doing this, either by minimizing the so-called cut between the groups, the mincut problem or by maximizing the modularity. Both are discussed below using network spectral embedding.

\subsubsection{Minimizing the cut functional\label{Cut functional}}

A useful tutorial on spectral clustering is given by  \cite{vonl:2007}. A more recent alternative account is given in \cite{zhen:2016}.

Given a graph $G = (V,E)$ with adjacency matrix $A$ we would like to
find a partition of $V$ in groups $V_1,\ldots,V_k$ such that the
number of edges between each group is minimized. This leads to the
mincut problem.

Let $W(V_i,V_j) \doteq \frac{1}{2} \sum_{m  \in V_i,l \in V_j} w_{ml}$, where $w_{ml}$ is the weight for the edge between  the nodes $v_m$ and $v_l$. In the unweighted situation  $w_{ml}$ is 1 if there is an edge between $v_m$ and $v_l$ and 0 if not. Let $\bar{V}_i$ be  the complement of $V_i$. The mincut  approach to clustering is simply defined for a given $k$ by choosing the partition $V_1,\ldots,V_k$ which minimizes the normalized cut size
$$
\mbox{NCut}(V_1,\ldots,V_k) \doteq \frac{1}{2} \sum_{i=1}^{k}\frac{W(V_i,\bar{V}_i)}{\mbox{vol}(V_i)} = \sum_{i=1}^{k}\frac{\mbox{cut}(V_i,\bar{V}_i)}{\mbox{vol}(V_i)},
$$
where $\mbox{vol}(V_i) =\sum_{v_{l} \in V_i}d_l$, $d_l$ being the weighted degree of $v_l$. A similar criterion is the RatioCut criterion.

The normalized Laplace matrix can be written as ${\bf L}_N = {\bf D}^{-1/2}{\bf L}{\bf D}^{-1/2}$. Let ${\bf H}$ be the $n \times d$ matrix whose columns are the $d$ eigenvectors corresponding to the $d$ smallest (non-zero) eigenvalues of ${\bf L}_N$. The $n$ $d$-dimensional row vectors of ${\bf H}$ then constitute an embedding of the nodes of the graph minimizing the normalized cut-functional of the graph. These embedding vectors are then used as a point of departure for clustering and classification of nodes.

\subsubsection{Maximizing the modularity\label{Modularity}}

Modularity is an alternative concept in the use of spectral methods in clustering. Modularity was introduced by the highly cited papers of \cite{girv:newm:2002} and \cite{newm:girv:2004}, and after that has been further developed as in \cite{newm:2006b}. See also \cite{bick:chen:2009} for an alternative using a nonparametric point of view. 

It was seen in the previous subsection that the principle underlying the cut-size algorithms is that a good division of a network is one in which there are few edges between communities. \cite{newm:2006b} states that this is not necessarily  what one should look for. He argues that a good division is one in which there are fewer than {\it expected} edges between communities. 

This idea, then, is quantified using the measure of modularity. 
Assume first that there are two potential classes. Again we suppose that the network contains $n = |V|$ nodes, and we introduce the vector $s$, whose $i$th component is given by $s_i = 1$ if node  $v_i$ belongs to group 1 and $s_i=-1$ if it belongs to group 2. The edge between nodes $v_i$ and $v_j$ is characterized by the adjacency matrix ${\bf A}$. The element $A_{ij}$ then represents the ``number of edges'' between $v_i$ and $v_j$. The expected number of edges between $v_i$ and $v_j$ if edges are placed at random is $d_id_j/2m$, where  $d_i$ and $d_j$ are the degrees of the nodes and  $m= \frac{1}{2}\sum_i d_i$ (undirected network). The modularity is then defined by
\begin{equation}\label{eq:Q}
Q = \frac{1}{4m} \sum_{ij}\Bigl(A_{ij}-\frac{d_id_j}{2m}\Bigr)s_is_j = \frac{1}{4m}s^T{\bf B}s,
\end{equation}
where the matrix ${\bf B}$ is defined by
$$
B_{ij}=A_{ij}-\frac{d_i d_j}{2m}.
$$
This is easily generalized to the case of $k$ classes, and the modularity is maximized by computing the eigenvectors of the ${\bf B}$ matrix. Corresponding to ${\bf H}$, let ${\bf S}$ be the $n \times d$ matrix whose columns are the eigenvectors corresponding to the top $d$ eigenvalues of ${\bf B}$. The $n$ $d$-dimensional row vectors of ${\bf S}$ then constitute an embedding of the $n$ nodes of the network maximizing the modularity.

\subsubsection{The Louvain method for community detection\label{Louvain method}}

The so-called Louvain method for community detection based on modularity was introduced in a paper by \cite{blon:guil:lamb:lefe:2008}. They start with a network with $n$ nodes, and where each node defines a community. Then one goes successively through the nodes of the net and for each node $v_i$, with neighbors $v_j$ one investigates the gain in modularity if $v_i$ is removed from its community and placed in the community of $v_j$. The node $v_i$ is then placed in the community for which this gain is maximum (in case of a tie, a breaking rule is used). An updating formula for the change in the modularity $Q$ is given in \cite{blon:guil:lamb:lefe:2008}. This is continued until the whole graph has been covered. In the next round the procedure in the first round is repeated, but this time with the communities formed in the first step as entities. This is continued until there is no increase in $Q$. 

There is no eigenvalue problem that needs to be solved in this algorithm. This makes it possible to apply the Louvain algorithm for substantially larger networks. One example that the authors refer to is a mobile phone company with a network composed of 2.6 million users. 

\subsubsection{Statistical modeling, SBMs and finding communities\label{SBM1}}

The methods in Sections \ref{Cut functional}-\ref{Louvain method} all belong to the algorithmic approach. An intuitively reasonable object function is maximized or minimized to find communities in a network. This is in line with the most popular approach to statistical embedding, where as such no statistical model is involved. There are no parameters that should be estimated, and in terms of which the fit of the model can be assessed.

These two different approaches, the algorithmic versus the statistical modeling one, have recently been discussed in several papers. The most recent one seems to be \cite{peix:2021}, who is staunchly critical to the algorithmic approach in general and to the methods of finding communities in Sections \ref{Cut functional}-\ref{Louvain method} in particular. The author demonstrates that maximizing the modularity $Q$ of subsection \ref{Modularity} could lead to falsely finding communities in a completely random environment. On the other hand he gives examples where in given situations use of $Q$ leads to underestimation of the number of communities. This may be part of a general problem of some machine learning algorithms, at least it is something that deserves closer attention, as indicated in the third keypoint of Section \ref{Conclusion}.

Peixoto argues for parametric statistical models from which networks can be {\it generated}, and where the structure of the net depends on the type of statistical models used and on the values of the parameters of these models. The generated model structures can be compared to real life networks, and parameters may be estimated by seeking to fit a generated model structure to the real life data. The most used statistical model is probably the so-called stochastic block model, SBM, where a block may be thought of as a community. The history of these models goes back at least to \cite{holl:lask:lein:1983}. Another early publication for a slightly more general model is \cite{hoff:raft:hand:2002}. There are several papers on the theoretical aspects of the SBM that will be briefly mentioned in Section \ref{SBM2}. A review paper is \cite{lee:wilk:2019}. Here we will base ourselves on \cite{karr:newm:2011} and \cite{newm:rein:2016}, since they are directly and explicitly related to maximizing modularity $Q$, Newman being the main originator of the modularity principle.

In the simplest undirected stochastic block model (many more complicated ones are possible) each of the $n$ nodes is assigned to one of $k$ blocks (communities), and undirected  edges are placed independently between node pairs with probabilities that are a function only of the block membership of the nodes. If we denote by $b_i$ the block to which node $i$ belongs, then one can define a $k \times  k$ matrix of probabilities such that the matrix element $p_{v_{i},v_{j}}$ is the probability of an edge between nodes $i$ and $j$. These probabilities are the $k^2$ parameters of the model, and there are several ways of estimating them for a given real data network.

Unfortunately, however, this simple block model does not work well for many real world networks, and tends to give bad results in obtaining plausible communities. There are generalizations of the simple SBM model, but they may lead to models that are far more difficult to estimate. One relatively simple generalization is the degree corrected stochastic block model (dcSBM) that seems to work much better on real life networks. We have used it to generate simulated networks in Section \ref{Visualization}. The dcSBM was suggested by \cite{karr:newm:2011}. It allows for heterogeneity in the number of degrees for the nodes, which is a phenomenon that is often observed in practice, whereas the simple SBM results in a model where each node has the same expected degree, which in many cases is clearly unrealistic. Karrer and Newman also demonstrate that in a certain approximative sense the dcSBM can be related to the modularity function $Q$ from Equation \eqref{eq:Q}.

On the famous karate club example the dcSBM does very well, much better than the simple SBM, which actually fails completely for this example.

There are many exciting statistical and probabilistic issues that are presently being worked on, and we will briefly mention some of them in subsection \ref{SBM2}.

\subsection{Embedding a network using Skip-Gram\label{Skip-Gram main}}

It should be noted at first that for large networks the cut-size spectral clustering method and the modular method (possibly with the exception of the Louvain method) run into problems because it is costly to solve eigenvalue problems for the high dimensions that may occur in network embedding.

These problems are to a large degree alleviated in a neural net based
Skip-Gram procedure. This procedure was first developed in word
embedding in a language text (from this the nomenclature
``Skip-Gram''). Here the eigenvalue problem is eliminated altogether,
and the neural net training is speeded up using so-called negative
sampling or hierarchical processing. At this point the reader may wish
to browse through Section 2 in the Supplement \citep{supp}, which contains a relatively detailed account of natural language embedding. This may, we believe, be of some independent interest. An effort will be made to make the current section on embedding of networks self-contained, just leaving some details to Section 2 of the Supplement.

What is needed, then,  to extend word processing to networks where words are replaced by nodes and the vocabulary with the network itself? The simple answer  is the concept of a neighborhood. 

In natural language processing, defining a neighborhood of a word in a text is not difficult:  simply taking $n_1$ and $n_2$, $n_i \geq 0$ context words in front and after the word respectively. One may think that a corresponding neighborhood around a node is easily defined, but not quite so, because here there is no natural ``past'' or ``future''. 

Before  embarking on the neighborhood problem, partly to define notation, let us formally write up the analog of the Skip-Gram model, presented in some detail in the language analysis in Section 2 of the Supplement,  for a network.  The notation $N(v)$ is used for the neighborhood of a node $v \in V$ in a network $G=(E,V)$. Neighborhoods are more precisely defined in Section \ref{Neighborhood sampling}.
The analysis  to be presented next applies mainly to the static undirected case.
Extensions to directed, heterogeneous and dynamic networks are briefly discussed in separate subsections.  

We let $f$ be the mapping from $V$ to the embedding feature space $\mathbb{R}^m$. The goal is to associate each node $v$ in $V$ with a feature vector $f(v)$ in $\mathbb{R}^{m}$. When representing the whole network in this way we obtain a $n \times m$ matrix with $n=|V|$.

\subsubsection{The Skip-Gram\label{Skip-Gram}}

We proceed to formulate the Skip-Gram architecture for an undirected symmetric network.  One seeks to optimize an objective function in finding a representation $f(v)$  such that  the conditional probability for obtaining individually the elements in $N(v)$, given an input node $v$, is maximized; i.e,
find $f$ such that 
\begin{equation}
\sum_{v \in V} \log P(N(v)|f(v)) 
\label{53a}
\end{equation}
is maximized. 

The maximization is done by training a one-layer hidden neural network
which has as possible inputs $n$ vectors, one for each node in the
network. A fixed input vector has as desired output a probability
distribution on the nodes. It should be concentrated as well as
possible to the neighbors (suitably defined) of the input node. The
idea is to train the neural net through its hidden layer so that this
is achieved to the highest possible degree. Only linear
transformations are used from the input layer to the hidden layer and
essentially also from the hidden layer to the output, although a
logistic type transformations is used to transform the outputs to
probabilities. A few basic facts of neural networks are given in
Section 2.1 of the Supplement \citep{supp}.

The training is done successively by going through this process for each input node several times and is stopped when the deviation from the obtained  probability distribution on the outputs is close enough to the ideal desired one, which is completely concentrated on the sought neighboring nodes. At each step of this procedure each node has an input vector representation and an output vector representation. It is the output vector representation that is of interest since it describes the relation between a node and its neighbors. This training process strives to maximize the function in (\ref{53a}).

To make this optimization problem tractable, the following two
assumptions are made (not always made explicitly in the language processing papers).

\begin{itemize}

\item[1)] Conditional independence: The conditional likelihood is factorized as
\begin{equation}
p(N(v)|f(v)) = \prod_{n_i \in N(v)} P(n_i|f(v)).
\label{53b}
\end{equation}
 
\item[2)] Symmetry in feature space and softmax: A source  node
and a neighborhood node have a symmetric effect on each other in the embedding feature space. Accordingly, the conditional likelihood for every source-neighborhood pair is modeled as a softmax unit, parameterized by a dot product of their features
\begin{equation}
P(n_i|f(v)) = \frac{\exp(f(n_i) \cdot f(v))}{\sum_{u \in V} \exp(f(u) \cdot f(v))}. 
\label{53c}
\end{equation}
This is nothing but a suitable parametrization of the multinomial
logistic regression model, but in the data science literature "softmax
unit" is preferred. Formula \eqref{53c} may be compared to the
development in Section 2.4 in the Supplement \citep{supp}.

\end{itemize}

With the above assumptions and taking logarithms in (\ref{53c}), the objective function in Equation (\ref{53a}) simplifies to
\begin{equation}
\max_{f} \sum_{v \in V} \Bigl[-\log(\sum_{u \in V} \exp(f(v) \cdot f(u))) +\sum_{n_i \in N(v)}f(n_i) \cdot f(v) \Bigr]. 
\label{53d}
\end{equation}

In the training of the neural net one avoids solving a high dimensional eigenvalue problem, but there is an obvious computational issue involved. As the size of the network increases with $n$, the neural net with the associated input and output vectors representations becomes heavy to update. For each step of the training, in principle, all of these representations have to be updated. The updating of the node input vectors is cheap, but learning the output vectors, which are the vectors of interest, is expensive. For each training instance one has to iterate through every node of the network, cf.\ the summation over $u$ in \eqref{53c} and  \eqref{53d}, compute the output and the prediction error and finally use the prediction error in a gradient descent algorithm to find the new output vector representation.

The idea of negative sampling, first introduced in
\cite{miko:suts:chen:corr:dean:2013b} in text analysis, makes the
training process amenable by not sampling over the entire network for
each update of a node, but rather a small sample of nodes. Obviously,
the output nodes in the neighborhood of a given node should be
included in the update sample, i.e., the last sum of \eqref{53d}. They represent the ground truth and are termed positive
samples. In addition a small number $k$ of nodes (noise or negative
samples) should be updated. \cite{miko:suts:chen:corr:dean:2013b}
suggest that $k=5-20$ are useful for small training sets, whereas for
large training sets $k=2-5$ may be sufficient, see Section 2.6 of the
Supplement for more details \citep{supp}. The sampling is via a probability mechanism where each word (node) is sampled according to its frequency in the text. It will be seen below how this can be done in the network case. In addition, \cite{miko:suts:chen:corr:dean:2013b} recommends, from empirical experience, that in the further analysis each frequency should be raised to the power of 3/4 (cf.\ again Section 2.6 of the Supplement). This seems also to have been adopted in the network version of negative sampling. Clearly, a more thorough statistical analysis, also including the choice of $k$, would be of interest We refer again to the first item of the three keypoints of Section \ref{Conclusion}.

We will return to the question of negative sampling in the next subsection, where a sampling strategy $S$ is introduced for creating neighborhoods of a node $v$, such that the neighborhood $N_V(S)$ depends on $S$. 

\subsubsection{Neighborhood sampling strategies\label{Neighborhood sampling}}

Various authors have suggested different sampling strategies of the nodes of a network. We will go through three main strategies which seem to be representative of this field as of the last 5 years. All of these contain parameters to be chosen for which, to our knowledge, an optimality theory is lacking.

\cite{perr:alfr:skie:2014} device a sampling strategy they call ``DeepWalk''.
Consider a node $v$, and denote by $w_{vu}$ the weight of its (undirected) edge with another node $u$. Let the degree variable be $d_v = \sum_u w_{vu}$. Then start a random walk from $v$ by letting it choose the one-step neighbor $u$ with probability $P(u|v) = w_{uv}/d_v$. Next, repeat this for the node $u$, and so on until $L$ steps, say, have been obtained. The walk may return to $v$ for one or more of its steps.
This procedure is now repeated $\gamma$ times obtaining $\gamma$ random walks starting in $v$. These may be compared to text segments in natural language processing. 
Analog to a moving window in a language text we now let a window of size $2K+1$, where $2K+1 \leq L$, glide along the random walk paths. For each window, there is a center node numbered $u^{\prime}$, 
$K \leq u^{\prime} \leq L-K$, and we define a neighborhood $N_S(u^{\prime})$ and $K$ nodes prior to $u^{\prime}$ and K nodes after $u^{\prime}$ in the considered random walk path. For each such configuration we apply the Skip-Gram procedure \eqref{53a} - \eqref{53d}.  In this way, for each node $v$ we generate $\gamma \times (L-2K)$ segments of nodes. 
Note that this creation of segments in paths of random walks can be carried out before the optimization process takes place. When applied to all of the nodes of the network it results in a collection of $n \times \gamma \times (L-2K)$ segments of nodes that correspond to windows of words in a language text. This sets up a frequency distribution over the nodes corresponding to the frequency distribution of words in the vocabulary in a text. Negative sampling of nodes can then be applied to this frequency distribution of nodes.

In the DeepWalk set-up the random walk can only go to one of the nearest neighbors in the next step with the probability $P(u|v) = w_{vu}/d_v$. 
\cite{grov:lesk:2016} argue that a combination of so-called  breadth first sampling, BFS and depth first sampling, DFS, should be used.

The LINE (Large-scale Information Network Embedding) was introduced by \cite{tang:qu:wang:zhan:yan:mei:2015b}. They  use a slightly different optimization criterion than (\ref{53b}). Somewhat similarly to \cite{grov:lesk:2016} LINE introduces the concepts of first and second order proximities. In each of these papers on sampling strategies there are a number of comparative experiments on information networks such as Wikipedia, Flickr, YouTube to evaluate the properties of each method.

\cite{qiu:dong:ma:li:wang:tang:2018} obtain a unifying view of the DeepWalk and LINE among other algorithms. In the more recent \cite{qiu:dong:ma:li:wang:tang:2019}
they look at the practical and algorithmic aspects of an implied matrix factorization with associated sparse matrices, resulting in the algorithm NetSMF. Software packages are available for all of the algorithms mentioned in this section.

\subsection{Directed network\label{Directed}}

In many applications of networks one deals with a directed network, e.g.\ in causality networks. This is a network where the weight on edges between nodes $v_i$ and $v_j$ may be different, so that $w_{ij} \neq w_{ji}$, and one may even have 
$w_{ij} > 0$ but $w_{ji} = 0$. \cite{rohe:qin:yu:2016} have looked at this from a spectral graph point of view. For a directed graph the adjacency matrix, giving the edge weights $w_{ij}$, is not symmetric. When the adjacency matrix is not symmetric, the left and right eigenvectors are in general not the same. This can be given an interpretation as ``sending'' and ``receiving'' nodes, and it can be argued that these should be clustered separately.

Directed graphs have also been attempted incorporated in the Skip-Gram
procedure, see e.g.\ \citet[p. 2944]{zhou:liu:liu:gao:2017}. The undirected sampling strategy described in Section
\ref{Neighborhood sampling} can again essentially be used. To
illustrate, let $w_{ij}$ be the weight of the edge in a transition
from $v_i$ to $v_j$. In a money laundering investigation, for example, where the nodes may be bank accounts, $w_{ij}$ may be proportional to the number of transactions from account $i$ to account $j$. Similarly, one may define $w_{ji}$. The probability of going from node $v_i$ to $v_j$ can then be given as $p_{ij} = w_{ij}/d_i$, where $d_i = \sum_{j \in N_{S}(i)} w_{ij}$ and $N_S(i)$ is the first order neighborhood of $v_i$.  
This is extendable to higher order neighborhoods as in \cite{grov:lesk:2016}.

\subsection{Heterogeneous network representation\label{Heterogeneous}}

Heterogeneous here refers to a situation where there are different types of nodes in a network, and there may be different types of edges. If these are treated with homogeneous techniques neglecting the heterogeneity, inferior results may result. 

Two papers will be briefly mentioned, one is an extension of the LINE approach, the other is an extension of the DeepWalk methodology. In these two papers the Skip-Gram algorithm is applied on so-called metapaths, paths consisting of a sequence of relations defined between {\it different} node types. The introduction of metapaths to heterogeneous graphs came before the Skip-Gram procedure. See \cite{sun:nori:han:yan:yu:yu:2012}. 

It is natural also to mention the extension of LINE found in the PTE (Predictive Text Embedding) of \cite{tang:qu:mei:2015a}. PTE deals with a text network embedding, but the method is applicable to a general network.


\cite{dong:chaw:swam:2017}  introduce a form of random walk sampling for heterogeneous networks which is analogous to or extends the sampling procedures in \cite{perr:alfr:skie:2014} and \cite{grov:lesk:2016}. Skip-Gram is combined with the metapath sampling as discussed by  \cite{sun:nori:han:yan:yu:yu:2012}.

Although there are different types of nodes in $V$, their representations are all mapped into the same latent space $\mathbb{R}^{m}$. 

\subsection{Embedding of dynamic networks\label{Dynamic}}

Most of the work on embedding of networks has been done on static networks. There is no time dimension involved to trace the dynamic evolution of the network. In many situations this is of course not very realistic. Consider for example a bank network. New accounts are opened, other accounts are closed. New types of transactions between accounts are appearing, others are becoming old and less relevant. Or in more general network language: New nodes are coming into the network, others are removed. New edges are created, others are discarded. Weights between edges may easily change in time. In a heterogeneous network new types of nodes may enter the system, others may leave. An early empirical investigation of changes in social networks is contained in \cite{koss:watt:2006}.  See also \cite{gree:doyl:cunn:2011}.

An obvious brute force solution is to use a moving window and then do an embedding, and possible clustering in each window. But clearly such a procedure is time consuming and non-efficient if there are many (overlapping) windows. One would like to have an updating algorithm that can keep information in the previous window and combine it with new information in the new window. To our knowledge the literature here is quite limited. 

\cite{zhou:yang:ren:wu:zhua:2018} 
consider triads as basic units of a network. A triadic closure process is aiming to capture the network dynamics and to learn representation vectors for each node at different time steps.

There is also a recent attempt to generalize the entire Skip-Gram methodology to a dynamic framework. This can be seen in \cite{du:wang:song:lu:wang:2018}. 
They utilize that a network may not change much during a short time in dynamic situations, thus the embedding spaces should not change too much either. 
A related paper venturing into heterogeneous networks  meta paths is \cite{bian:koh:dobb:divo:2019}. \cite{zhu:pan:li:liu:wang:2017}, takes a more statistical modeling point of view on dynamic networks. The paper is briefly reviewed in the next subsection. Clearly, the theme of dynamic networks is an open and challenging field for data scientists and statisticians. Much recent work is summed up, mostly from a machine learning point of view, in \cite{kaze:goel:jain:koby:seth:fors:poup:2020}.

\subsection{Network embedding: Data science and machine learning versus statistical modeling\label{Data science}}

An overwhelming part of the literature on network embedding can be found in the machine learning journals and in proceedings on data and computational science. The emphasis has been on deriving methods that ``work'', i.e.\ can be used in practical applications. Certain parts of some of the methods used are quite ad hoc such as the argument in \cite{miko:suts:chen:corr:dean:2013b} where from empirical evidence the word count is raised to 3/4 power in the distribution forming the basis of the negative sampling. This has been followed up in later literature and does seem to work well. But it is not clear why. Moreover, there are few quantitative expressions of uncertainty or on statistical properties of the obtained results.

Many of the algorithms and methods discussed in this paper contain input parameters or hyper parameters, including the choice of the dimension of the embedding space. An important issue in both theory and practice is the setting of these parameters. The problem has to be treated with care to avoid instability in the embedded structure. The problem is briefly mentioned in Section \ref{Mapper}, but the problem is relevant also in a more general context. 

Broadly speaking, statistical methods use theoretically derived methods to choose hyper parameters necessary to fully specify a method, while the typical machine learning approach is to rely on hyper parameter optimization or so-called tuning. The former may require assumptions that are too strong or cannot be checked in practice. The latter typically requires additional data or re-training of models based on randomly dividing the data into subsets (cross validation), which is computationally costly and comes with an uncertainty component due to the randomness in the data splitting.  Many machine learning practitioners may enforce a rather basic and ad-hoc trial and error optimization approach. Still,  methods like Bayesian optimization \citep{shahriari2015taking} have gained significant momentum in the recent years. Bayesian optimization aims at solving the optimization problem using as few evaluations as possible. While the method uses statistical theory through its reliance on Gaussian processes, the hyper parameter selection problem is still based on optimization and possesses the aforementioned drawbacks.
We think the machine learning methods could benefit from theoretically derived hyper parameter choices. There have been some attempts at choosing parameters for machine learning methods through the statistical information criteria approach \citep{claeskens2008information, lunde2020information}, but it does not yet seem to have found its place in machine learning.
The theoretical difficulty of deriving such criteria due to the lack of proper likelihoods in the training of the machine learning methods is an obvious obstacle. To avoid this, it might be possible to go in the direction of the generalized information criterion (GIC) \citep{konishi2008information} which does not require a likelihood, but rather relies on functionals of the data generating distribution and their associated influence functions. In any case, going forward, we believe it is worth looking in the direction of theoretically derived selection procedures for the machine learning community, and have as such identified this in our list of keypoints in Section \ref{Conclusion}.

\subsubsection{Stochastic block modeling\label{SBM2}}

The issues mentioned above appear to lead to a gap between data/computational science using algorithmic approaches and more traditional (and modern) statistical thinking. There is a clear need for results bridging this gap, as argued in the second keypoint in Section \ref{Conclusion}. This should be helpful for both disciplines.  There is some good news: As indicated in Section \ref{SBM1} there is a recent trend represented by stochastic block models and related models to bring statistical modeling and statistical inference on these models into network embedding and community detection.

The concept of formal statistical consistency has been brought into recent network embedding literature. Stochastic block modeling has the property that under certain conditions consistency is obtained in the sense that if the method is applied to a network actually generated from a specific block model, then it can correctly recover the block model asymptotically. 

In particular Peter Bickel and his collaborators have taken up various problems of asymptotic theory for stochastic block models and related models. This includes hypothesis testing in \cite{bick:sark:2016}, asymptotic normality in \cite{bick:choi:chan:zhan:2013}, nonparametrics in \cite{bick:chen:2009}. Works more specifically directed towards asymptotics of spectral clustering can be found in \cite{rohe:chat:yu:2011} and in \cite{lei:rina:2015}. Most of these works require a delicate asymptotic balancing  between the number of nodes, the degree of the nodes, and the number of communities. An example of a heterogeneous model which is analyzed rigorously from a statistical point of view is \cite{zhan:chen:2018}. For instance, the proposed modularity function is shown to be consistent in a heterogeneous stochastic block model framework. It is related to the  \cite{bick:chen:2009} paper.

A very important problem both in practice and in theory is the problem of determining the number of communities in community detection. In earlier literature this number was actually taken to be known. In statistical likelihood based models one has attempted to find this number by letting it be an unknown parameter in the likelihood and then do likelihood integration. \cite{wang:bick:2017} look at the problem from an underestimation and overestimation point of view. \cite{newm:rein:2016} propose replacing the original Bernoulli type likelihood by an approximated Poisson likelihood, which is easier to handle computationally. \cite{peix:2021} discusses AIC and BIC type approaches to this problem.

There has been made progress in the numerical estimation of the parameters in stochastic block type models. Typically, a Bayesian approach has been used with extensive use of Markov Chain Monte Carlo. But we think it is fair to say that the dimension of the networks attacked by stochastic block modeling has been considerably less than  the most general used algorithmic Skip-Gram models of Section \ref{Skip-Gram}.

\subsubsection{Time series modeling in networks\label{Time series network}}

A recent example of rigorous statistical modeling of a dynamic network is \cite{zhu:pan:li:liu:wang:2017}. They model the network structure by a network vector autoregressive model. This model assumes that the response of each node at a given time point is a linear combination of (a) its previous value, (b) the average of connected neighbors, (c) a set of node-specific covariates and (d) independent noise. More precisely, if $n$ is the network size, let $Y_{it}$ be the response collected from the $i$th subject (node) at time $t$. Further, assume that a $q$-dimensional node-specific random vector $Z_i = (Z_{i1},\ldots,Z_{iq})^{T} \in \mathbb{R}^{q}$ can be observed. Then the model for $Y_{it}$ is given by
\begin{equation}
Y_{it} = \beta_0+Z_i^{T} \gamma + \beta_1 n_i^{-1} \sum_{j=1}^{n} a_{ij}Y_{j,t-1} + \beta_2 Y_{i,t-1} + \varepsilon_{it}.
\label{57a}
\end{equation}
Here, $n_i = \sum_{j \neq i} a_{ij}$, $a_{ii}=0$, is the total number of neighbors of the node $v_i$ associated with $Y_i$, so it is the degree of $v_i$. The term $\beta_0+Z_i^{T} \gamma$ is the impact of covariates on node $v_i$, whereas $n_i^{-1} \sum_{j=1}^{n} a_{ij}Y_{j,t-1}$ is the average impact from the neighbors of $v_i$. The term $\beta_2 Y_{i,t-1}$ is the standard autoregressive impact. Finally the error term $\varepsilon_{it}$ is assumed to be independent of the covariates and iid normally distributed.

Given this framework, conditions for stationarity are obtained, and least squares estimates of parameters are derived and their asymptotic distribution found. 

They give an example analyzing a Sina Weibo data set, which is the largest twitter-like social medium in China. The data set contains weekly observations of $n=2{,}982$ active followers of an official Weibo account. 

An extension of the model (\ref{57a}) is contained in \cite{zhu:pan:2020}. 

There are a number of differences between the network vector
autoregression modeled by (\ref{57a}) and the dynamic network
embeddings treated in Section \ref{Dynamic}. First of all, (\ref{57a})
treats the dynamics of the nodes themselves and not of an
embedding. Even if the autoregressive model does introduce some
(stationary) dynamics in time, the parameters are static; i.e.\ no new
nodes are allowed, and the relationship between them is also static as
modeled by the matrix ${\bf A} = \{a_{ij}\}$. From this point of view, as the authors are fully aware of, the model (\ref{57a}) is not realistic for the dynamics that takes place in practice for many networks.
On the other hand the introduction of a stochastic model that can be analyzed by traditional methods of inference is to be lauded. A worthwhile next step is to try to combine more realistic models with a stochastic structure (regime type models for the parameters?) that is amenable to statistical inference.

For some very recent contributions to network autoregression, see \cite{armi:foki:krik:2022} and references therein.

\section{Embedding in 2 or 3 dimensions and visualization\label{Visualization}}

Visualization is an important part of data analysis. The problem can be stated as finding a good 2- or 3-dimensional representation of high dimensional data
and often with a large number of samples. Principal component analysis offers one possibility where the data are projected on the 2 or 3 first principal components. Although very useful, since it is linear and projects on a hyper plane, it generally fails to give a good characterization in cases where the data are concentrated on a nonlinear manifold which is a subset of $\mathbb{R}^{p}$.

It is appropriate to conclude this survey on embedding by the topic of visualization, where in principle any of the treated methods in this survey can be used by choosing the embedding dimension $m$ to be 2. However, we have chosen to concentrate on three methods that are powerful and much used, and which are based on the main ideas in Sections \ref{Nonlinear}, \ref{Topological} and \ref{Network}, respectively. The $t$-SNE
algorithm was developed by \cite{vand:hint:2008} and \cite{vand:2014}. It is based on ideas handling the connection between a high dimensional $x$-scale and a low dimensional $y$-scale which are inherent already in multidimensional scaling. But unlike most earlier attempts $t$-SNE is based on comparisons of {\it probability distributions} on the $x$ and $y$-scale, which seems much more sensible in a nonlinear problem than applying moments and covariances.

\cite{tang:liu:zhan:mei:2016} introduced LargeVis which is based on techniques reviewed in Section \ref{Network}, especially the Skip-Gram procedure treated in Section \ref{Skip-Gram main}. Finally, \cite{mcin:heal:melv:2018} use methods from topological data analysis akin to ideas in Section \ref{Topological} to derive their algorithm UMAP. Illustrations of the use of the three methods are given in Section \ref{Illustrating example}.


\subsection{$t$-SNE\label{SNE}}

SNE is an acronym for Stochastic Neighbor Embedding. That embedding and visualization technique was introduced by \cite{hint:rowe:2002}. 
The $t$ in $t$-SNE refers to further developments in \cite{vand:hint:2008} using a $t$-distribution approximation on the $y$-scale.

Starting with SNE, the similarities between the points on the $x$-scale and $y$-scale is sought expressed in terms of pairwise Gaussian approximations. On the $x$-scale high dimensional Euclidean distances are expressed in conditional probabilities. The similarity of a data point $X_i$ to a data point $X_j$ is expressed as a Gaussian conditional probability $p_{j|i}$ such that for pairs of nearby data points, $p_{j|i}$ would be relatively high, whereas for widely separated points, $p_{j|i}$ could be infinitesimally small. The essential idea is to preserve the internal structure of the high-dimensional data by keeping similar data points close and dissimilar data points far apart, in the low-dimensional space. Mathematically $p_{j|i}$ is given by
\begin{equation}
p_{j|i} = p_{j|i}(x_j|x_i) = \frac{\exp(-||x_j-x_i||^2/2\sigma_i^2)}{\sum_{k \neq i} \exp(-||x_k-x_i||^2/2\sigma_i^2)}, 
\label{61a}
\end{equation}
where $\sigma_i^2$ is the variance of the Gaussian that is centered on the data point $x_i$. The parameter $\sigma_i$ is chosen so that the probability distribution $P_i$, induced by $p_{j|i}$ for all $j$-s different from $i$, has a perplexity specified by the user. Here the perplexity of $P_i$ is given by
$$
\mbox{Perp}_i = 2^{-\sum_{j} p_{j|i} \log_{2} p_{j|i}}.
$$
See \cite{hint:rowe:2002} for more details.

The similarities on the $x$-scale is sought mapped into corresponding similarities in the low dimensional $y$-scale by modeling the  conditional probabilities by
$$
q_{j|i} = \frac{\exp(-||y_j-y_i||^2)}{\sum_{k \neq i} \exp(-||y_k-y_i||^2)}.
$$
The coordinates $Y_i$ of a data point $X_i, i=1,\ldots,n$ are then sought determined by minimizing the Kullback-Leibler distance (or cross entropy) between the $p_{j|i}$ and $q_{j|i}$, i.e.\ by minimizing the cost function
$$
C = \sum_i \mbox{KL}(P_i||Q_i) = \sum_{i,j} p_{j|i} \log \frac{p_{j|i}}{q_{j|i}}.
$$
The minimization of the cost function with respect to the $y$-coordinates can be done by using a gradient descent method, and the $y$-s are initialized by random, Gaussian values.

The SNE algorithm is hampered by a cost function which is quite difficult to optimize in practice, and there is a so-called ``crowding'' problem in the sense that far apart points on the $x$-scale may be mapped in such a way that the joint probability $q_{ij}$ may be even smaller than $p_{ij}$. These problems are attacked in $t$-SNE by symmetrization, modeling {\it joint} probabilities $p_{ij}$ and $q_{ij}$ and by using a $t$-distribution as an approximation at the $y$-scale having points in the tails mapped such that $q_{ij}$ is larger than $p_{ij}$ to avoid the crowding effect. This trick is also present for other local techniques for multidimensional scaling.

To avoid problems that may be caused by outliers on the $x$-scale the ``joint probabilities'' on the $x$-scale are in fact computed as $p_{ij}= (p_{i|j}+p_{j|i})/2n$, which ensures $\sum_j p_{ij} > 1/2n$ for all data points $X_i$, such that each data point makes a significant contribution to the cost function. Further,  on the $y$-scale a $t$ distribution structure of one degree of freedom is used,
$$
q_{ij} = \frac{(1+||y_i-y_j||^2)^{-1}}{\sum_{k \neq \ell} (1+||y_k-y_{\ell}||^2)^{-1}},
$$
where it should be noted that a double sum is now used in the denominator. The cost function is given by
$$
C =  \sum_{i,j} p_{ij}\log \frac{p_{ij}}{q_{ij}}.
$$
The details of the optimization can again be found in \cite{vand:hint:2008}. In that paper there is also a series of experiments comparing $t$-SNE with the Sammon mapping of MDS and the ISOMAP and LLE, where the $t$-SNE does extremely well. 

The $t$-SNE algorithm is speeded up in the paper by \cite{vand:2014} by not going over all possible pairs $(x_i,x_j)$ but only essentially over nearest neighbors. 

\subsection{LargeVis\label{LargeVis}}


\cite{tang:liu:zhan:mei:2016} propose a new algorithm for visualization, LargeVis. It starts with a speeded up {\it approximate} nearest neighbor algorithm that has complexity $O(n)$ as compared to $O(n\log n)$ for the speeded up nearest neighbor algorithms of \cite{vand:2014}. The \cite{tang:liu:zhan:mei:2016} algorithm is built upon random projection trees but significantly improved by using neighbor exploring. The basic idea of this, similarly to the LINE construct in \cite{tang:qu:wang:zhan:yan:mei:2015b} and referenced in Section \ref{Neighborhood sampling}, is that ``the neighbor of my neighbor is also likely to be my neighbor''. Specifically, a few random projection trees are built to construct an approximate k-nearest neighbor graph, the accuracy of which may not be so high. Then for each node of the graph, the neighbors of its neighbor are searched, which are also likely to be candidates of its nearest neighbor. The accuracy may then be improved by multiple iterations. The claim is that the accuracy of this k-nearest neighbor  graph quickly improves to almost 100\% without investing in many trees. For the weights of the nearest neighbor graph essentially the same procedure as in $t$-SNE is used. The graph is symmetrized by setting the weights between $x_i$ and $x_j$ to $w_{ij} = \frac{p_{j|i}+p_{i|j}}{2n}$, where $p_{i|j}$ and $p_{j|i}$ are defined via (\ref{61a}).
Before using the LargeVis algorithm itself a pre-processing step can be used where the dimension is reduced to say 100 by using the Skip-Gram network embedding technique explained in Section \ref{Skip-Gram main}. The negative sampling technique of \cite{miko:suts:chen:corr:dean:2013b} is used in the Skip-Gram step. 

For the time complexity of the optimization, done with asynchronous
stochastic gradient descent, each stochastic gradient step takes
$O(sM)$, where $M$ is the number of negative samples, say $M$ is from $5 - 10$, and $s$ is the number of dimensions of the low dimensional space, $s=2,3$. Therefore the overall complexity is $O(sMn)$, which is linear in the number of nodes.


\subsection{UMAP\label{UMAP}}

Sections \ref{Manifold} and \ref{Persistent} were concerned with
topological methods in manifold learning and persistence homology. In
particular, filters  of simplicial complexes were used in Section 1.2
of the Supplement \citep{supp}. In the first part of \cite{mcin:heal:melv:2018}, these filters are generalized to simplicial sets. In addition, components of fuzzy set theory, category theory and functor theory are used to compute a fuzzy topological representations. 

Letting $\{Y_1,\ldots, Y_n\} \subseteq \mathbb{R}^m$ and $\{X_1,\ldots,X_n\} \subseteq \mathbb{R}^{p}$ with $m \ll p$, in visualization we have a situation where $m$ is 2 or 3. 

To compare two fuzzy sets generated by $\{X_1,\ldots,X_n\}$ and $\{Y_1,\ldots,Y_n\}$, respectively, fuzzy set cross entropy is used in UMAP. The use of advanced concepts of algebraic topology makes the first part of this paper hard to read. In the computational part of the paper, however, inspired by motivations and ideas of the first part, the authors specialize to a $k$-neighborhood graph situation where the analogy with $t$-SNE and LargeVis is easier to appreciate.

As with other $k$-neighbor graph based algorithms, UMAP, can be described in two phases. In the first phase a particular weighted $k$-neighbor graph is constructed. In the second phase a low dimensional layout of this graph is made. The theoretical basis for UMAP in the first part of \cite{mcin:heal:melv:2018} provides novel approaches to both of these phases.  

Let $\{X_1,\ldots,X_n\}$ be the input data set with a jointly given matrix ${\bf D}$ that can be thought of as consisting of Euclidean distances between the data vectors. For each $X_i$ one can compute the set of $k$ nearest neighbors $\{X_{i_{1}},\ldots,X_{i_{k}}\}$. There are many choices of a nearest neighbor algorithm. \cite{mcin:heal:melv:2018} use the algorithm of \cite{dong:mose:li:2018}.

This can be used to define a weighted directed graph $G^{\prime} = (V,E,w)$. The nodes of $G^{\prime}$ are the set $\{X_1,\ldots,X_n\}$ the directed edges are $\{(X_i,X_{i_{j}})| 1 \leq j \leq k, 1 \leq i \leq n\}$ and a weight function defined in \cite{mcin:heal:melv:2018}. Let ${\bf A}$ be the weighted adjacency matrix of $G^{\prime}$. An undirected graph $G$ is obtained by introducing the symmetric adjacency matrix
$$
{\bf B} = {\bf A}+{\bf A}^{T}-{\bf A} \circ {\bf A}^{T},
$$ 
where $\circ$ denotes the Hadamard (pointwise) product.

The $\{X_1,\ldots,X_n\}$ data set is next connected to a low dimensional data set $\{Y_1,\ldots,Y_n\}$, where the dimension is 2 or 3 if visualization is considered.
The transition from $\{X_1,\ldots,X_n\}$ to $\{Y_1,\ldots,Y_n\}$ is accomplished by a force directed graph  
layout algorithm. The history of this kind of graph layout goes far back, 
\cite{tutt:1963}. A more recent account can be found in \cite{kobo:2012}.
The details of the algorithm as used in UMAP with an iterative
application of attractive and repulsive forces are given in
\citet[p. 14]{mcin:heal:melv:2018}. It should be noted that the
terminology of attractive and repulsive forces is used in
\cite{vand:hint:2008} as well, but unlike their paper where there is a
random set-like initialization, in UMAP a spectral layout (cf.\
Sections \ref{Laplace eigenmaps} and \ref{Spectral}) is used to
initialize the embedding. This is claimed to provide faster
convergence and greater stability within the algorithm. It should be
noted that also for the implementation of their algorithm negative
sampling, as treated in Section \ref{Skip-Gram
  main}, plays an important role in reducing the computational burden.

\subsubsection{The importance of initialization of $t$-SNE and UMAP}

One noteworthy difference between $t$-SNE and UMAP is the initialization: The embeddings of $t$-SNE are, at least as the default choice, initialized randomly, while the embeddings of UMAP are initialized by Laplacian eigenmaps. According to recent experiments by \cite{kobak:linderman:2021}, UMAP with random initialization seems to preserve the global structure as bad as $t$-SNE with random initialization,
while $t$-SNE with ``informative initialization''  (PCA in this case)  performs as
well as UMAP with informative initialization. \citeauthor{kobak:linderman:2021} argue therefore that 1) the UMAP algorithm per se does not have any advantage over $t$-SNE when it comes to preservation of the global structure, and 2) these algorithms should by default use informative initialization. For the statistician, this informative initialization might be handled in a more formal way, for example expressed by appropriate priors in the Bayesian paradigm. More formal approaches to questions as these are warranted as, e.g., in biology, reproducibility of such embeddings is essential \citep{becht:mcinnes:healy:duterte:kwok:ng:ginhoux:newell:2019}.

\subsection{A brief comparison of $t$-SNE, LargeVis and UMAP}

A number of experiments were performed in \cite{mcin:heal:melv:2018}
with a comparison to $t$-SNE and LargeVis. The UMAP works on par with
or better than these algorithms for those examples. 

All of the embedding algorithms have been demonstrated to work well in a number of quite complicated situations. Nevertheless, as pointed out by \citeauthor{mcin:heal:melv:2018}, it is important to be aware of some weaknesses of these algorithms that could create fruitful challenges for further research. 

$t$-SNE, LargeVis and UMAP all lack the strong interpretability of PCA and it is difficult to see that something like a factor analysis can be performed.

One of the core assumptions is that it is assumed that there exists a lower dimensional manifold structure in the data. If this is not so, there is always the danger that a spurious noise driven embedding can be the result. This danger is reduced as the sample size increases.  Developing an asymptotic analysis and finding more robust algorithms is clearly a challenge.

For all three algorithms a number of approximations are made, such as the use of approximate nearest neighbor algorithms and negative sampling used in optimization. Particularly for small sample sets the effect of these approximations may be non-negligible.

\subsection{An illustrating example\label{Illustrating example}}

\begin{figure}[ht!]
  \begin{center}
    \begin{subfigure}{1\textwidth}
      \centering
      \includegraphics[width=1\textwidth]{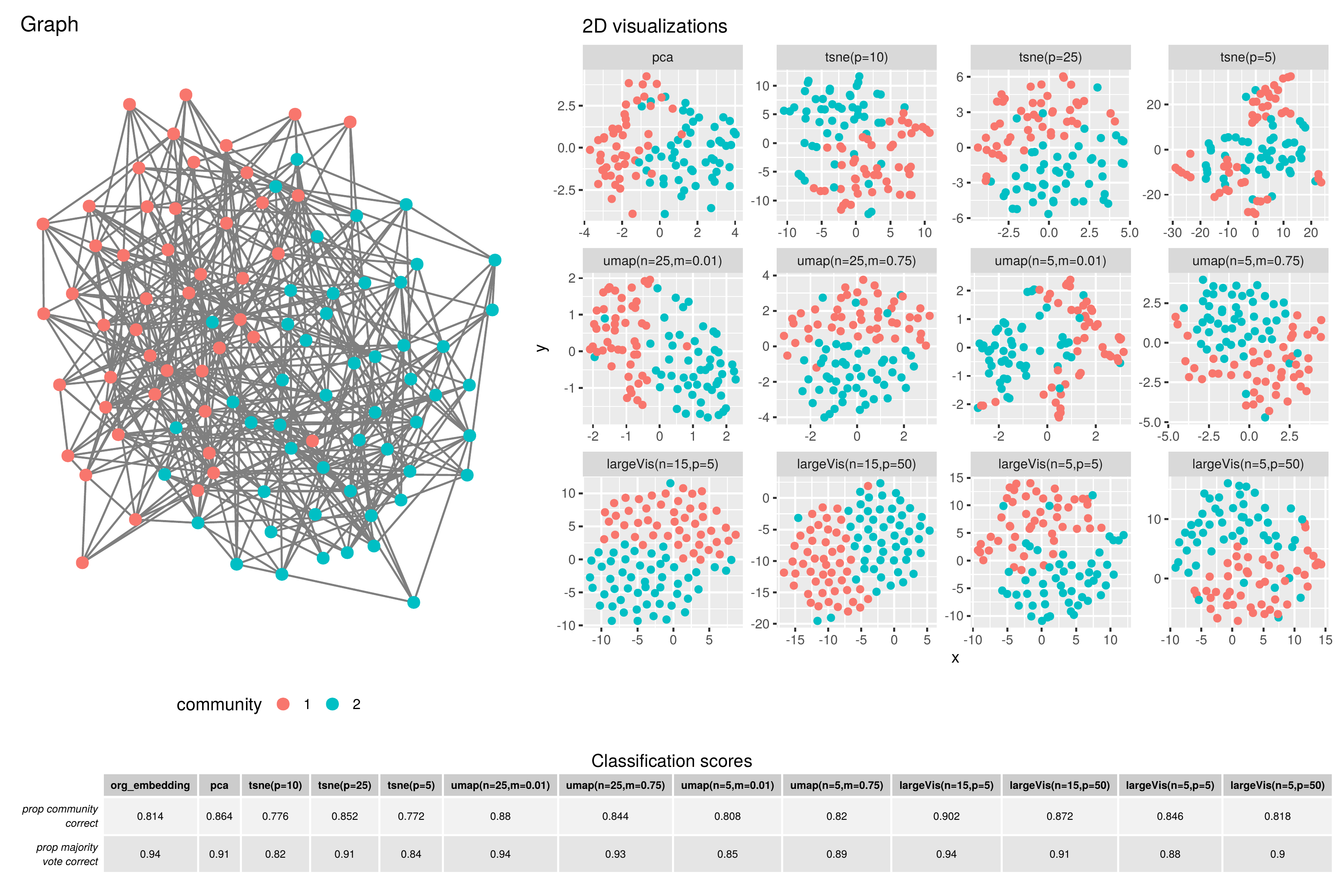}
      \caption{Homogeneous graph from stochastic block model.}
      \label{fig:sub5a}
    \end{subfigure}
    \begin{subfigure}{1\textwidth}
      \centering
    \includegraphics[width=1\textwidth]{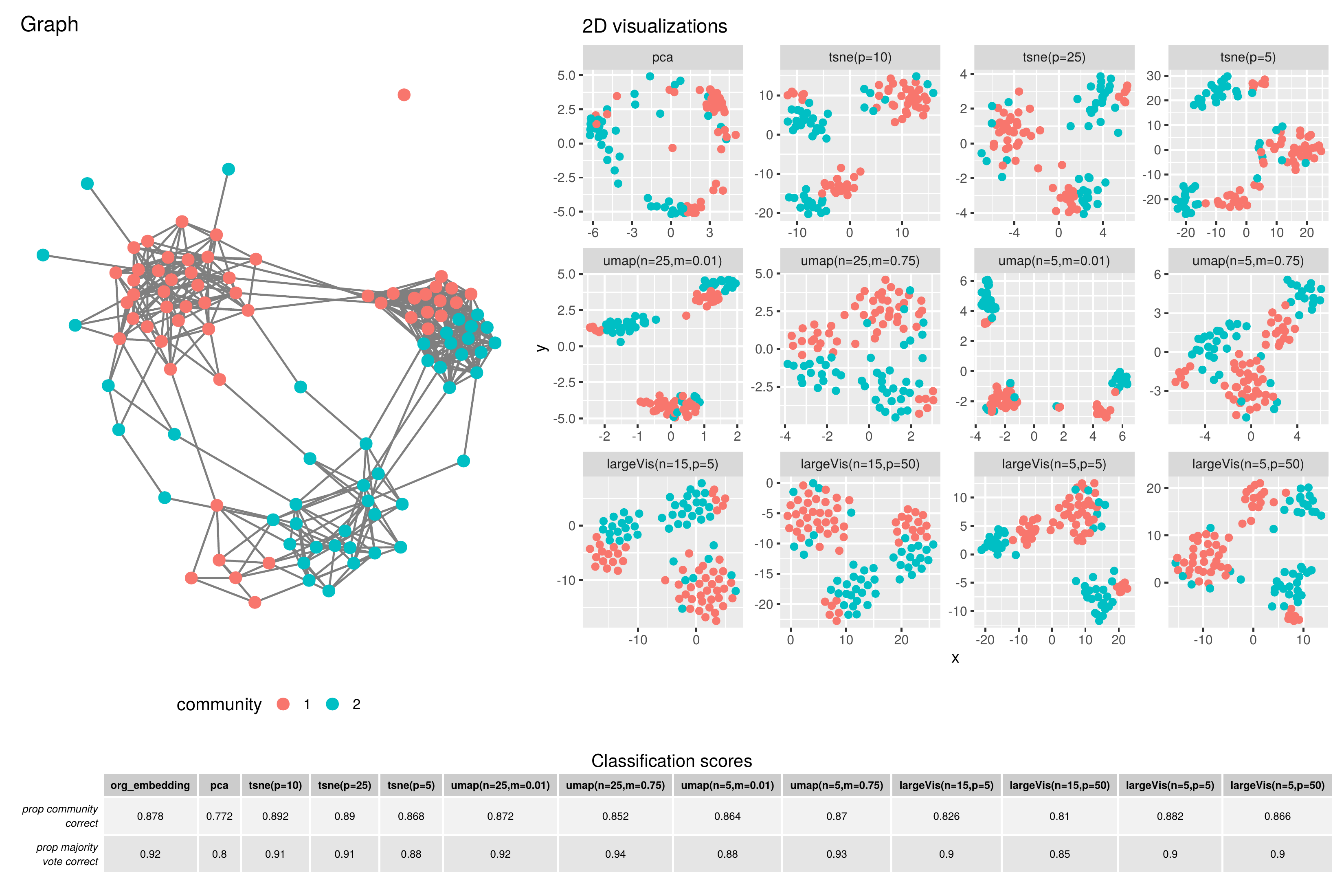}  
      \caption{Heterogeneous graph from a combination of three stochastic block models.}
      \label{fig:sub5b}
    \end{subfigure}    
\end{center}
\caption{Graphs, visualizations and classification results with a $k$-nearest neighbors algorithm with $k=5$.
 \label{fig:5}}
\end{figure}

The illustrating example consists of two networks, each having two different types of nodes (colored red and blue, respectively) corresponding to two different communities. The first one, the homogeneous graph in Fig.~\ref{fig:sub5a}, is very simple and  is simulated from a stochastic block model \citep{karr:newm:2011}, mentioned in Section \ref{SBM1}, with 2 communities, 100 nodes, average node degree $d = 10$, and ratio of between-community
edges over within-community edges $\beta=0.4$. In this setup the the number of edges per node is Poisson
distributed with expected number of edges of 10. This simple network has very little overlap between the two types of nodes.

\afterpage{\clearpage}

The second one is somewhat more complex,
the heterogeneous graph in Fig.~\ref{fig:sub5b}, and is simulated from three
subgraphs {\bf a}, {\bf b} and {\bf c}, that has 2 communities each:
\begin{description}
\item[Graph a:] 30 nodes, average node degree $d = 7$, ratio of
  between-block edges over within-block edges $\beta=0.2$
\item[Graph b:] 30 nodes, average node degree $d = 15$, ratio of
  between-block edges over within-block edges $\beta=0.4$
\item[Graph c:]  40 nodes, average node degree $d = 7$, ratio of
  between-block edges over within-block edges $\beta=0.2$, and an
  unbalanced community proportion; a probability of 3/4 for community 1
  and a probability of 1/4 for community 2
\end{description}
To link graphs a, b and c, some random edges are added between nodes from
the same community\footnote{For each pair of nodes between a pair of graphs, say Graph {\bf a} and
  {\bf c}, a new link is randomly sampled with a probability of 0.01, and
   links  connecting two nodes from the same community are kept.}.

The purpose of the illustrating example is to examine how well these network structures are managed by $t$-SNE, LargeVis and UMAP, how robust they are to parameter choices inherent in the three methods, and how they compare with traditional principal component analysis (PCA) visualization.

The visualization is done in two steps. First the networks are embedded in $\mathbb{R}^ m$ with $m=64$ using the Skip-Gram routine node2vec
with (cf.\ Section \ref{Neighborhood sampling})
$L = 30$ nodes in each random walk and $\gamma = 200$ walks per node, and a
word2vec window length of $K=5$ where all nodes are included. The second step is to reduce the point cloud in $\mathbb{R}^ {64}$ to $\mathbb{R}^ 2$, i.e., the visualization step using PCA and the three visualization algorithms with a selection of different tuning parameters. (In $t$-SNE, $p$ is the perplexity parameter; in LargeVis $n$ is the number of negative samples, $p$ the total weight of positive interactions; in UMAP $n$ is the number of nearest-neighbors, $m$ is a distance parameter, where low $m$ gives clumpier embeddings.) The results are given in Figs.~\ref{fig:sub5a} and \ref{fig:sub5b}.

Underneath the figures are given classification scores for the two types of nodes (communities) in the study. These are classified on a neighborhood basis. In the first line of each sub-table the class of a node is determined using the average of the 5 nearest neighbors; in the second by the majority vote among these 5 nearest neighbors. The first column ``org\_embedding'' gives the classification results for the 64-dimensional embedding in step 1.

For the simple network, PCA does well, on par with the three other
visualization algorithms, both visually and in the classification. The
tuning parameters does not seem to make much of a difference with the
exception of $t$-SNE with $p=5$. For the more complicated network, PCA
is in trouble both visually and with respect to classification. In
this case the dependence on tuning parameters seems to be greater, but
most of the visualizations manage to pick out the three subgraphs
${\bf a}$, ${\bf b}$, and ${\bf c}$. For all values of the tuning
parameters $t$-SNE, LargeVis and UMAP all do clearly better than
PCA. Somewhat surprisingly, perhaps, the embedding in 64 dimensions
gives result not very different from those of the three visualizations
routines. We also did experiments with other embedding dimensions
ranging from 2 to 256. Again the classification results were not much
different. This could be due to the fact that the number of nodes and
links in these experiments are very modest compared to the real data
experiments in the Skip-Gram references given in Sections
\ref{Skip-Gram} and \ref{Neighborhood sampling}, which has number of
nodes and links of an entirely different order. A more involved
illustrating example (but still with a moderate number of nodes) is
given in Section 3 of the Supplement \citep{supp}.



\section{Some concluding remarks \label{Conclusion}}

Principal components work well for linearly generated Gaussian data. It may also work well for other types of data and is probably still the most important statistical embedding method. But, on the other hand, it is not difficult to find examples where it does not work. The search for nonlinear extensions started long ago with the MDS method. In fact, multidimensional scaling methods contain ideas that have been found relevant in several recent nonlinear algorithms.

There is no universally superior method that works better than any of the others in all situations. For Gaussian or approximately Gaussian data ordinary principal components should be preferred. If the distribution can be approximated locally by a Gaussian, the potential of locally Gaussian methods as outlined in \cite{tjos:otne:stov:2021a} could be investigated. Other nonlinear methods depend on local linear structures in the data. For data sets with holes or cavities, topological data analysis is a natural option. Data that form a network has artificial neural network methods as an obvious candidate. The Skip-Gram method of Section \ref{Skip-Gram} is based on a single layer artificial network. Deep learning algorithms are based on multiple layer neural networks and is an attractive alternative for more complicated dependencies. The neural network approaches have an advantage in their speed, making it possible to treat ultra-high dimensional data sets with complex relationships. They avoid the problem of solving an eigenvalue problem of very high dimension present in a number of other methods.  

The field of statistical embedding has had an explosive development in the recent years, not the least because of the need to interpret, represent, cluster and classify very large data sets, this  being an important  part of the Big Data revolution. This may be particularly true for the embedding of networks, since an increasing part of ultra-large data sets comes in the form of networks, such networks being of importance in an increasing number of applications. 

In this paper we have covered selected methods of nonlinear embedding generalizing PCA, topological embeddings in persistence diagrams, network embedding and embedding to dimension 2 (i.e., visualization). In addition, in the  course of the review, we have pointed to some cases of an apparent and arguably widening gap between developments in data science, including computer and algorithmic based methods, and more traditional statistical modeling methods. We have also sought to point out specific issues that could benefit from more input from statisticians. These may be conveniently summed up in the following keypoints:
\begin{itemize}
\item [1.] In quite a few algorithms there are parameters to be chosen, and the performance of the algorithm may depend quite strongly on these choices. Examples can be found in Skip-Gram, spectral community detection, the Mapper, and there are others. There is a need for well-founded methods for making in some sense optimal or near optimal  choices of such parameters -- in some cases as an alternative to the computational expensive empirical optimization routines which typically also have a randomness component.
As mentioned in Section \ref{Data science}, information criteria based solution is one option, in particular likelihood-free methods like GIC might be one way to go about this.
\item[2.] It is highly desirable to reduce the gap between machine learning algorithmic techniques and statistical modeling. A good example of a bridging attempt is the stochastic block models for which one can do statistical inference and which has also  resulted in good network algorithms. One needs more of this!
\item[3.] More critical statistical work is needed to test the sanity and robustness of algorithms. One example is the close investigation of the modularity algorithm reported on in Section \ref{SBM1}. It is useful to put algorithms to  stress tests, but it is important to find a balancing point between such  criticism and perceived usefulness of an algorithm.
\end{itemize}
It is important, however, to point out that this is a two-way relationship. 
We are hopeful that interaction between machine learning and statistical modeling could bring about synergy effects for both disciplines. 


\section*{Funding}

This work was supported by the Norwegian Research Council grant 237718 (BigInsight).

\section*{Supplementary material}
The Supplement  \citep{supp} contains more details on persistence
  diagrams, simplical complexes and word embedding, as well as a more
involved variant of the network example in Section \ref{Illustrating example}.

\bibliographystyle{apalike}
\bibliography{dagemb}

\begin{thebibliography}{}

\bibitem[Aizerman et~al., 1956]{aize:brav:rozo:1964}
Aizerman, M.~A., Braverman, E., and Rozonoer, L. (1956).
\newblock Theoretical foundations of the potential function method in pattern
  recognition learning.
\newblock {\em Automation and Remote Control}, 25:821--137.

\bibitem[Armillotta et~al., 2022]{armi:foki:krik:2022}
Armillotta, M., Fokianos, K., and Krikidis, I. (2022).
\newblock Generalized linear models network autoregression.
\newblock In {\em Network Science}, pages 112--125.
\newblock International Conference on Network Science.

\bibitem[Becht et~al.,
  2019]{becht:mcinnes:healy:duterte:kwok:ng:ginhoux:newell:2019}
Becht, E., McInnes, L., Healy, J., Dutertre, C.-A., Kwok, I.~W., Ng, L.~G.,
  Ginhoux, F., and Newell, E.~W. (2019).
\newblock Dimensionality reduction for visualizing single-cell data using umap.
\newblock {\em Nature biotechnology}, 37(1):38--44.

\bibitem[Belkin and Niyogi, 2002]{belk:niyo:2002}
Belkin, M. and Niyogi, P. (2002).
\newblock Laplacian eigenmaps and spectral techniques for embedding and
  clustering.
\newblock In {\em Advances in Information Processing Systems}. MIT Press,
  Cambridge.
\newblock T.K. Leen and T.G. Dietterich and V. Treps eds.

\bibitem[Belkin and Niyogi, 2003]{belk:niyo:2003}
Belkin, M. and Niyogi, P. (2003).
\newblock Laplacian eigenmaps for dimensionality reduction and data
  representation.
\newblock {\em Neural Computation}, 15:1373--1396.

\bibitem[Bian et~al., 2019]{bian:koh:dobb:divo:2019}
Bian, R., Koh, Y.~S., Dobbie, G., and Divoli, A. (2019).
\newblock Network embedding and change modeling in dynamic heterogeneous
  networks.
\newblock In {\em Proceedings of the 42nd International ACM SIGIR Conference on
  Research and Development in Information Retrieval}, pages 861--864.

\bibitem[Bickel and Chen, 2009]{bick:chen:2009}
Bickel, P. and Chen, A. (2009).
\newblock A nonparametric view of network models and {N}ewman-{G}irvan and
  other modularities.
\newblock {\em Proceedings National Academy of Science. PNAS 0907096106}.

\bibitem[Bickel et~al., 2013]{bick:choi:chan:zhan:2013}
Bickel, P., Choi, D., Chang, X., and Zhang, H. (2013).
\newblock Asymptotic normality of maximum likelihood and its variational
  approximation for stochastic blockmodels.
\newblock {\em Annals of Statistics}, 41:1922--1946.

\bibitem[Bickel and Sarkar, 2016]{bick:sark:2016}
Bickel, P. and Sarkar, P. (2016).
\newblock Hypothesis testing for automated community detection in networks.
\newblock {\em Royal Statistical Society, Ser. B}, 78:253--273.

\bibitem[Blondel et~al., 2008]{blon:guil:lamb:lefe:2008}
Blondel, V.~D., Guillaume, J.-L., Lambiotte, R., and Lefebvre, E. (2008).
\newblock Fast unfolding of communities in large networks.
\newblock {\em Journal of statistical mechanics: theory and experiment},
  2008(10):P10008.

\bibitem[Boser et~al., 1992]{bose:guyo:vapn:1992}
Boser, B., Guyon, I., and Vapnik, V. (1992).
\newblock A training algorithm for optimal margin classifiers.
\newblock In Fifth Annual Workshop on COLT, Pittsburgh, ACM.

\bibitem[Budur et~al., 2015]{budu:lee:kong:2015}
Budur, E., Lee, S., and Peti, K. (2015).
\newblock Structural analysis of criminal network and predicting hidden links
  using machine learning.
\newblock arXiv:1507.05739v3.

\bibitem[Bukkuri et~al., 2021]{bukk:ando:darc:2021}
Bukkuri, A., Andor, N., and Darcy, I. (2021).
\newblock Applications of topological data analysis on oncology.
\newblock {\em Frontiers in Artificial Intelligence: Machine Learning and
  Artificial Intelligence}, 4:1--14.

\bibitem[Cannings and Samworth, 2017]{cann:samw:2017}
Cannings, T. and Samworth, R. (2017).
\newblock Random projection ensemble classification.
\newblock {\em Journal of the Royal Statistical Society, Ser. B}, 79:959--1035.

\bibitem[Carlsson, 2009]{carl:2009}
Carlsson, G. (2009).
\newblock Topology and data.
\newblock {\em Bulletin of the American Mathematical Society}, 46:255--308.

\bibitem[Carri\`{e}re et~al., 2018]{carr:mich:oudo:2018}
Carri\`{e}re, M., Michel, B., and Oudot, S. (2018).
\newblock Statistical analysis and parameter selection for mapper.
\newblock {\em Journal of Machine Learning Research}, 19:1--39.

\bibitem[Carri\`{e}re and Rabad\'{a}n, 2020]{carr:raba:2020}
Carri\`{e}re, M. and Rabad\'{a}n, R. (2020).
\newblock Topological data analysis of single-cell hi-c contact maps.
\newblock In {\em Topological Data Analysis}, pages 147--162. Springer, New
  York.
\newblock Abel Symposia Booke Series, ABEL, volume 15.

\bibitem[Chazal and Michel, 2017]{chaz:mich:2017}
Chazal, F. and Michel, B. (2017).
\newblock An introduction to topological data analysis: fundamental and
  practical aspects for data scientists.
\newblock arXiv: 1710.04019v1.

\bibitem[Chazal and Michel, 2021]{chaz:mich:2021}
Chazal, F. and Michel, B. (2021).
\newblock An introduction to topological data analysis: fundamental and
  practical aspects for data scientists.
\newblock {\em Frontiers in Artificial Intelligence: Machine Learning and
  Artificial Intelligence}, 4:1--28.

\bibitem[Chen et~al., 2015a]{chen:geno:wass:2015c}
Chen, Y., Genovese, C., and Wasserman, L. (2015a).
\newblock Asymptotic theory for density ridges.
\newblock {\em Annals of Statistics}, 43:1896--1928.

\bibitem[Chen et~al., 2015b]{chen:ho:free:geno:wass:2015a}
Chen, Y., Ho, S., Freemen, P., Genovese, C., and Wasserman, L. (2015b).
\newblock Cosmic web reconstruction through density ridges: methods and
  algorithm.
\newblock {\em Monthly Notices of the Royal Astronomical Society},
  454:1140--1156.

\bibitem[Chen et~al., 2015c]{chen:ho:tenn:mand:crof:2015b}
Chen, Y., Ho, S., Tenneti, A., Mandelbaum, R., andT. DiMatteo, R.~C., Freeman,
  P., Genovese, C., and Wasserman, L. (2015c).
\newblock Investigating galaxy-filament alignments in hydrodynamic simulations
  using density ridges.
\newblock {\em Monthly Notices of the Royal Astronomical Society},
  454:3341--3350.

\bibitem[Claeskens et~al., 2008]{claeskens2008information}
Claeskens, G., Croux, C., and Van~Kerckhoven, J. (2008).
\newblock An information criterion for variable selection in support vector
  machines.
\newblock {\em The Journal of Machine Learning Research}, 9:541--558.

\bibitem[Coifman and Lafon, 2006]{coif:lafo:2006}
Coifman, R. and Lafon, S. (2006).
\newblock Diffusion maps.
\newblock {\em Applied and Computational Harmonic Analysis}, 21:5--30.

\bibitem[Cormen et~al., 2022]{cormen:leisersion:rivest:stein:2022}
Cormen, T.~H., Leiserson, C.~E., Rivest, R.~L., and Stein, C. (2022).
\newblock {\em Introduction to algorithms}.
\newblock MIT press.

\bibitem[Crawford et~al., 2020]{craw:mono:chen:mukh:raba:2020}
Crawford, L., Monod, A., Chen, A., Mukherjee, S., and Rabad\'{a}n, R. (2020).
\newblock Predicting clinical outcomes in glioblastoma: an application of
  topological and functional data analysis.
\newblock {\em Journal of the American Statistical Association},
  115:1139--1150.

\bibitem[Cui et~al., 2019]{cui:wang:pei:zhu:2019}
Cui, P., Wang, X., Pei, J., and Zhu, W. (2019).
\newblock A survey on network embedding.
\newblock {\em IEEE transactions and Knowledge Engineering}, 31:833--852.

\bibitem[de~Silva and Tenenbaum, 2002]{desi:tene:2002}
de~Silva, V. and Tenenbaum, J. (2002).
\newblock Global versus local methods in nonlinear dimensionality reduction.
\newblock {\em Advances in neural information processing systems}, 15.

\bibitem[Devroye and Wise, 1980]{devr:wise:1980}
Devroye, L. and Wise, G. (1980).
\newblock Detection of anormal behavior via nonparametric estimation of the
  support.
\newblock {\em Siam Journal of Applied Mathematics}, 38:480--488.

\bibitem[Dong et~al., 2018]{dong:mose:li:2018}
Dong, W., Moses, C., and Li, K. (2018).
\newblock Efficient $k$-nearest neighbour graph construction for generic
  similarity measures.
\newblock In Proceedings of the 20th International Conference of the World Wide
  Web, pp. 577-586, New York.

\bibitem[Dong et~al., 2017]{dong:chaw:swam:2017}
Dong, Y., Chawla, N., and Swami, A. (2017).
\newblock Metapath2vec: Scalable representation learning for heterogeneous
  networks.
\newblock Kid 17, 2017, Halifax, NS, Canada.

\bibitem[Du et~al., 2018]{du:wang:song:lu:wang:2018}
Du, L., Wang, Y., Song, G., Lu, Z., and Wang, J. (2018).
\newblock Dynamic network embedding: An extended approach for skip-gram based
  network embedding.
\newblock Proceedings of the 27th International joint conference on Artificial
  Intelligence, IJ(AI-18).

\bibitem[Duchamp and Stuetzle, 1996]{duch:stue:1996}
Duchamp, T. and Stuetzle, W. (1996).
\newblock Extremal properties of principal curves.
\newblock {\em Annals of Statistics}, 24:1511--1520.

\bibitem[Edelsbrunner et~al., 2002]{edel:letc:zomo:2002}
Edelsbrunner, H., Letcher, D., and Zomorodian, A. (2002).
\newblock Topological persistence and simplification.
\newblock {\em Discrete Computational Geometry}, 28:511--533.

\bibitem[Genovese et~al., 2012]{geno:pero:verd:wass:2012}
Genovese, C., Perone-Pacifico, M., Verdinelli, I., and Wasserman, L. (2012).
\newblock Manifold estimation and singular deconvolution under {H}ausdorff
  loss.
\newblock {\em Annals of Statistics}, 40:941--963.

\bibitem[Genovese et~al., 2014]{geno:pero:verd:wass:2014}
Genovese, C., Perone-Pacifico, M., Verdinelli, I., and Wasserman, L. (2014).
\newblock Nonparametric ridge estimation.
\newblock {\em Annals of Statistics}, 42:1511--1545.

\bibitem[Ghojogh et~al., 2021]{ghoj:ghod:karr:crow:2021}
Ghojogh, B., Ghodsi, A., Karray, F., and Crowley, M. (2021).
\newblock Johnson-lindenstrauss lemma, linear and nonlinear random projections,
  random fourier features and random kitchen sinks: Tutorial and survey.
\newblock arXiv:2108.04172v1.

\bibitem[Ghrist, 2017]{ghri:2017}
Ghrist, R. (2017).
\newblock Homological algebra and data.
\newblock Lecture notes.

\bibitem[Girvan and Newman, 2002]{girv:newm:2002}
Girvan, M. and Newman, M. (2002).
\newblock Community structure in social and biological networks.
\newblock {\em Proc. Natl. Acad. Sci. (PNAS)}, 99:7821--7826.

\bibitem[Greene and Cunningham, 2011]{gree:doyl:cunn:2011}
Greene, D. and Cunningham, P. (2011).
\newblock Tracking the evolution of communities in dynamic social networks.
\newblock Report Idiro Technologies, Dublin, Ireland.

\bibitem[Gretton, 2019]{gret:2019}
Gretton, A. (2019).
\newblock Introduction to {R}{K}{H}{S}, and some simple kernel algorithms.
\newblock Lecture notes.

\bibitem[Grover and Leskovec, 2016]{grov:lesk:2016}
Grover, A. and Leskovec, J. (2016).
\newblock node2vec: Scalable feature learning for networks.
\newblock In {\em Proceedings of the 22nd ACM SIGKDD international conference
  on Knowledge discovery and data mining}, pages 855--864.

\bibitem[Haghverdi et~al., 2015]{hagh:buet:thei:2015}
Haghverdi, L., Buettner, F., and Theis, F. (2015).
\newblock Diffusion maps for high-dimensional single-cell analysis of
  differentiation data.
\newblock {\em Bioinformatics}, 31:2989--2998.

\bibitem[Hastie, 1984]{hast:1984}
Hastie, T. (1984).
\newblock Principal curves and surfaces.
\newblock Laboratory for Computational Statistics Technical Report 11, Stanford
  University, Department of Statistics.

\bibitem[Hastie and Stuetzle, 1989]{hast:stue:1989}
Hastie, T. and Stuetzle, W. (1989).
\newblock Principal curves.
\newblock {\em Journal of the American Statistical Association}, 84:502--516.

\bibitem[Hastie et~al., 2019]{hast:tibs:frie:2009}
Hastie, T., Tibshirani, R., and Friedman, J. (2019).
\newblock {\em The Elements of Statistical Learning}.
\newblock Springer, New York.

\bibitem[Hinton and Roweis, 2002]{hint:rowe:2002}
Hinton, G. and Roweis, S. (2002).
\newblock Stochastic neighbour embedding.
\newblock {\em Advances in Neural Information Processing Systems}, 15:833--840.

\bibitem[Hinton and Salakhutdinov, 2006]{hint:sala:2006}
Hinton, G. and Salakhutdinov, R. (2006).
\newblock Reducing the dimensionality of data with neural networks.
\newblock {\em Science}, 313:504--507.

\bibitem[Hoff et~al., 2002]{hoff:raft:hand:2002}
Hoff, P., Raftery, A., and Handcock, M. (2002).
\newblock Latent space approaches to social network analysis.
\newblock {\em Journal of the American Statistical Association}, 97:1090--1098.

\bibitem[Holland et~al., 1983]{holl:lask:lein:1983}
Holland, P., Laskey, K., and Leinhardt, S. (1983).
\newblock Stochastic blockmodels: First steps.
\newblock {\em Social Networks}, 5:109--137.

\bibitem[Hotelling, 1933]{hote:1933}
Hotelling, H. (1933).
\newblock Analysis of a complex of statistical variables into principal
  components.
\newblock {\em Journal of Educational Psychology}, 24:417--441.

\bibitem[Hotelling, 1936]{hote:1936}
Hotelling, H. (1936).
\newblock Relations between two sets of variates.
\newblock {\em Biometrika}, 28:321--377.

\bibitem[Hyv{\"a}rinen and Oja, 2000]{hyva:oja:2000}
Hyv{\"a}rinen, A. and Oja, E. (2000).
\newblock Independent component analysis: algorithms and applications.
\newblock {\em Neural Networks}, 13:411--430.

\bibitem[Johnson and Lindenstrauss, 1984]{john:lind:1984}
Johnson, W. and Lindenstrauss, J. (1984).
\newblock Extensions of lipschitz mapping into a hilbert space.
\newblock {\em Contemporary Mathematics}, 26:189--206.

\bibitem[Joliffe, 2002]{joli:2002}
Joliffe, I. (2002).
\newblock {\em Principal Component Analysis}.
\newblock Springer, New York.

\bibitem[Karrer and Newman, 2011]{karr:newm:2011}
Karrer, B. and Newman, M. (2011).
\newblock Stochastic blockmodels and community structure in networks.
\newblock {\em Physical Review E}, 83.

\bibitem[Kazemi et~al., 2020]{kaze:goel:jain:koby:seth:fors:poup:2020}
Kazemi, S., Goel, R., Jain, K., Kobyzev, I., Sethi, A., Forsyth, P., and
  P.Poupart (2020).
\newblock Representation learning for dynamic graphs: A survey.
\newblock {\em Journal of Machine Learning Research}, 21:1--73.

\bibitem[Kim et~al., 2019]{kim:rina:wass:2019}
Kim, J., Rinaldo, A., and Wasserman, L. (2019).
\newblock Minimax rates for estimating the dimension of a manifold.
\newblock {\em Journal of Computational Geometry}.

\bibitem[Kobak and Linderman, 2021]{kobak:linderman:2021}
Kobak, D. and Linderman, G.~C. (2021).
\newblock Initialization is critical for preserving global data structure in
  both t-sne and umap.
\newblock {\em Nature biotechnology}, 39(2):156--157.

\bibitem[Kobourov, 2012]{kobo:2012}
Kobourov, S. (2012).
\newblock Spring embedders and forced directed graph drawing algorithms.
\newblock arXiv 1201.3011.

\bibitem[Kohonen, 1982]{koho:1982}
Kohonen, T. (1982).
\newblock Self-organized formation of topologically correct feature map.
\newblock {\em Biological Cybernetics}, 43:59--69.

\bibitem[Konishi and Kitagawa, 2008]{konishi2008information}
Konishi, S. and Kitagawa, G. (2008).
\newblock {\em Information criteria and statistical modeling}.
\newblock Springer.

\bibitem[Kossinets and Watts, 2006]{koss:watt:2006}
Kossinets, G. and Watts, D. (2006).
\newblock Empirical analysis of an evolving social network.
\newblock {\em Science}, 311:88--90.

\bibitem[Kuno and Suga, 1966]{kuno:suga:1966}
Kuno, U. and Suga, Y. (1966).
\newblock Multidimensional mapping of piano pieces.
\newblock {\em Japanese Psychological Research}, 8:119--124.

\bibitem[Lee and Wilkinson, 2019]{lee:wilk:2019}
Lee, C. and Wilkinson, D. (2019).
\newblock A review of stochastic block models and extensions for graph
  clustering.
\newblock {\em Applied Network Science}, pages 1--50.

\bibitem[Lei and Rinaldo, 2015]{lei:rina:2015}
Lei, J. and Rinaldo, A. (2015).
\newblock Consistency of spectral clustering in stochastic block models.
\newblock {\em Annals of Statistics}, 43:215--237.

\bibitem[Levina and Bickel, 2004]{levi:bick:2005}
Levina, E. and Bickel, P. (2004).
\newblock Maximum likelihood estimation of intrinsic dimension.
\newblock In Saul, L., Weiss, Y., and Bottou, L., editors, {\em Advances in
  Neural Information Processing Systems}, volume~17. MIT Press.

\bibitem[Li et~al., 2007]{li:hast:chur:2007}
Li, P., Hastie, T., and Church, K. (2007).
\newblock Nonlinear estimators and tail bounds for dimension reduction in $l_1$
  using cauchy random projections.
\newblock {\em Journal of Machine Learning Research}, 8:2497--2532.

\bibitem[Little et~al., 2011]{litt:magg:rosa:2011}
Little, A., Maggioni, M., and Rosasco, L. (2011).
\newblock Multiscale geometric methods for estimating intrinsic dimension.
\newblock Proc. SampTA 4:2.

\bibitem[Lunde et~al., 2020]{lunde2020information}
Lunde, B. {\AA}.~S., Kleppe, T.~S., and Skaug, H.~J. (2020).
\newblock An information criterion for automatic gradient tree boosting.
\newblock {\em arXiv preprint arXiv:2008.05926}.

\bibitem[Luxburg, 2007]{vonl:2007}
Luxburg, U.~V. (2007).
\newblock A tutorial on spectral clustering.
\newblock {\em Statistics and Computing}, 17:395--416.

\bibitem[Markov, 1958]{mark:1958}
Markov, A. (1958).
\newblock Insolubility of the problem of homeomorphy.
\newblock Proc. Intern Congress of Mathematicians.

\bibitem[McInnes et~al., 2018]{mcin:heal:melv:2018}
McInnes, L., Healy, J., and Melville, J. (2018).
\newblock {U}{M}{A}{P}: Uniform manifold approximation for dimension reduction.
\newblock arXiv:1802.03426v2.

\bibitem[Mikolov et~al., 2013]{miko:suts:chen:corr:dean:2013b}
Mikolov, T., Sutskever, I., Chen, K., Corrado, G., and Dean, J. (2013).
\newblock Distributed representation of words and phrases and their
  composability.
\newblock In Advances in Neural Information Processing Systems 26: Proceedings
  Annual 27th Conference on Neural Information Processing Systems. Lake Tahoe,
  Nevada, USA.

\bibitem[Mornelli et~al., 2005]{morn:gigu:peti:2007}
Mornelli, C., Giguere, C., and Petit, K. (2005).
\newblock The efficiency/security trade-off in criminal networks.
\newblock {\em Social Networks}, 29:143--153.

\bibitem[Newman, 2006]{newm:2006b}
Newman, M. (2006).
\newblock Modularity and community structure in networks.
\newblock {\em Proceedings of the National Academy of Science, PNAS},
  103:8577--8582.

\bibitem[Newman, 2020]{newm:2020}
Newman, M. (2020).
\newblock {\em Networks}.
\newblock Oxford University Press.
\newblock 2nd revised edition.

\bibitem[Newman and Girvan, 2004]{newm:girv:2004}
Newman, M. and Girvan, M. (2004).
\newblock Finding and evaluating community networks.
\newblock {\em Physical Review E}, 69:02613--1 -- 02613--15.

\bibitem[Newman and Reinert, 2016]{newm:rein:2016}
Newman, M. and Reinert, G. (2016).
\newblock Estimating the number of communities in a network.
\newblock {\em Physical Review Letters}, 137.

\bibitem[Niyogi et~al., 2008]{niyo:smal:wein:2008}
Niyogi, P., Smale, S., and Weinberger, S. (2008).
\newblock Finding the homology of submanifolds with high confidence from random
  samples.
\newblock {\em Discrete and Computational Geometry}, 39:419--441.

\bibitem[Otneim et~al., 2020]{otne:jull:tjos:2020}
Otneim, H., Jullum, M., and Tj{\o}stheim, D. (2020).
\newblock Pairwise local {F}isher and naive {B}ayes: Improving two standard
  discriminants.
\newblock {\em Journal of Econometrics}, 216:284--304.

\bibitem[Ozertem and Erdogmus, 2011]{ozer:erdo:2011}
Ozertem, U. and Erdogmus, D. (2011).
\newblock Locally defined principal curves and surfaces.
\newblock {\em Journal of Machine Learning Research}, 12:1249--1286.

\bibitem[Pearson, 1901]{pear:1901}
Pearson, K. (1901).
\newblock On lines and planes of closest fit to systems of points in space.
\newblock {\em Philosophical Magazine}, 2:559--572.

\bibitem[Peixito, 2021]{peix:2021}
Peixito, T. (2021).
\newblock Descriptive vs. inferential community detection: pitfalls, myths and
  half-truths.
\newblock arXiv:2112.00183v1.

\bibitem[Perozzi et~al., 2014]{perr:alfr:skie:2014}
Perozzi, B., Al-Rfou, R., and Skiena, S. (2014).
\newblock Deepwalk: Online learning of social representations.
\newblock In {\em Proceedings of the 20th ACM SIGKDD international conference
  on Knowledge discovery and data mining}, pages 701--710.

\bibitem[Qiao and Polonik, 2021]{qiao:polo:2021}
Qiao, W. and Polonik, W. (2021).
\newblock Algorithms for ridge estimation with convergence guarantees.
\newblock arXiv:2014.12314v1.

\bibitem[Qiu et~al., 2018]{qiu:dong:ma:li:wang:tang:2018}
Qiu, J., Dong, Y., Ma, H., Li, J., Wang, K., and Tang, J. (2018).
\newblock Network embedding as matrix factorization.: unifying deepwalk,
  {L}{I}{N}{E}, {P}{T}{E}, and node2vec.
\newblock Proceedings WSDM, ACM, New Tork, NY, USA.

\bibitem[Qiu et~al., 2019]{qiu:dong:ma:li:wang:tang:2019}
Qiu, J., Dong, Y., Ma, H., Li, J., Wang, K., and Tang, J. (2019).
\newblock Net{S}{M}{F}: Large-scale network embedding as sparse matrix
  factorization.
\newblock In Proceedings of the 2019 World Wide Web Conference, May 13-17, San
  Francisco, CA, USA.

\bibitem[Ravisshanker and Chen, 2019]{ravi:chen:2019}
Ravisshanker, N. and Chen, R. (2019).
\newblock Topological data analysis (tda) for time series.
\newblock arXiv: 1909.10604v1.

\bibitem[Rohe et~al., 2011]{rohe:chat:yu:2011}
Rohe, K., Chatterjee, S., and Yu, B. (2011).
\newblock Spectral clustering and the high-dimensional stochastic blockmodel.
\newblock {\em Annals of Statistics}, 39:1878--2015.

\bibitem[Rohe et~al., 2016]{rohe:qin:yu:2016}
Rohe, K., Qin, T., and Yu, B. (2016).
\newblock Co-clustering directed graphs to discover asymmetries and directional
  communities.
\newblock {\em Proceeding National Academy of Science}, 113:12679--12684.

\bibitem[Roweis and Saul, 2000]{rowe:saul:2000}
Roweis, S. and Saul, L. (2000).
\newblock Nonlinear dimensionality reduction by locally linear embedding.
\newblock {\em Science}, 290:2323--2326.

\bibitem[Sammon, 1969]{samm:1969}
Sammon, J. (1969).
\newblock A nonlinear mapping for data structure analysis.
\newblock {\em IEEE Transactions on Computers}, 18:403--409.

\bibitem[Sch{\"o}lkopf et~al., 2005]{scho:smol:mull:2005}
Sch{\"o}lkopf, B., Smola, A., and M{\"u}ller, K.-L. (2005).
\newblock Kernel principal components.
\newblock {\em Lecture Notes in Computer Science}, 1327:583--588.

\bibitem[Shahriari et~al., 2015]{shahriari2015taking}
Shahriari, B., Swersky, K., Wang, Z., Adams, R.~P., and De~Freitas, N. (2015).
\newblock Taking the human out of the loop: A review of bayesian optimization.
\newblock {\em Proceedings of the IEEE}, 104(1):148--175.

\bibitem[Singh et~al., 2007]{sing:memo:carl:2007}
Singh, G., Memoli, F., and Carlsson, G. (2007).
\newblock Topological methods for the analysis of high dimensional data sets
  and 3d object recognition.
\newblock In {\em Eurographics Symposium on Point Based Graphics}. The
  Eurographics Association.
\newblock M. Botsch and R. Pajarola.

\bibitem[Sun et~al., 2012]{sun:nori:han:yan:yu:yu:2012}
Sun, Y., Norick, B., Han, J., Yan, X., Yu, P., and Yu, X. (2012).
\newblock Integrating meta-path selection with user-guided object clustering in
  heterogeneous information networks.
\newblock In {\em KDD '12: Proceedings of the 18th ACM SIGKDD international
  conference on Knowledge discovery and data mining}, page 1348–1356.

\bibitem[Tang et~al., 2016]{tang:liu:zhan:mei:2016}
Tang, J., Liu, J., Zhang, M., and Mei, Q. (2016).
\newblock Visualizing large-scale and high-dimensional data.
\newblock In {\em Proceedings of the 25th international conference on world
  wide web}, pages 287--297.

\bibitem[Tang et~al., 2015a]{tang:qu:mei:2015a}
Tang, J., Qu, M., and Mei, Q. (2015a).
\newblock {P}{T}{E}: Predictive text embedding through large-scale
  heterogeneous text networks.
\newblock arXiv. 1508.00200v1.

\bibitem[Tang et~al., 2015b]{tang:qu:wang:zhan:yan:mei:2015b}
Tang, J., Qu, M., Wang, M., Zhang, M., Yan, J., and Mei, Q. (2015b).
\newblock {{L}{I}{N}{E}: Large-scale information network embedding}.
\newblock In {\em Proceedings of the 24th international conference on world
  wide web}, pages 1067--1077.

\bibitem[Tenenbaum et~al., 2000]{tene:silv:lang:2000}
Tenenbaum, J., de~Silva, V., and Langford, J. (2000).
\newblock A global geometric framework for nonlinear dimensionality reduction.
\newblock {\em Science}, 290:2319--2323.

\bibitem[Tj{\o}stheim et~al., 2022a]{supp}
Tj{\o}stheim, D., Jullum, M., and L{\o}land, A. (2022a).
\newblock {Supplement to ``Statistical embedding: Beyond principal
  components''}.
\newblock DOI here.

\bibitem[Tj{\o}stheim et~al., 2022b]{tjos:otne:stov:2021}
Tj{\o}stheim, D., Otneim, H., and St{\o}ve, B. (2022b).
\newblock Statistical dependence: Beyond pearson's $\rho$.
\newblock {\em Statistical Science}, 37(1):90--109.

\bibitem[Tj{\o}stheim et~al., 2022c]{tjos:otne:stov:2021a}
Tj{\o}stheim, D., Otneim, H., and St{\o}ve, B. (2022c).
\newblock {\em Statistical Modeling Using Local Gaussian Approximation}.
\newblock Academic Press.

\bibitem[Torgerson, 1952]{torg:1952}
Torgerson, W. (1952).
\newblock Multidimensional scaling: 1 theory and method.
\newblock {\em Psychometrica}, 29:1--27.

\bibitem[Tutte, 1963]{tutt:1963}
Tutte, W. (1963).
\newblock How to draw a graph.
\newblock {\em Proceedings of the London Mathematical Society}, 13:743--768.

\bibitem[van~der Maaten, 2014]{vand:2014}
van~der Maaten, L. (2014).
\newblock Accelerating t-{S}{N}{E} using tree-based algorithms.
\newblock {\em Journal of Machine Learning Research}, pages 3221--3245.

\bibitem[van~der Maaten and Hinton, 2008]{vand:hint:2008}
van~der Maaten, L. and Hinton, G. (2008).
\newblock Visualizing data using t-{S}{N}{E}.
\newblock {\em Journal of Machine Learning research}, 9:2579--2605.

\bibitem[van~der Maaten et~al., 2009]{vand:post:vand:2009}
van~der Maaten, L., Postma, E., and van~der Herik, J. (2009).
\newblock Dimensionality reduction: A comparative review.
\newblock Tilburg Centre for Creative Computing, TiCC TR 2009.005.

\bibitem[Wang and Bickel, 2017]{wang:bick:2017}
Wang, Y. and Bickel, P. (2017).
\newblock Likelihood-based model selection for stochastic block models.
\newblock {\em Annals of Statistics}, 45:500--528.

\bibitem[Wasserman, 2018]{wass:2018}
Wasserman, L. (2018).
\newblock Topological data analysis.
\newblock {\em Annual Review of Statistics and its Applications}, 5:501--532.

\bibitem[Wilson and Zhu, 2008]{wils:zhu:2008}
Wilson, R. and Zhu, P. (2008).
\newblock A study of graph spectra for computing graphs and trees.
\newblock {\em Journal of Pattern Recognition}, 4:2833--2841.

\bibitem[Xie et~al., 2018]{xie:li:xue:2018}
Xie, H., Li, J., and Xue, H. (2018).
\newblock A survey of dimensionality reduction techniques based on random
  projection.
\newblock arXiv:1706.04371v4.

\bibitem[Young and Householder, 1938]{youn:hous:1938}
Young, G. and Householder, A. (1938).
\newblock Discussion of a set of points in terms of their mutual distances.
\newblock {\em Psychometrika}, 3:19--22.

\bibitem[Zhang and Chen, 2020]{zhan:chen:2018}
Zhang, J. and Chen, Y. (2020).
\newblock Modularity based community detection in heterogeneous networks.
\newblock {\em Statistica Sinica}, 30(2):601--629.

\bibitem[Zheng, 2016]{zhen:2016}
Zheng, Q. (2016).
\newblock Spectral techniques for heterogeneous social networks.
\newblock PhD thesis, Queen's University, Ontario, Canada.

\bibitem[Zhou et~al., 2017]{zhou:liu:liu:gao:2017}
Zhou, C., Liu, Y., Liu, X., and J.Gao (2017).
\newblock Scalable graph embedding for asymmetric proximity.
\newblock Proceedings of the 31st AAAI Conference on Artificial Intelligence.

\bibitem[Zhou et~al., 2018]{zhou:yang:ren:wu:zhua:2018}
Zhou, L., Yang, Y., Ren, X., Wu, F., and Zhuang, Y. (2018).
\newblock Dynamic network embedding by modeling triadic closure process.
\newblock 32nd AAAI Conference on Artificial Intelligence.

\bibitem[Zhu and Pan, 2020]{zhu:pan:2020}
Zhu, X. and Pan, R. (2020).
\newblock Grouped network vector autoregression.
\newblock {\em Statistica Sinica}, 30(3):1437--1462.

\bibitem[Zhu et~al., 2017]{zhu:pan:li:liu:wang:2017}
Zhu, X., Pan, R., Li, G., Liu, Y., and Wang, H. (2017).
\newblock Network vector autoregression.
\newblock {\em Annals of Statistics}, 45:1096--1123.

\bibitem[Zomordian and Carlsson, 2005]{zomo:carl:2005}
Zomordian, A. and Carlsson, G. (2005).
\newblock Computing persistent homology.
\newblock {\em Discrete Computational Geometry}, 33:249--274.

\end{thebibliography}


\begin{thebibliography}{}

\bibitem[Bengio et~al., 2003]{beng:duch:vinc:jauv:2003}
Bengio, Y., Ducharme, R., Vincent, P., and Jauvin, C. (2003).
\newblock A neural probabilistic language model.
\newblock {\em Journal of Machine Learning Research}, 3:1137--1155.

\bibitem[Bubenik, 2015]{bube:2015}
Bubenik, P. (2015).
\newblock Statistical topological data analysis using persistence landscapes.
\newblock {\em Journal of Machine Learning Research}, 16:77--102.

\bibitem[Carriere and Oudot, 2019]{carr:oudo:2019}
Carriere, M. and Oudot, S. (2019).
\newblock Sliced {W}asserstein kernel for persistence diagrams.
\newblock arXiv 1803.07961v1.

\bibitem[Cartsens and Horadam, 2013]{cart:hora:2013}
Cartsens, C. and Horadam, K. (2013).
\newblock Persistent homology of collaboration networks.
\newblock {\em Mathematical Problems in Engineering}, pages 1--7.

\bibitem[Chazal et~al., 2011]{chaz:cohe:mego:2011}
Chazal, F., Cohen-Steiner, D., and M\'{e}got, Q. (2011).
\newblock Geometric inference for probability measures.
\newblock {\em Foundations of Computational Mathematics}, 11:733--751.

\bibitem[Chazal et~al., 2015]{chaz:fasy:lecc:rina:wass:2015}
Chazal, F., Fasy, B., Lecci, F., Rinaldo, A., and Wasserman, L. (2015).
\newblock Stochastic convergence of persistence landscapes and silhouettes.
\newblock {\em Journal of Computational Geometry}, 6:140--161.

\bibitem[Chazal et~al., 2016]{chaz:mass:mich:2016}
Chazal, F., Massart, P., and Michel, B. (2016).
\newblock Rates of convergence for robust geometric inference.
\newblock {\em Electronic Journal of Statistics}, 10:2243--2286.

\bibitem[Chazal and Michel, 2021]{chaz:mich:2021}
Chazal, F. and Michel, B. (2021).
\newblock An introduction to topological data analysis: fundamental and
  practical aspects for data scientists.
\newblock {\em Frontiers in Artificial Intelligence: Machine Learning and
  Artificial Intelligence}, 4:1--28.

\bibitem[Curry et~al., 2018]{curr:mukh:turn:2018}
Curry, J., Mukherjee, S., and Turner, K. (2018).
\newblock How many directions determine a shape and other sufficiency results
  for two topological transforms.
\newblock arXiv:1805.09782.

\bibitem[Edelsbrunner and Harer, 2010]{edel:hare:2010}
Edelsbrunner, H. and Harer, J. (2010).
\newblock {\em Computational Topology: An Introduction}.
\newblock American Mathematical Society.

\bibitem[Goldberger and Levy, 2014]{gold:levy:2014}
Goldberger, Y. and Levy, O. (2014).
\newblock word2vec explained: Deriving mikolov et al.'s negative-sampling
  word-embedding method.
\newblock arXiv: 1402.3722v.1.

\bibitem[Kusano and Hiraoka, 2016]{kusa:fuku:hira:2016}
Kusano, G. and Hiraoka, Y. (2016).
\newblock Persistence weighted gaussian kernel for topological data analysis.
\newblock Proceedings of the 33rd International Conference on Machine Learning,
  New York.

\bibitem[Le and Mikolov, 2014]{le:miko:2014}
Le, Q. and Mikolov, T. (2014).
\newblock Distributed representations of sentences and documents.
\newblock arXiv:1405.4053v2.

\bibitem[Maroulas et~al., 2020]{maro:nasr:obal:2020}
Maroulas, V., Nasrin, F., and Obello, C. (2020).
\newblock A bayesian framework for persistent homology.
\newblock {\em SIAM Journal of Mathematical Sciences}, 2.

\bibitem[Mikolov et~al., 2013a]{miko:chen:corr:dean:2013a}
Mikolov, T., Chen, K., Corrado, G., and Dean, J. (2013a).
\newblock Efficient estimation of word representations in vector space.
\newblock CoRR, abs/1301,3781.

\bibitem[Mikolov et~al., 2013b]{miko:le:suts:2013c}
Mikolov, T., Le, V., and Sutskever, I. (2013b).
\newblock Exploiting similarities among languages for machine translation.
\newblock arXiv:1309.4168.

\bibitem[Mikolov et~al., 2013c]{miko:suts:chen:corr:dean:2013b}
Mikolov, T., Sutskever, I., Chen, K., Corrado, G., and Dean, J. (2013c).
\newblock Distributed representation of words and phrases and their
  composability.
\newblock In Advances in Neural Information Processing Systems 26: Proceedings
  Annual 27th Conference on Neural Information Processing Systems. Lake Tahoe,
  Nevada, USA.

\bibitem[Mnih and Hinton, 2008]{mnih:hint:2008}
Mnih, A. and Hinton, G. (2008).
\newblock A scalable hierarchical distributed language model.
\newblock NIPS Proceedings 2008.

\bibitem[Moon and Lazar, 2020]{moon:laza:2020}
Moon, C. and Lazar, N. (2020).
\newblock Hypothesis testing for shapes using vectorized persistence diagrams.
\newblock arXiv:2006.0n46.

\bibitem[Morin and Bengio, 2005]{mori:beng:2005}
Morin, F. and Bengio, Y. (2005).
\newblock Hierarchical probabilistic neural network language model.
\newblock AISTATS.

\bibitem[Obayashi and Hiraoka, 2017]{obay:hira:2017}
Obayashi, I. and Hiraoka, Y. (2017).
\newblock Persistence diagrams with linear machine learning models.
\newblock arXiv preprint 1706.10082.

\bibitem[Ravisshanker and Chen, 2019]{ravi:chen:2019}
Ravisshanker, N. and Chen, R. (2019).
\newblock Topological data analysis (tda) for time series.
\newblock arXiv: 1909.10604v1.

\bibitem[Reininghaus et~al., 2015]{rein:hube:baue:kwit:2015}
Reininghaus, J., Huber, S., Bauer, U., and Kwitt, R. (2015).
\newblock A stable multi-scale kernel for topological machine learning.
\newblock In Proceedings of the IEEE Conference on Computer Vision and Pattern
  Recognition.

\bibitem[Rong, 2016]{rong:2016}
Rong, X. (2016).
\newblock word2vec parameter learning explained.
\newblock arXiv:1411.2738v4.

\bibitem[Schmidhuber, 2015]{schm:2015}
Schmidhuber, J. (2015).
\newblock Deep learning in neural networks: An overview.
\newblock {\em Neural Networks}, 61:85--117.

\bibitem[Umeda, 2017]{umed:2017}
Umeda, Y. (2017).
\newblock Time series classification via topological data analysis.
\newblock {\em Transactions of the Japanese society for Artificial
  Intelligence}, 32:1--12.

\bibitem[Wasserman, 2018]{wass:2018}
Wasserman, L. (2018).
\newblock Topological data analysis.
\newblock {\em Annual Review of Statistics and its Applications}, 5:501--532.

\end{thebibliography}

\end{document}


\maketitle

\section{Persistence diagrams and simplical complexes \label {Persistence diagrams}}

Assume that we observe a sample $X_1,\ldots,X_n$ drawn from a distribution $P$ supported on a set $S$, and let us define the empirical distance function
$$
\hat{d}(x) = \min_{1 \leq i \leq n}||x-X_i||.
$$
It should be noted that lower level sets $\hat{L}_{\varepsilon}$  defined by $\hat{L}_{\varepsilon} = \{x: \hat{d}(x) \leq \varepsilon\}$ are precisely the union of balls described in Equation (13) in the main paper, i.e.,
$$
\hat{L}_{\varepsilon} = \{x:\hat{d}(x) \leq \varepsilon\} = \cup_{i=1}^{n} B(X_i,\varepsilon).
$$
The persistence diagram $\hat{D}$ defined by these lower level sets is an estimate of the underlying diagram $D$.

The empirical distance function is often used for defining the persistence diagram of a data set in computational topology. However, as pointed out by \cite{wass:2018}, from a statistical point of view this is a poor choice, as it is highly non-robust. Wasserman points out several more robust alternatives. One of them is the so called DTM distance introduced by \cite{chaz:cohe:mego:2011} 
given by
$$
\hat{d}_m^2(x) = \frac{1}{k}\sum_{i=1}^{k}||x-X_i(x)||^2,
$$
where $k = [mn]$ is the largest integer less than or equal to $mn$ and with $0 \leq m \leq 1$ being a scale parameter. Further, $X_j(x)$ denotes the data after re-ordering them so that $||X_1(x)-x|| \leq ||X_2(x)-x|| \leq \cdots$. This means that $\hat{d}_m^2(x)$ is the average squared distance to the $k$-nearest neighbors Other alternative references to a robustified 
distance measure are given in \cite{wass:2018}.

Actually, in more complicated situations, the persistence diagram is not computed directly from $\hat{L}_{\varepsilon}$, but from so-called simplical complexes. This approach is particularly interesting since it generalizes the embedding of a point cloud in a graph as described in Sections 3.4 and 3.5 in the main manuscript. We will give a brief description here. Much more details can be found in  \cite{chaz:mich:2021}.

First, recall the definition of a simplex: Given a set $\mathbb{X} = \{X_0,\ldots,X_k\} \subset \mathbb{R}^{p}$ of $k+1$ ``affinely independent'' (i.e., the vectors $(X_0,X_1, \ldots X_k)$ are linearly independent), the $k$-dimensional simplex $\sigma = [X_0,\ldots,X_k]$ spanned by $\mathbb{X}$ is the convex hull of $\mathbb{X}$. For instance, for $k=1$ the simplex is simply given by the line from $X_0$ to $X_1$. The points of $\mathbb{X}$ are called the nodes of $\sigma$ and the simplices spanned by the subsets of $\mathbb{X}$ are called the faces of $\sigma$. A geometric simplical complex $K$ in $\mathbb{R}^{p}$ is a collection of simplices such that (i) any face of a simplex of $K$ is a simplex of $K$, (ii) the intersection of any two simplices of $K$ is either empty or a common face of both.

As seen in Sections 3.4 and 3.5 in the main paper, connecting pairs of nearby data points by edges leads to the standard notion of a neighboring graph from which the connectivity of the data can be analyzed and clustering can be obtained, including non-convex situations, as described in Section 3.4. Using simplical complexes, where simplical complexes of dimension 1 are graphs, one can go beyond this simple form of connectivity. In fact a central idea in TDA is to build higher dimensional equivalents of neighboring graphs by not only connecting pairs but also $(k+1)$-tuples of nearby data points. This enables one to identify new topological features such as cycles and voids and their higher dimensional counterparts. Regarding embedding of networks, as treated in Section 5, such a technique could possibly be used to discover cycles in networks such as criminal rings in fraud detection, say.

Simplical complexes are mathematical objects that have both topological and algebraic properties. This makes them especially useful for TDA
There are two main examples of complexes in use. They are the Vietoris-Rips complex and the $\breve{\mbox{C}}$ech complex. The Vietoris-Rips complex $V_{\varepsilon}(\mathbb{X})$ can be introduced in a metric space $(M,d)$. It is the set of simplices  $\mathbb{X} = [X_0,\ldots,X_k]$ such that $d_{\mathbb{X}}(X_i,X_j) \leq \varepsilon$ for all $(i,j)$. The $\breve{\mbox{C}}$ech complex $C_{\varepsilon}(\mathbb{X})$ is defined as the simplices $[X_0,\ldots,X_k]$ such that the $k+1$ balls $B(X_i,\varepsilon)$ have a nonempty intersection.  

These definitions should be compared to the use of ball-coverings in Section 4 of the main paper and level sets defined in the present subsection. It can in fact be shown that the homology of $\hat{L}_{\varepsilon}$ is the same as the homology of $C_{\varepsilon}$. The homology of $C_{\varepsilon}$ can be computed using basic matrix operations. All relevant computations can be reduced to linear algebra. This gives a method of computing homology and persistent homology relating the complexes as $\varepsilon$ varies as briefly mentioned in our simple introductory example of chain of circles, or the more involved example involving Ranunculoids, in Section 4.2 of the main paper (see \cite{edel:hare:2010}). 
In fact, it is computationally easier to work out the algebra for the Vietoris-Rips complex $V_{\varepsilon}$. It can be shown that the persistent homology defined by $V_{\varepsilon}$ approximates the persistent homology defined by $C_{\varepsilon}$.

Given a subset $\mathbb{X}$ of a compact metric space $(M,d)$, the families of Vietoris-Rips complexes, $\{V_{\varepsilon}(\mathbb{X})\}_{\varepsilon \in \mathbb{R}}$ and the family of $\breve{\mbox{C}}$ech complexes, $\{C_{\varepsilon}(\mathbb{X})\}_{\varepsilon \in \mathbb{R}}$ are filtrations, that is, nested families of complexes. As indicated earlier, the parameter $\varepsilon$ can be considered as a data resolution level at which one considers the data set $\mathbb{X}$. For example if $\mathbb{X}$ is a point cloud in $\mathbb{R}^{p}$, the filtration $\{C_{\varepsilon}\}$ encodes the topology of the whole family of unions of balls $\mathbb{X}^{\varepsilon} = \cup_{X \in \mathbb{X}}B(X,\varepsilon)$ as $\varepsilon$ goes from 0 to $\infty$. 

As in the example in Section 4.2 of the main paper, the homology of a filtration $\{F_{\varepsilon}\}$ changes as $\varepsilon$ increases: new connected components can appear, existing components can merge, loops and cavities may appear or be filled. Persistence homology tracks these changes, identifies the appearing features, and attaches a lifetime to them. The resulting information can be encoded as a set of intervals, the bar-code, or equivalently, as a multiset of points in $\mathbb{R}^2$, where the coordinates of each point is the start and end point of the corresponding interval. In \cite{chaz:mich:2021} a formal definition of bar-code and persistence diagram is given via the concept of persistence module which again is defined in terms of an indexed family of vector spaces and a doubly-indexed family of linear maps.

\subsection{Persistent landscapes, functional spaces and appli\-cations\label{Persistent landscape}}

The space of persistence diagrams is not a function space in the sense that it is not a Hilbert space. This may make it more difficult to directly apply methods from statistics and machine learning. For example,  the definition of a mean persistence diagram is not obvious and unique \citep[p. 28]{chaz:mich:2021}.  
Further, according to \citet[p. 29]{chaz:mich:2021} the highly nonlinear nature of diagrams prevents them from being used as a standard feature of machine learning algorithms. An exception, however, is \cite{obay:hira:2017}. 

\cite{bube:2015} introduced persistence landscapes. The persistence landscape is a collection of continuous linear functions obtained by transforming the points of the persistence diagram into tent functions. This function space can be given a Hilbert space structure (in fact a more general structure of a separable Banach space in Bubenik's original paper). The random structure created by $X_1,\ldots,X_n$ may then be represented by Hilbert space variables, and it becomes meaningful to consider means, variances and a central limit theorem.  
The vector space structure of persistent landscapes and similar constructions 
 may appear to be more directly extendable to machine learning, in particular to kernel methods, cf.\ also Section 3.7 in the main paper, in reproducing kernel Hilbert space (see for instance \cite{rein:hube:baue:kwit:2015}, \cite{kusa:fuku:hira:2016} and \cite{carr:oudo:2019}).
It can safely be stated that combining TDA and persistence homology with machine learning is becoming an active research direction  with results having potential for unsolved practical problems.

Clearly, the bar codes, the persistence diagrams and Betti numbers
can also be used directly as feature extractors for
classification problems.
In particular, these have been used for network
characterizations in \cite{cart:hora:2013}. Possibly such features can
be used as a supplement to the network embedding and clustering
methods presented in  Section 5 in the main paper of this survey.

Connections between persistent homology and deep learning has also started to be explored. \cite{umed:2017} has done this in a time series context. Another application to time series is \cite{ravi:chen:2019}. 

For applications to specific problems we refer to references in \cite{wass:2018} and \cite{chaz:mich:2021}. \citeauthor{wass:2018} discusses briefly applications to the cosmic web, images and proteins, \citeauthor{chaz:mich:2021} discuss applications to protein binding configurations and classification of sensor data. 

\subsection{Statistical inference\label{Statistical inference}}

A central concept in inference for persistence diagrams is the bottleneck distance. Given two diagrams $C_1$ and $C_2$, the bottleneck distance is defined by 
$$
\delta_{\infty}(C_1,C_2) = \inf_{\gamma} \sup_{z \in C_1}||z-\gamma(z)||_{\infty},
$$
where $\gamma$ ranges over all bijections between $C_1$ and $C_2$. Intuitively, this is like overlaying the two diagrams and asking how much one has to shift the diagrams to make them the same \citep{wass:2018}. The practical computation of the bottleneck distance amounts to the computation of perfect matching in a bipartite graph for which classical algorithms can be used \citep{chaz:mich:2021}. 

The bottleneck distance is a natural tool to express stability of persistence diagrams. An alternative distance measure is the Wasserstein distance.
The bottleneck distance is also a natural tool in statistical inference on persistent  landscapes, cf.\ \cite{chaz:fasy:lecc:rina:wass:2015}.

The (estimated) persistence diagram $\hat{C}$ is based on a finite
collection of random variables $X_1,\ldots,X_n$. One might think of a
true persistence diagram $C$ as $n \to \infty$. A central question is
then whether there is such a thing as consistency, and is it possible
to introduce confidence intervals? Such questions have been considered
by \citet[Section 5.7; see especially Section 5.7.4]{chaz:mich:2021} and is based on the bottleneck distance between $\hat{C}$ and $C$.

For many applications, in particular when the point cloud does not come from a (perturbation of) a geometric structure, the persistence diagram will look quite complicated. In particular, there will be a number of cases where the life time is quite short and consequently with representative points close to the diagonal. The question then arises whether these points can be considered as noise and should therefore be eliminated from the diagram. One needs a concept of statistical significance to make such an evaluation, and again the bottleneck distance can be used as a tool. When estimating a persistence diagram $C$ with an estimator $\hat{C}$ one may look for a quantile type number $\eta_{\alpha}$ such that
\begin{equation}
P(d_{\infty} \geq \eta_{\alpha}) \leq \alpha, 
\label{A14a}
\end{equation}
for $\alpha \in (0,1)$. This can be taken as a point of departure for computation of confidence intervals and significance tests.

It is necessary to translate (\ref{A14a}) into something that can be computed. This can be done by the bootstrap as in \cite{chaz:mass:mich:2016}. 
Let $(X_1^*,\ldots,X_n^*)$ be a sample from the empirical measure defined from the observations $(X_1,\ldots,X_n)$. Moreover, let $\hat{C}^*$ be the persistence diagram derived from this sample. One can then take as an estimate of $\eta_{\alpha}$ the quantity $\hat{\eta}_{\alpha}$ defined by 
$$
P[d_{\infty}(\hat{C}^*,\hat{C}) > \hat{\eta}_{\alpha}|X_1,\ldots,X_n] = \alpha,
$$
where it is straightforward to estimate $\hat{\eta}_{\alpha}$ by Monte Carlo integration. \cite{chaz:mass:mich:2016} have shown that the bootstrap is valid  when computing the sub-level sets of a density estimator. Using the bottleneck bootstrap and given a certain significance level, a band can be constructed parallel to the diagonal of the persistence diagram, and such that points in this level are considered as noise. A bootstrap algorithm can also be used to construct confidence bands for landscapes as shown in \cite{chaz:mass:mich:2016}. 

There are a number of problems of interest for statisticians in TDA. \cite{chaz:mich:2021} in particular mentions four topics:
\begin{enumerate}
\item Proving consistency and studying the convergence rates of TDA methods.
\item Providing confidence regions for topological features and discussing the significance of estimated topological quantities.
\item Selecting relevant scales (i.e.\ selecting $\varepsilon$ in the examples discussed above) at which topological phenomenons should be considered as functions of observed data.
\item Dealing with outliers and providing robust methods for TDA.
\end{enumerate}
In addition, one may want to introduce the block bootstrap to take better care of dependence structures There are also recent contributions to hypothesis testing, \cite{moon:laza:2020}, sufficient statistics, \cite{curr:mukh:turn:2018}, and Bayesian statistics for topological data analysis, \cite{maro:nasr:obal:2020}.

\section{Embedding and word feature repre\-sen\-tation of a language text \label{Embedding word}}

Sections 5.2 and 5.3 of the main paper describe the importance of  embedding of networks and its use in feature extraction, in clustering, characterization and classification for ultra-large data sets. It was pointed out in Section 5.3 that a main methodology for this is the Skip-Gram procedure which was developed in the context of word embedding for a natural language. The purpose of the present section is twofold. First,  language processing is of considerable independent interest. Second, it provides more details on the Skip-Gram procedure, its background and its use. Although this material is couched in terms of language analysis, we believe that when read in conjunction with Section 5.3 of the main paper, it will also provide added insight into the details of network embedding.

\subsection{A  few basic facts of neural nets\label{Basic facts}}

The  Skip-Gram procedure is based on a neural network with a single hidden layer, and we therefore include a brief summary of neural networks in this supplement.

Neural networks are used for a number of problems in prediction, classification and clustering.  The developments perhaps stagnated somewhat in the early seventies, but received renewed interest the last decades, following a massive increase in computational power. Currently, there is an intense activity involving among other things deep learning, where some remarkable results have been obtained. See \cite{schm:2015} for a relatively recent overview. 

Assume that we are given an $n$-vector $x$ as input. In a neural network approach one is interested in transforming $x$ via linear combinations of its components and possibly a nonlinear transformation of these linear combinations. This transformation constitutes what is called a hidden layer. Then this might be sent through a new transformation of the same type to create a new hidden layer and eventually to an output layer $y$ which should be as close as possible to a target vector $t$. If there is more than one hidden layer, it is said to be a deep network, its analysis being a base for so-called deep learning. In this supplement, mainly dealing with the background of the Skip-Gram, only the case of one hidden layer will be treated, that, in our context, will be formed by a linear transformation.

Given the input layer, the first step in forming the hidden layer is to form linear combinations 
\begin{equation}
h_i = \sum_{j=1}^{n}w_{ij}x_j,
\label{A21a}
\end{equation} 
where $i=1,\ldots,m$. Note that implicitly, there may be a constant term by taking $x_1$, say, equal to 1. (This is sometimes termed the bias term of the linear combination.)

In the case of one hidden layer, the output layer is  given by 
$$
y_j = \sum_{i=1}^{m}w_{ij}^{\prime}h_i,
$$
for $j=1,\ldots,q$. In subsequent applications for language and network embedding models $q = \mbox{dim}(y) = \mbox{dim}(x) = n$.

In a classification problem, $y_j$ may be associated  with an unnormalized probability for a class $j$, which in Section 5.3 of the main paper is the appropriate neighborhood of a node $v_j$ in a network. In such cases the output layer is also transformed. A common transformation is the so-called softmax function given by
\begin{equation}
\mbox{softmax}(y_j) = \frac{\exp (y_j)}{\sum_{i=1}^{n} \exp (y_i)}.
\label{A21b}
\end{equation}
This is recognized (if there is no hidden layer) as the multinomial logistic regression model which is a standard tool in classification. 

Using a training set, the coefficients (or weights) $w_{ij}$ and 
$w_{ij}^{\prime}$ are determined by a penalty function measuring the distance between the output $y$ and the target vector $t$, for example measured by the loss function $E = ||y-t||^2$. In a classification and clustering problem the training set consists of input vectors $x$ belonging to known classes $i$ (known words in the vocabulary in the text). The target vector is a so-called ``one hot'' vector having 1 at the component $j$ for the given target word and zeros elsewhere. The weights are adjusted such that the output vector is as close as possible to this vector, which means that the softmax function should be maximized for this particular component and ideally $\exp(y_i) \approx 0$ for $i \neq j$. 

The error function is evaluated for each of the samples coming in as inputs, and the gradient of the error function with respect to $y$ is evaluated with the weights being re-computed and updated in the direction of the gradient  by stochastic gradient descent. 

The weights $w_{ij}^{\prime}$ for the output layer is computed first and then $w_{ij}$ by the chain differentiation rule using so-called back propagation. Details are given in e.g.\ the appendix of \cite{rong:2016}. Schematically this may be represented by
$$
w_{ij}^{(\mbox{new})} = w_{ij}^{(\mbox{old})} - \varepsilon \frac{\partial E}{\partial w_{ij}}
$$
and similarly for $w^{\prime}_{ij}$. Initial values for the weights can be chosen by drawing from a set of uniform variables. Below the updating scheme will be illustrated on word representation of natural languages, which  next can be applied to embedding of networks.
  
\subsection{Word feature representation of natural languages\label{Word feature}}

Consider a natural language text. We start with a set of input vectors $x_i$, $i=1,\ldots,n$, where $n$ is the number of words in the vocabulary of the text, and $x_i$ represents word $i$ in the vocabulary. Each vector is of dimension $n$, where $x_i$ has a one in position $i$ of the vector and zeros elsewhere (``one-hot'' encoded vector). Let $m$ be the dimension of the desired word embedding feature representation. The dimension may be quite large. Common choices are in the range $100 - 1000$. Let the one-hot vector for the word $w_i$, word number $i$ in the vocabulary, be $x_i$. Further, consider a $n \times m$ weight matrix $ {\bf W}$. Define the $m$-dimensional hidden units $h_i$, $i=1,\ldots, m$ (without a nonlinear transformation) by
\begin{equation}
h_i = {\bf W}^{T}x_i  \doteq  v_{w_{i}}^{T},
\label{A22a}
\end{equation}
which is essentially copying the $m$-dimensional $i$th row of ${\bf W}$ to $h_i$. The vector $v_{w_{i}}$ is the input word representation vector for word number $i$ in the vocabulary, or the feature vector $f_i$ of the word $w_i$ . This means that the link (activation) function of the hidden layer units is simply {\it linear}. The weights, i.e., the vector word representation can then be learned by the neural network given appropriate targets and a penalty function.

An obvious question is whether a nonlinear transformation is needed. \cite{beng:duch:vinc:jauv:2003}, in their pioneering paper suggest an added  nonlinearity, whereas the approach of \cite{miko:chen:corr:dean:2013a,miko:suts:chen:corr:dean:2013b} is entirely linear, but using the softmax transformation henceforth. The latter papers also have some other ingredients which have made them extremely influential. 

An essential feature of the papers by \cite{miko:chen:corr:dean:2013a,miko:suts:chen:corr:dean:2013b} and related papers is that they have found clever approximations to simplify and speed up the calculations of \cite{beng:duch:vinc:jauv:2003}.

\subsection{The \citeauthor{miko:chen:corr:dean:2013a} approach: word2vec\label{Mikolov approach}}

We have already presented the input linear representation of word vectors as rows of the weight matrix ${\bf W}$, see \eqref{A22a}. The output layer should consist of conditional probabilities of words in the vocabulary as in \cite{beng:duch:vinc:jauv:2003}, but  \citeauthor{miko:chen:corr:dean:2013a} has a purely linear transformation to the output layer prior to the softmax transformation.

As a further simplification we assume that we have a window passing over a given text with the window  consisting of just two words $w_t,w_{t-1}$ in position $t$ and $t-1$ of the text. Here, $w_t$ is the target word of the text $w_{O}$, $w_{t-1}$ is the input word $w_I$, and the conditional probability $P(w_t|w_{t-1})$ can also be written $P(w_{O}|w_{I})$. This means that there is only one context word $w_{I}$ for the output word, whereas in the case of \cite{beng:duch:vinc:jauv:2003} there were $l-1$ context words. (Note that in Skip-Gram, and the use of it in network embedding, the context words are more naturally being thought of as target words belonging to the output.) To describe the transition from the hidden layer to the output layer we introduce a new $m \times n$ dimensional weight matrix ${\bf W}^{\prime} = \{w_{ij}^{\prime}\}$. Let $v_{w_{j}}^{\prime}$ be the $j$th column of the matrix ${\bf W}^{\prime}$ (it has dimension $m$). It is the output vector representation of word number $j$ in the vocabulary.
Then the $n$-dimensional output vector is defined by 
$$
y = ({\bf W}^{\prime})^{T}h,
$$ 
where $h=v_{w_{I}}$. Component $y_j$ is given by
\begin{equation}
y_j = (v_{w_{j}}^{\prime})^{T}h,\; j=1,\ldots,n.
\label{A23a}
\end{equation}
To obtain the posterior distribution one uses softmax as defined in (\ref{A21b}),
\begin{equation}
P(w_j|w_{I}) \doteq u_j = \frac{\exp(y_j)}{\sum_{i=1}^{n}\exp(y_i)},  
\label{A23b}
\end{equation}
where now $u_j$ is the transformed output of the $j$th unit in the output layer. By substitution, one obtains
\begin{equation}
P(w_j|w_{I}) = \frac{\exp\Bigl((v_{w_{j}}^{\prime})^{T} v_{w_{I}}\Bigr)}{\sum_{i=1}^{n}\exp\Bigl(w_{i}^{\prime})^{T}v_{w_{I}}\Bigr)}. 
\label{A23c}
\end{equation}
It should be noted that one gets two distinct word representations $v_w$ and $v_{w}^{\prime}$ for each word $w$ in the vocabulary, one input and one output word vector. The output vector is the relevant one in the sense that the context relations are baked into it. Since the system is completely linear, there are no extra parameters to be learned from the network, ``just'' the matrices ${\bf W}$ and ${\bf W}^{\prime}$.

The network is trained by stochastic gradient descent as in \cite{beng:duch:vinc:jauv:2003} and most other neural network applications. Given the input word $w_{I}$ and the output word $w_{O}$, one is interested in maximizing the conditional probability $P(w_{O}|w_{I})$; i.e., finding the index $j = j^{\star}$ and the corresponding probability $u_j$ in the output layer so that, using (\ref{A23b}),
\begin{equation}
\max u_{j} = \max P(w_{O}|w_{I})  \quad \mbox{or} \quad \max \log u_{j} = y_{j^{\star}} - \log \sum_{i=1}^{n} \exp(y_{i}).
\label{A23d}
\end{equation}
By taking derivatives   one gets the update equation
$$
(w_{ij}^{\prime})^{(\mbox{new})} = (w_{ij}^{\prime})^{(\mbox{old})}-\eta e_j h_i,
$$
or
\begin{equation}
(v_{w_{j}}^{\prime})^{(\mbox{new})} = (v_{w_{j}}^{\prime})^{(\mbox{old})} - \eta  e_j  h_i. 
\label{A23e}
\end{equation}
for $j=1,\ldots, n$, where $\eta > 0$ is the learning rate and $e_j = u_j-t_j$ with $t_j = 1(j= j^{\star})$. One has to go through every word in the vocabulary, check its output probability $u_j$, and compare $u_j$ with its targeted output, either 0 or 1.

Going through the same exercise for the transition between the input and the hidden layer, one obtains (see \cite{rong:2016} for details) for the update equation if $w_{I} = w_i$ 
$$
v_{w_{i}}^{(\mbox{new})} = v_{w_{i}}^{(\mbox{old})} - \eta F,
$$ 
where $F$ is the vector whose $i$th component, using back propagation, is given by $\sum_{j=1}^{n}e_j w_{ij}^{\prime}$. Recall that $v_{w_{I}}^{T}$ is a row of ${\bf W}$, the ``input word vector'' of the  only context word $w_{I}=w_i$, and it is the only row of ${\bf W}$ whose derivative is non-zero. All the other rows will remain unchanged after this iteration, since their derivatives are zero.

The generalization from a one word context to a context with several words is quite straightforward in the \cite{miko:chen:corr:dean:2013a,miko:suts:chen:corr:dean:2013b} set-up. They distinguish between two ways of doing this, the CBOW and the Skip-Gram model.

Traditional text classification is based solely on frequencies in the text of words in the vocabulary. This is the bag of words (BOW) approach. \cite{miko:chen:corr:dean:2013a,miko:suts:chen:corr:dean:2013b} take context into account resulting in a continuous bag of words (CBOW). We are then essentially back to the situation in \cite{beng:duch:vinc:jauv:2003} where there are $C = l-1$ context words and we want to maximize $P(w_{O}|w_{1},\ldots , w_{C}$), but  \citeauthor{miko:chen:corr:dean:2013a} assume linearity in the concatenated $C$ words in such a way that the concatenated word vector corresponding to $[w_{1},\ldots, w_{C}]$ is simply given by the average $\frac{1}{C}(v_{w_{1}}+ \cdots + v_{w_{C}})$ of the individual pairwise word vectors.  The hidden layer is then given by
\begin{align}
h &= \frac{1}{C}{\bf W}^{T}(x_1+x_2+ \cdots x_C) \nonumber \\
&= \frac{1}{C}(v_{w_{1}} + \dots + v_{w_{C}}).
\label{A23f}
\end{align} 
This is the CBOW assumption. With this assumption one is more or less back to the one-context word updates. The loss function can be written (cf.\ (\ref{A23a}) and (\ref{A23d})),
\begin{gather}
E = -\log P(w_{O}|w_{1},\cdots w_{C}) \nonumber \\
= -y_{j^{\star}} + \log \sum_{i=1}^{n} \exp(y_i) = -(v_{w_{O}}^{\prime})^{T}  h + \log \sum_{i=1}^{n} \exp((v_{w_{i}}^{\prime})^{T}  h), 
\label{A23g}
\end{gather}
which is the same as (\ref{A23d}), the objective of the one-word context model, except that $h$ is different, being defined as in (\ref{A23f}) instead of in (\ref{A22a}). This leads to an update equation for the output words which is identical to (\ref{A23e}), whereas the update equation for input words has to be updated separately for every word $w_{c},\;c = 1,\ldots, C$, namely
$$
v_{w_{c}}^{(\mbox{new})} = v_{w_{c}}^{(\mbox{old})} - \frac{1}{C}  \eta F,
$$
where $F$ is defined as before.

\subsection{The Skip-Gram model\label{Skip-Gram}}

The Skip-Gram model is in a sense the opposite of the CBOW model, and this is the situation considered in the network embedding in Section 5.3. It is also different from the Bengio model. For a window centered at the word $w_{I}$, the window contains $C/2$ (with $C$ being an even number) words before the center word $w_{I}$ and $C/2$ word after the center word, so that in the notation of \cite{beng:duch:vinc:jauv:2003} the window consists of the words $[w_{t+C/2},\ldots, w_t,\ldots,w_{t-C/2}]$. Sliding the window, the objective is to predict each of the $C$ context words (i.e.\ maximize the conditional probability) $[w_{t+C/2},\ldots,w_{t+1},w_{t-1},\ldots,w_{t-C/2}]$  given the input word $w_{I}=w_t$. Here, conditional independence is assumed, so that the conditional probability for each context word is maximized separately.

For the input word representation the derivation in the two word case is the same as the present situation for the input word and with the same definition of the hidden layer $h$, so that we still have $h_I =  v_{w_{I}}^{T}$. Instead of outputting one (multinomial) distribution, we are outputting $C$ (multinomial) distributions. But, importantly, each output is computed using the same matrix ${\bf W}^{\prime}$ mapping the hidden layer into the output layer. (This means that the {\it sequencing} of the context words does not matter, only {\it which} words are there in the window). Moreover,
$$
P(w_{c,j}|w_{I})  = \frac{\exp(y_{c,j})}{\sum_{i=1}^{n}\exp(y_i)},
$$
where $w_{c,j}$, $c=1,\ldots,C$, $j=1,\ldots,n$, and where the index $j$ is referring to the number in the vocabulary of the word $w_{O,c}$.
Further for $h=v_{w_{i}}$, 
$$
y_{c,j}  = (v_{w_{j}}^{\prime})^{T}   h,
$$
for $c=1,\ldots,C$, where $v_{w_{j}}^{\prime}$ is the output vector for the $j$th word $w_j$ in the vocabulary, and also $v_{w_{j}}^{\prime}$ is taken from the $j$th column of weight matrix ${\bf W}^{\prime}$ transforming the hidden layer to the output layer.

The derivations of the parameter update equations are similar to the one-word context. Assuming conditional independence, the loss function in (\ref{A23g}) is changed to 
$$
E = -\log P(w_{O,1},\dots,w_{O,C}|w_{I}) = -\sum_{c=1}^{C}(v_{w_{c}}^{\prime})^{T}  v_{w_{I}} + C \log \sum_{i=1}^{n} \exp\{(v_{w_{i}}^{\prime})^{T}  v_{w_{I}}\}.
$$
The updating equations can be derived by taking derivatives similarly to the CBOW case, and we refer to \cite{rong:2016} for details.

In spite of the relatively simple linear structure of CBOW and Skip-Gram, it  makes for some quite astonishing properties that goes beyond simple syntactic regularities. This is obtained using just very simple algebraic operations in the word representation space $\mathbb{R}^m$, such that for example the embedded word vector(``King'')-word vector(``Man")+word vector(``Woman'') has a high probability of having the word vector(``Queen'') as its closest word vector, as measured by cosine distance in word feature space $\mathbb{R}^m$. Several similar examples are given in \cite{miko:chen:corr:dean:2013a,miko:suts:chen:corr:dean:2013b}, and they have also examined quite systematically the capabilities of CBOW and Skip-Gram compared to other word representation routines in solving such tasks.

\subsection{The computational issue\label{Computational issue}}

For all of the word models presented so far, there is a computational issue. As the size of the vocabulary and the size of the training text set increase, they are heavy to update. For the two-word, the CBOW and the Skip-Gram models there are two vector representations for each word in the vocabulary: the input vector $v_w$ and the output vector $v_w^{\prime}$. Learning the input vectors is cheap, but learning the output vectors is expensive. From the update equations (\ref{A23b}), (\ref{A23c}), (\ref{A23d}) and (\ref{A23e}) it is seen that to update $v_w^{\prime}$ for each training instance, one has to iterate through every word $w_j$ in the vocabulary, compute $y_j$, the  prediction error $e_j$  and finally use the prediction error to update the output vector $v_{w_{j}}^{\prime}$.

Such kind of computations makes it difficult to scale up to large vocabularies or large training corpora. The obvious solution to circumvent this problem is to limit the number of output vectors that must be updated per training instance. There are two main approaches for doing this, hierarchical softmax and negative sampling. Both approaches optimize {\it only} the computation for updates for output vectors. 

Hierarchical softmax is an efficient way of computing softmax \citep{mori:beng:2005,mnih:hint:2008}. With this method the frequency of words appearing in texts is taken into account. In hierarchical softmax the list of words from word 1 to word $n$ is replaced by a binary Huffman encoded tree with the $n$ words appearing at the leaves (outer branches) of the tree. The probability of the occurrence of a word given an input word is computed from a probability path from the root of the tree to the given word. This reduces the number of operations in an update from $n$ to $\log _2 n$, e.g.\ for $n$ = 1 million = $10^{6}$, the number of operations are reduced to $6  \log_2 10 \approx 20$. We refer to \cite{mori:beng:2005} and \cite{mnih:hint:2008} for a detailed description of hierarchical softmax.

\subsection{Negative sampling\label{Negative sampling}}

The idea of negative sampling is far more straightforward than hierarchical softmax. It is sampling-based, and for each updating instance, only a sample of output vectors are used. This seems to be an, perhaps {\it the}, essential idea that makes Skip-Gram work so well.

Obviously the output words; i.e.\ $w_{O}$ in CBOW and each of the
words $w_{O,c}$ for $c=1,\ldots,C$ in the Skip-Gram procedure should
be included in the updating sample. They represent the ground truth
and are termed  positive samples. In addition, a certain number $k$ of
word vectors (noise or negative samples) are updated, such that
$k=5-20$ are useful for small training sets, whereas for large
training sets, $k= 2-5$ may be sufficient
\citep{miko:suts:chen:corr:dean:2013b}. The sampling is carried out via a
probability mechanism where each word is sampled according to its
frequency  $f(w_i)$ in the text. In addition,
\citeauthor{miko:suts:chen:corr:dean:2013b} recommend from empirical
experience that each word is given a weight equal to its frequency
(word count) raised to the $3/4$ power. The probability for selecting
a word (vector) is just its weight divided by the sum of weights
for all words, i.e.,  
$$
P_n(w_i) = \frac{f(w_i)^{3/4}}{\sum_{j=1}^{n}f(w_j)^{3/4}}.
$$  

In addition, in word2vec, instead of using the loss functions (\ref{A23d}) and (\ref{A23g}) constructed from multinomial distributions, the authors argue that the following simplified training objective is capable of producing high-quality word embeddings:
\begin{equation}
E = -\log \sigma((v_{w_{O}}^{\prime})^{T} h) - \sum_{w_j \in {\mathcal W}_{\mbox{neg}}} \log \sigma(-(v_{w_{j}}^{\prime})^{T}  h), 
\label{A25a}
\end{equation}
where $\sigma(u)$ is the logistic function given by $\sigma(u) = 1/(1+\exp(-u))$ and ${\mathcal W}_{\mbox{neg}}$ is the collection of negative samples for the given update. Further, $w_{O}$ is the output word (the positive sample), $v_{w_{O}}^{\prime}$ is the output vector; $h$ is the value of the hidden layer with $h = \frac{1}{C}\sum_{c=1}^{C} v_{w_{c}}$ in the CBOW model and $h = v_{w_{I}}$ in the Skip-Gram model. Note that \citeauthor{miko:chen:corr:dean:2013a} write (\ref{A25a}) as
$$
E = -\log \sigma((v_{w_{O}}^{\prime})^T h)-\sum_{i=1}^k E_{w_{i}\sim P_n(w)} \log \sigma(-(v_{w_{i}}^{\prime})^T h).
$$
To obtain the update equations we again use the chain rule of differentiation. First, the derivative of $E$ with respect to $(v_{w_{j}}^{\prime})^{T} h$ is computed as 
\[
\frac{\partial E}{\partial ((v_{w_{j}}^{\prime})^{T} h)} = 
\left\{
\begin{tabular} {cc}
$\sigma((v_{w_{j}}^{\prime})^{T} h)-1$ & if $w_j = w_{O}$\\
$\sigma((v_{w_{j}}^{\prime})^{T} h)$ & if $w_j \in {\mathcal W}_{\mbox{neg}}$
\end{tabular}
\right\},
\]
which results in the derivative being equal to $\sigma((v_{w_{j}}^{\prime})^{T} h)-t_j$ where $t_j$ is the label of word $w_j$ such that $t_j = 1$ if $w_j$ is a positive sample, and 0 otherwise. Next, we take the derivative of $E$ with regard to the output vector of the word $w_j$,
$$
\frac{\partial E}{\partial v_{w_{j}}^{\prime}} = \frac{\partial E}{\partial ((v_{w_{j}}^{\prime})^{T}  h)}\frac{\partial ((v_{w_{j}}^{\prime})^{T} h)}{\partial v_{w_{j}}^{\prime}} = \Bigl(\sigma((v_{w_{j}}^{\prime})^{T} h)-t_j\Bigr) h.
$$
This results in the following update equation for the output vector
$$
v_{w_{j}}^{\prime \; (\mbox{new})} = v_{w_{j}}^{\prime \; (\mbox{old})} - \varepsilon \Bigl(\sigma((v_{w_{j}}^{\prime})^{T} h) - t_j\Bigr) h,
$$
which only needs to be applied to $w_j \in \{w_{O}\} \cup {\mathcal W}_{\mbox{neg}}$ instead of every word in the vocabulary. This equation can be used both for CBOW and the Skip-Gram model. For the Skip-Gram model, the equation has to be applied for one context word at a time.

To back-propagate the error to the hidden layer and thus update the input vectors of words, it is necessary to take the derivative of $E$ with regard to the hidden layer's output, obtaining
\begin{align*}
  \frac{\partial E}{\partial h}
  &= \sum_{w_j \in \{w_{O}\} \cup {\mathcal
                                W}_{\mbox{neg}}} \frac{\partial
                                E}{\partial (v_{w_{j}}^{\prime})^{T}
                                h}\frac{\partial
                                (v_{w_{j}}^{\prime})^{T} h}{\partial
                                h} \\
  &= \sum_{w_j \in \{w_{O}\} \cup {\mathcal W}_{\mbox{neg}}} \Bigl(\sigma((v_{w_{j}}^{\prime})^{T} h)-t_j\Bigr)v_{w_{j}}^{\prime} \doteq F.
\end{align*}
Using this, one can obtain update equations for the input vectors of the CBOW and Skip-Gram models.

\subsection{Some results\label{Some results}}

There are a number of results for variously structured text data sets
in \cite{miko:chen:corr:dean:2013a,miko:suts:chen:corr:dean:2013b},
where it is seen that CBOW and Skip-Gram perform well compared to
other methods and that with negative sampling or hierarchical softmax
the methods can be applied to vocabularies in the millions and text
samples in the billions of words. Choices of parameters such as the
number of context words (not much greater than 10), sample size of
negative samples, and dimension of word vectors are
discussed. Further, there are several experiments analyzing the
sensitivity of the results on applications to empirical data. The
Skip-Gram is a slightly heuristic method when combined with negative 
sampling (such as a sudden shift from one objective function to another one, raising the empirical frequencies to an exponent of {3/4}). The authors justify this from the empirical results obtained, which are quite impressive. There are several papers attempting to simplify and complement the rather brief description in the papers by \cite{miko:chen:corr:dean:2013a,miko:suts:chen:corr:dean:2013b}, and trying to give it  a firmer mathematical basis. We have found \cite{rong:2016} useful. The shift of objective function is sought explained in  \cite{gold:levy:2014}.

There are extensions to classification of text extending the context
of word-vector to the concept of paragraph-vector in
\cite{le:miko:2014}, but it is very concisely written. There is also a
paper on machine translation by \cite{miko:le:suts:2013c}. Software is easily available for all of the algorithms described in this section.

\section{ A more involved illustrating example \label{More involved} }

\begin{figure}[ht]
  \begin{center}
    \begin{subfigure}{1\textwidth}
      \centering
      \includegraphics[width=1\textwidth]{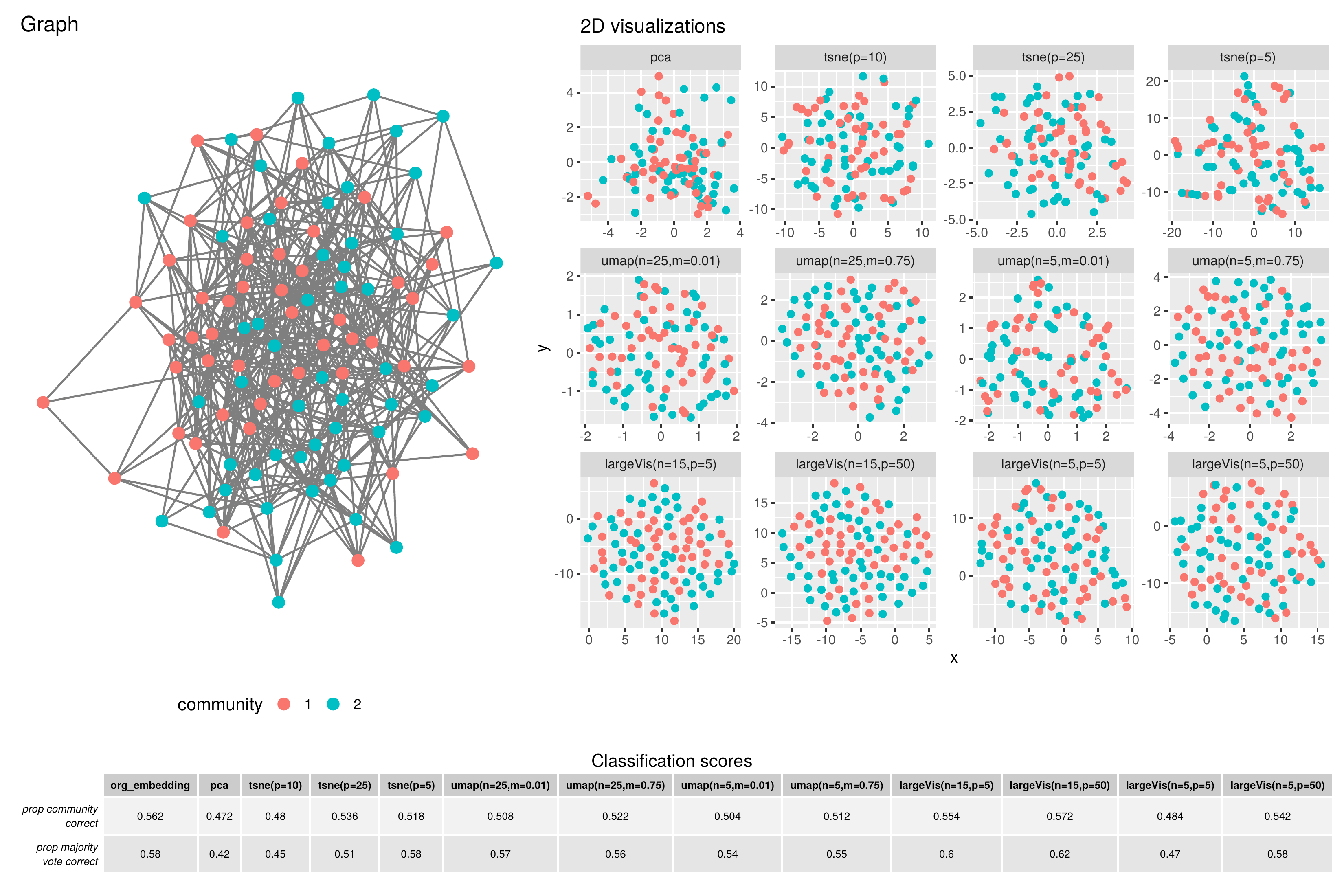}
      \caption{Homogeneous graph from degree corrected stochastic block model.}
      \label{fig:suba1a}
    \end{subfigure}
    \begin{subfigure}{1\textwidth}
      \centering
    \includegraphics[width=1\textwidth]{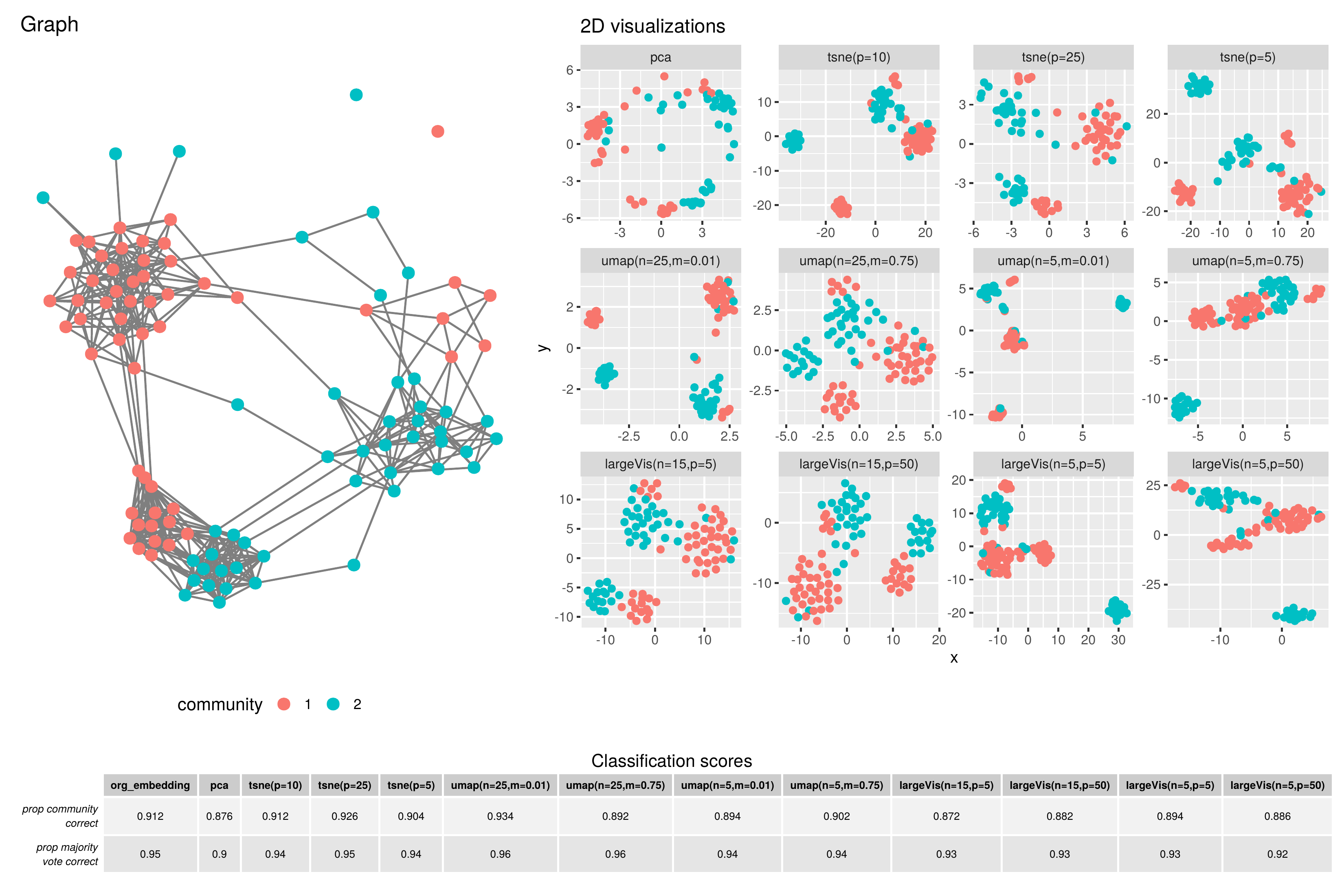}  
      \caption{Heterogeneous graph from a combination of three degree corrected stochastic block models.}
      \label{fig:suba1b}
    \end{subfigure}    
\end{center}
\caption{Graphs, visualizations and classification results with a $k$-nearest neighbors algorithm with $k=5$.
 \label{fig:a1}}
\end{figure}

Fig.\ \ref{fig:a1} contains more challenging variants of the graphs in
Fig.\ 5 in the main paper. The homogeneous graph in Fig.\ \ref{fig:suba1a} is  simulated from a stochastic block model  with 2 communities,
100 nodes, average node degree $d = 10$ and ratio of between-community
edges over within-community edges $\beta=0.75$, i.e.~it is generated from the same model as Fig.\ 5a in the paper, except that $\beta$ has increased form 0.4 to 0.75. As for Fig.\ 5 in the main paper, embeddings with dimension 64 were computed using
node2vec with 30 nodes in each walk with 200 walks per node, and a
word2vec window length of 10 where all words are included. The
accompanying 2-dimensional visualizations of the embeddings are done with PCA
and $t$-SNE, UMAP and LargeVis, all with different tuning parameters.

Compared to Fig.\ 5a in the main paper, the PCA is far inferior to
the three other embedded visualizations for this more involved example. 
Similarly to Fig.\ 5b in the main paper, the heterogeneous graph in Fig.\ \ref{fig:suba1b} is simulated from three
degree corrected stochastic block models (three subgraphs ${\bf a}$, ${\bf b}$ and ${\bf c}$, each with 2 communities:
\begin{description}
\item[Graph a:] 30 nodes, average node degree $d = 7$, ratio of
  between-community edges over within-community edges $\beta=0.1$
\item[Graph b:] 30 nodes, average node degree $d = 15$, ratio of
  between-community edges over within-community edges $\beta=0.2$
\item[Graph c:]  40 nodes, average node degree $d = 7$, ratio of
  between-community edges over within-community edges $\beta=0.1$, and an
  unbalanced community proportion; a probability of 3/4 for community 1
  and a probability of 1/4 for community 2
\end{description}
To link graphs a, b and c, some random edges are added between nodes from
the same community\footnote{For each pair of nodes between a pair of graphs, say Graph a and
  c, a new link is randomly sampled with a probability of 0.01, and
   links  connecting two nodes from the same community are kept.}. The
 results are somewhat similar to those of Fig.\ 5b of the main
 paper. Again,  PCA is inferior to the three other methods, but it is closer than in Fig.\  \ref{fig:suba1b}.

 \clearpage
 
\bibliographystyle{apalike}
\bibliography{dagemb}